\documentclass{article}

\PassOptionsToPackage{dvipsnames,table}{xcolor}

\usepackage[preprint]{neurips_2022} %[preprint]

\usepackage{amsmath,amsfonts,bm}

\def\eqref#1{equation~\ref{#1}}
\def\1{\bm{1}}

\def\vb{{\bm{b}}}
\def\vc{{\bm{c}}}
\def\vd{{\bm{d}}}

\def\vu{{\bm{u}}}
\def\vv{{\bm{v}}}

\def\vx{{\bm{x}}}

\def\vz{{\bm{z}}}

\def\evb{{b}}

\def\evd{{d}}

\def\mW{{\bm{W}}}

\DeclareMathAlphabet{\mathsfit}{\encodingdefault}{\sfdefault}{m}{sl}
\SetMathAlphabet{\mathsfit}{bold}{\encodingdefault}{\sfdefault}{bx}{n}

\def\gC{{\mathcal{C}}}
\def\gD{{\mathcal{D}}}

\def\gL{{\mathcal{L}}}

\def\sD{{\mathbb{D}}}
\newcommand{\R}{\mathbb{R}}

\DeclareMathOperator*{\argmax}{arg\,max}
\DeclareMathOperator*{\argmin}{arg\,min}

\DeclareMathOperator{\sign}{sign}

\usepackage[utf8]{inputenc} % allow utf-8 input
\usepackage[T1]{fontenc}    % use 8-bit T1 fonts
\usepackage{hyperref}       % hyperlinks
\usepackage{url}            % simple URL typesetting
\usepackage{booktabs}       % professional-quality tables
\usepackage{amsfonts}       % blackboard math symbols
\usepackage{nicefrac}       % compact symbols for 1/2, etc.
\usepackage{microtype}      % microtypography
\usepackage{xcolor}         % colors
\usepackage{xfrac}

\makeatletter
\def\@fnsymbol#1{\ensuremath{\ifcase#1\or \dagger\or *\or \ddagger\or
   \mathsection\or \mathparagraph\or \|\or **\or \dagger\dagger
   \or \ddagger\ddagger\or \mathsection\mathsection
   \or \mathparagraph\mathparagraph \or \|\|\else\@ctrerr\fi}}

\makeatother

\usepackage{authblk}

\usepackage{amsthm}
\newtheorem{theorem}{Theorem}[section]
\theoremstyle{definition}
\newtheorem{definition}{Definition}[section]

\newtheorem{lemma}{Lemma}[section]
\usepackage{graphicx}
\usepackage{multirow}
\usepackage{soul}
\usepackage{changepage,threeparttable}
\usepackage[caption = false]{subfig}
\usepackage{wrapfig,lipsum}
\usepackage{makecell}
\usepackage[ruled,linesnumbered,norelsize]{algorithm2e}

\SetKwComment{Comment}{$\triangleleft$\ }{}

\SetCommentSty{mycommfont}
\usepackage{pifont}% http://ctan.org/pkg/pifont
\newcommand{\cmark}{\ding{52}}%
\newcommand{\xmark}{\ding{56}}%
\usepackage{textcomp}

\usepackage{xr}
\makeatletter
\newcommand*{\addFileDependency}[1]{% argument=file name and extension
  \typeout{(#1)}
  \@addtofilelist{#1}
  \IfFileExists{#1}{}{\typeout{No file #1.}}
}
\makeatother

\newcommand*{\myexternaldocument}[1]{%
    \externaldocument{#1}%
    \addFileDependency{#1.tex}%
    \addFileDependency{#1.aux}%
}
\myexternaldocument{appendix}

\usepackage{tikz}
\usepackage{scalerel}
\newcommand{\tikztriangleright}[1][red,fill=red]{\scalerel*{\tikz \draw[rounded corners=0.1pt,#1] (0,-2.5pt)--++(0,5pt)--++(-30:5pt)--cycle;}{\triangleright}}
\usepackage{amssymb}

\title{Progressive Voronoi Diagram Subdivision: Towards A Holistic Geometric Framework for Exemplar-free Class-Incremental Learning}

\author[1]{\textbf{Chunwei Ma}}
\author[1]{\textbf{Zhanghexuan Ji}}
\author[2]{\textbf{Ziyun Huang}}
\author[1]{\textbf{Yan Shen}}
\author[1]{\textbf{Mingchen Gao}}
\author[1]{\textbf{Jinhui Xu}}
\affil[1]{%
    Department of Computer Science and Engineering\\
    University at Buffalo % \\
}
\affil[2]{%
    Computer Science and Software Engineering\\
    Penn State Erie % \\
}
\affil[1]{\texttt{\{chunweim,zhanghex,yshen22,mgao8,jinhui\}@buffalo.edu}}
\affil[2]{\texttt{\{zxh201\}@psu.edu}}

\begin{document}

\maketitle

\begin{abstract}
  Exemplar-free Class-incremental Learning (CIL) is a challenging problem because %the rehearsal of 
  rehearsing data from previous phases is strictly prohibited, causing catastrophic forgetting of Deep Neural Networks (DNNs). In this paper, we present \emph{iVoro}, a holistic framework derived from computational geometry. We found Voronoi Diagram (VD), a classical model for space subdivision, is especially powerful for solving the CIL problem, because VD itself can be constructed favorably in an incremental manner  
  % of its characteristic of convenient incremental construction
  % VD itself is friendly to incremental construction 
  -- the newly added sites (classes) will only affect the proximate classes, making the non-contiguous classes hardly forgettable.
  %impossible to be forgotten 
  % unchanged
  Furthermore, in order to find a better set of centers for VD construction, we colligate DNN with VD using Power Diagram and show that the VD structure can be optimized by integrating local DNN models using a divide-and-conquer algorithm. Moreover, our VD construction is not restricted to the deep feature space, but is also applicable to multiple intermediate feature spaces, promoting VD to be multi-centered VD that efficiently captures multi-grained features from DNN. Importantly, \emph{iVoro} is also capable of handling uncertainty-aware test-time Voronoi cell assignment and has exhibited high correlations between geometric uncertainty and predictive accuracy (up to ${\sim}0.9$). Putting everything together, \emph{iVoro} achieves up to $25.26\%$, $37.09\%$, and $33.21\%$ improvements on CIFAR-100, TinyImageNet, and ImageNet-Subset, respectively, compared to the state-of-the-art non-exemplar CIL approaches. In conclusion, \emph{iVoro} enables highly accurate, privacy-preserving, and geometrically interpretable CIL that is particularly useful when cross-phase data sharing is forbidden, e.g. in medical applications. Our code is available at https://machunwei.github.io/ivoro.
\end{abstract}
% GitHub\footnote{Temporarily available at https://anonymous.4open.science/r/Y37WdxZuo2i2-8F57.}.
% https://machunwei.github.io/ivoro.
% ==================== ==================== ==================== ==================== ====================
% Introduction
% ==================== ==================== ==================== ==================== ====================
\section{Introduction} \label{sec:intro}
\setlength\intextsep{-10pt} % change this value
\begin{wrapfigure}{r}{0.55\textwidth} % change this value
    \centering
    \subfloat{\includegraphics[height=1.0in]{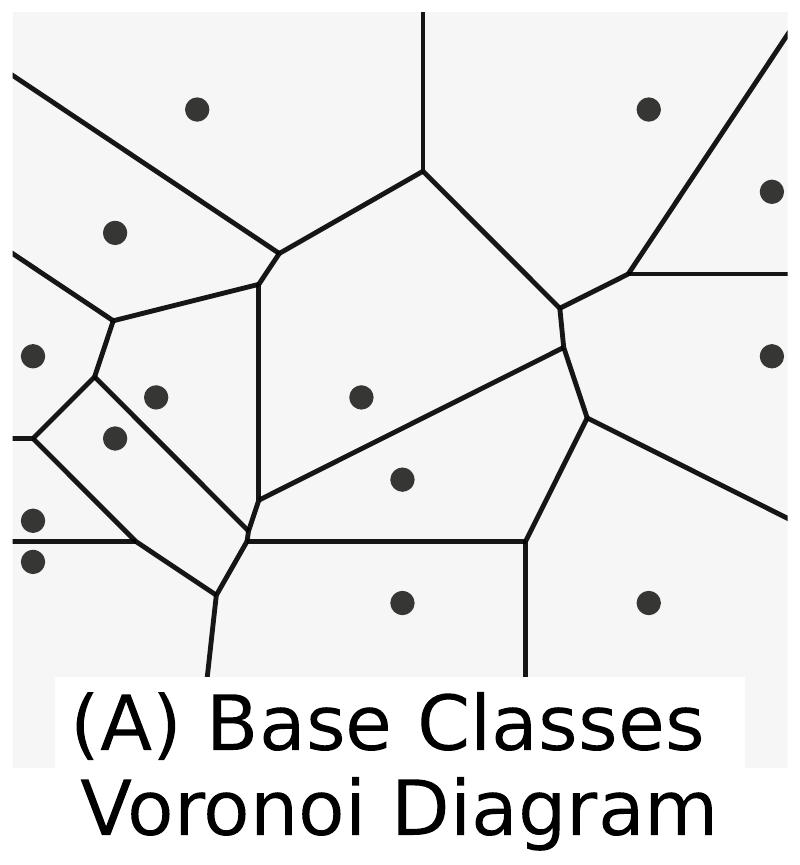}} 
    \subfloat{\includegraphics[height=1.0in]{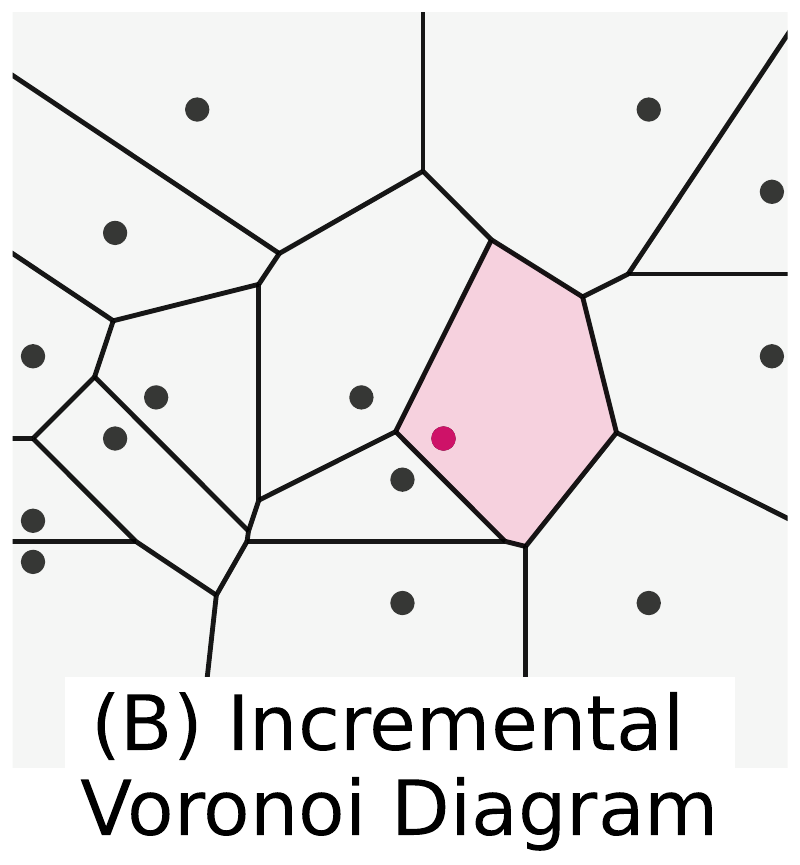}} 
    \subfloat{\includegraphics[height=1.0in]{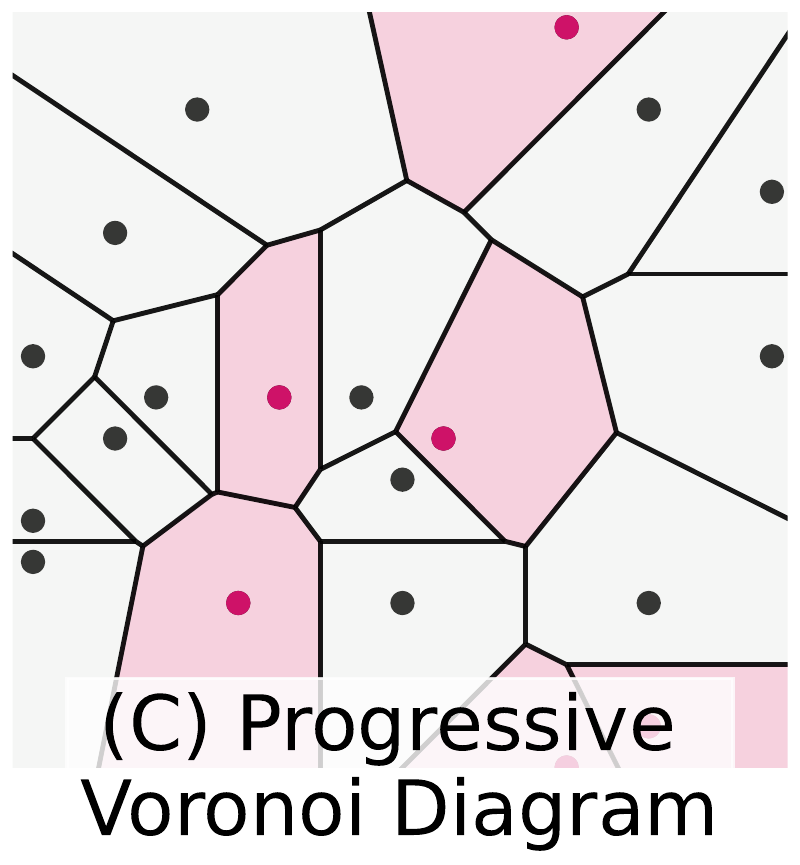}} 
    \vspace{-3mm}
	\caption{Schematic illustrations of Voronoi Diagram (VD) for base sites (A), and when a new site (B) or a clique of new sites (C) is added to the system.}
	\label{fig:vd_demo}
    \vspace{3mm} % change this value
\end{wrapfigure}

% introduce the background of IL
In many real-world applications, e.g., medical imaging-based diagnosis, the learning system is usually required to be expandable to new classes, for example, from common to rare inherited retinal diseases (IRDs)~\citep{miere2020deep}, or from coarse to fine chest radiographic findings~\citep{syeda2020chest}, and importantly, without losing the knowledge already learned. This motivates the concept of \emph{incremental learning} (IL)~\citep{hou2019learning,wu2019large,zhu2021prototype,liu2021rmm}, also known as \emph{continual learning}~\citep{parisi2019continual,9349197,chaudhry2019tiny}, which has drawn growing %a 
interests in recent years. Although Deep Neural Networks (DNNs) have become the \textit{de facto} method of choice due to their extraordinary ability of learning from complex data, they still suffer from severe catastrophic forgetting~\citep{mccloskey1989catastrophic,goodfellow2013empirical,kemker2018measuring} when adapting to new tasks that contain only unseen training samples from novel classes. 

% why exemplar-free? because of privacy
To mitigate this issue, Rebuffi et al.~\citep{rebuffi2017icarl} proposed the paradigm of memory-based class-incremental learning (CIL)~\citep{belouadah2019il2m,zhao2020maintaining,hou2019learning,castro2018end,wu2019large,liu2021adaptive,liu2020mnemonics,liu2021rmm} in which a small portion of samples (e.g., 20 exemplars per class) will be stored to use in the subsequent phases. However, the storing and sharing of data, e.g. medical images, may not be feasible due to privacy considerations. Another line of methods memorize (part of) network and increase the model capacity for new classes~\citep{rusu2016progressive,li2019learn,wang2017growing,yoon2017lifelong}, which may incur unbounded memory consumption for long task sequence. Hence, in this paper, we focus on the challenging exemplar-free CIL problem under the strictest memory and privacy constraints -- no stored data and fixed model capacity.

\begin{figure}
    \centering
    \subfloat{\includegraphics[width=0.32\paperwidth]{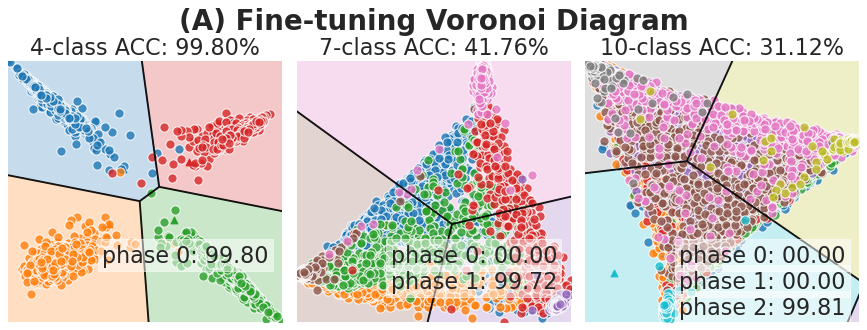}} 
    \subfloat{\includegraphics[width=0.32\paperwidth]{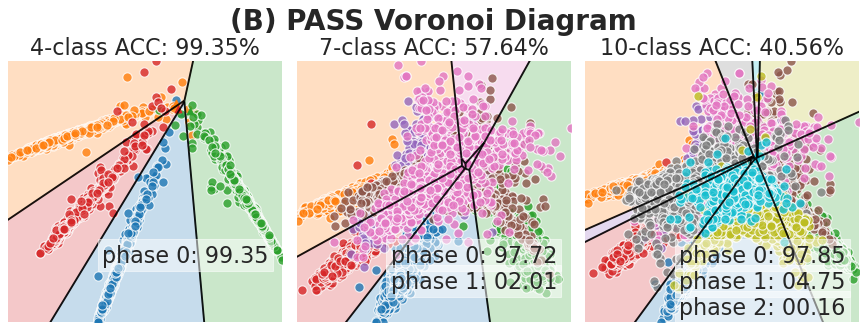}} \\ [-0.3ex]
    \subfloat{\includegraphics[width=0.32\paperwidth]{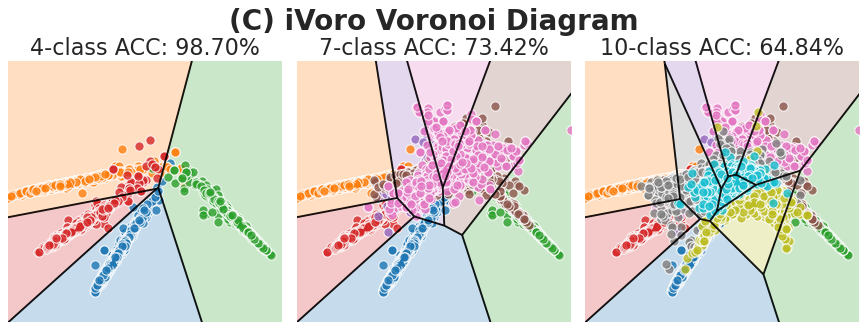}} 
    \subfloat{\includegraphics[width=0.32\paperwidth]{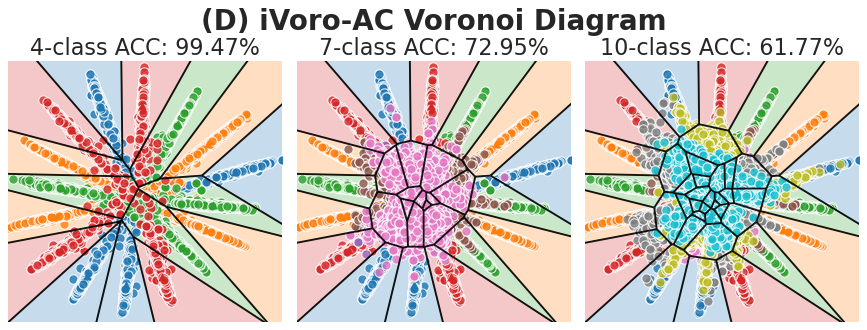}} 
    \caption{Visualization of of Voronoi Diagrams induced by (A) incremental fine-tuning, (B) PASS, (C) iVoro, and (D) iVoro-AC on MNIST dataset in $\R^2$ (best viewed in color). The dataset was split to 4, 3, and 3 disjoint classes. (See Appendix~\ref{supp:mnist} for details.)}\label{fig:mnist-vd}
    \vspace{-5mm} % change this value
\end{figure}

% three obstacles
Despite extensive research in recent years (see Appendix~\ref{supp:related} for a literature review), three challenges still pose an obstacle to successful CIL. 
\textbf{(I)} During the course of isolated training upon new data, the feature distributions of the old classes are usually dramatically changed (see Fig.~\ref{fig:mnist-vd} (A) for an illustration). Knowledge Distillation (KD)~\citep{hinton2015distilling} has become a routine in many CIL methods~\citep{li2017learning,schwarz2018progress,castro2018end,hou2019learning,dhar2019learning,douillard2020podnet,zhu2021prototype} to partially maintain the spatial distribution of old classes. However, KD loss is typically applied onto the whole network, and a strong KD loss may potentially degenerate the network's ability to adapt to novel classes. %(Fig.~\ref{fig:mnist-vd} (B)). 
\textbf{(II)} Without the full access to old data, the decision boundaries cannot be learned precisely, making it harder to discriminate between old and new classes. Taking inspiration from metric-based Few-shot Learning (FSL)~\citep{snell2017prototypical}, PASS~\citep{zhu2021prototype} memorizes a set of prototypes (feature centroids) and generates features augmented by Gaussian noise for a joint training in new phases. However, feature means might be suboptimal to represent the whole class, which is not necessarily normally distributed (Fig.~\ref{fig:mnist-vd} (B)). 
\textbf{(III)} Since the old classes and the new classes are learned in a disjoint manner, their distributions are likely to be overlapped, which becomes even severer in our exemplar-free setting as the old data is totally absent. 
To circumvent this issue, \textit{Task-incremental learning} (TIL)~\citep{shin2017continual,kirkpatrick2017overcoming,zenke2017continual,wu2018incremental,lopez2017gradient,buzzega2020dark,cha2021co2l,pham2021dualnet,fernando2017pathnet} assumes the phase within which a class was learned is known, which is generally unrealistic in practice. CIL, however, are not grounded on this assumption.
Self-supervised Learning (SSL)~\citep{lee2020self,chen2020simple,he2021masked}, on the other hand, has shown potential for alleviating task-level overfitting by learning representations transferable across phases. However, SSL in CIL is still largely underexplored and restricted to simple operations e.g. rotation.

In this paper, we tackle the CIL problem from a geometric point of view. We found that Voronoi Diagram (VD), a classic model for space subdivision that has been intensively studied for decades, bears a close analogy to incremental learning, because VD itself is friendly to incremental construction -- the newly added sites (classes) will roughly change only the cells of the neighboring classes, making the non-contiguous classes untouched and thus impossible to be forgotten. Based on this intuition, we decompose CIL into three subproblems: the delineation of Voronoi boundary \textbf{(a)} between old and old classes, \textbf{(b)} between new and new classes, and \textbf{(c)} between old and new classes, and accordingly design a complete pipeline that can handle \textbf{(1)} VD construction from prototypes, \textbf{(2)} progressive VD refinement during phases, \textbf{(3)} prototype optimization, \textbf{(4)} uncertainty quantification for Voronoi cell assignment, and \textbf{(5)} multi-layer VD for deep neural network, becoming a holistic geometric framework that %greatly 
significantly overcomes all the listed obstacles. The contributions can be summarized as follows:

\noindent\textbf{1. iVoro.} We begin with the simplest scenario in which the feature extractor is frozen after the first phase, and the prototypes ($\{\vc\}$) are used to construct VD, providing a strong baseline (denoted as iVoro) for CIL.

\noindent\textbf{2. iVoro-D.} iVoro treats the aforementioned three subproblems equally and determines the Voronoi boundaries all by bisecting prototypes. However, without considering data distribution, the bisector is certainly not optimal especially within a certain phase. We establish explicit connection between DNN and VD using Voronoi diagram reduction~\citep{ma2022fewshot} and show that local VDs (centered at $\{\tilde{\vc}\}$) can be optimized via DNN and be aggregated into the global VD by a divide-and-conquer (D\&C) algorithm (iVoro-D).

\noindent\textbf{3. iVoro-R.} While iVoro-D provides better Voronoi boundaries for classes within a phase (subproblems \textbf{(a)} and \textbf{(b)}), the old-to-new boundaries are still bisectors and need refinement (subproblem \textbf{(c)}). To do so, we use DNN to learn the "residue" from the vanilla prototypes ($\{\vc\}$) to better prototypes (Voronoi residual prototypes $\{\tilde{\vc}'\}$) that are better representatives of the data distribution (iVoro-R). 

\noindent\textbf{4. iVoro-AC/AI.} Geometrically, SSL-based label augmentation~\citep{lee2020self} will duplicate one Voronoi cell to be multiple (possibly disjoint) Voronoi cells (see Fig.~\ref{fig:mnist-vd} (D)), and this will cause ambiguity when assigning a query example to a cell, implying that uncertainty quantification cannot be neglected in test-time. Here we propose two protocols to resolve this ambiguity, namely, augmentation consensus (iVoro-AC) and augmentation integration (iVoro-AI). We also show that the entropy-based geometric variance~\citep{ding2020learning} is a good indicator of the uncertainty of this assignment, with high Pearson correlation coefficient up to ${\sim}0.9$.

\noindent\textbf{5. iVoro-L.} Until now, only deep features from the last layer are used for VD construction. However, the intermediate feature could also be informative to aid VD construction. Cluster-induced Voronoi Diagram (CIVD)~\citep{ChenHLX13, ChenHL017, HuangX20, HuangCX21}, that allows for multiple centers per Voronoi cell, has recently achieved remarkable success in metric-based FSL by incorporating heterogeneous features to VD~\citep{ma2022fewshot}. As a matter of fact, for a deep neural network, the feature induced by every layer can all be used to construct a VD. Finally, we also explore the idea to build a multi-centered CIVD by features from multiple layers, and show superior final performance (iVoro-L).

\noindent\textbf{6.} Experimentally, we exhaust the combinations of the above five contributions with comprehensive ablation studies. We also experiment with different numbers of phases, different sizes of the base phase, and features from different layers. Finally, our fully-fledged iVoro model achieves up to $25.26\%$, $37.09\%$, and $33.21\%$ improvements on CIFAR-100, TinyImageNet, and ImageNet-Subset, respectively, compared with the state-of-the-art non-exemplar CIL approaches.

% ==================== ==================== ==================== ==================== ====================
% Methodology
% ==================== ==================== ==================== ==================== ====================

\section{Methodology} \label{sec:method}

% ==================== Problem Setting ====================
\subsection{Preliminaries: Class-Incremental Learning} \label{sec:basics}
In CIL, the data comes as a stream and a single model is trained on current data locally without revisiting previous data, but should ideally be able to discriminate between \emph{all} classes it has seen so far. Specifically, let $\gD = \{\gD_t\}_{t=1}^T$ be the data stream in which $\gD_t = \{(\vx_{t,i}, y_{t,i})\}_{i=1}^{N_t}, \vx_{t,i} \in \sD, y_{t,i} \in \gC_t$ is the dataset at time step $t$. $\sD$ is an arbitrary domain, e.g., natural image, and $\gC_t$ is the set of classes at phase $t$. The dataset $\gD_t$ contains $N_{t,k}, k \in \{1,...,K_t\}$ samples for the $K_t$ classes (i.e. $N_t = {\textstyle\sum}_{k=1}^{K_t} N_{t,k} $). Notice that $\gC_i, \gC_j$ for two arbitrary phases $i, j$ are disjoint, i.e. $\gC_i \cap \gC_j = \emptyset, \forall i,j: i \neq j$. The unified model consists of a feature extractor $\phi$ and a classification head $\theta$. The feature extractor is a deep neural network $\vz = \phi(\vx), \vz \in \R^n$ that maps from image domain $\sD$ to feature domain $\R^n$, and is trained continuously at each phase $t$. In this section, $T$, $t$, and $\tau$ denote the total phase, the current phase, and the historical phase, respectively.
% ==================== Contribution 1 ====================
\subsection{Constructing Voronoi Diagrams: A Feature Extractor is All You Need} \label{sec:ivoro}
In many CIL methods, the feature extractor $\phi$ and classification head $\theta$ are jointly and continuously optimized during every phase $t$ guided by %sophisticatedly 
carefully 
designed losses~\citep{zhu2021prototype}. As a starting point, in this section, we freeze the feature extractor $\phi$ after the first phase and use a \emph{Voronoi Diagram} (i.e., a 1-nearest-neighbor classifiers) to be $\theta$ as an extremely simple baseline method (denoted as iVoro), upon which we will then gradually add component introduced in Sec.~\ref{sec:intro}. First, we introduce \emph{Power Diagram} (PD), a generalized version of VD:
\begin{definition}[Power Diagram and Voronoi Diagram] \label{def:power}
    Let $\Omega = \{\omega_1,...,\omega_K\}$ be a partition of the space $\R^n$, and $\gC = \{\vc_1,...,\vc_K\}$ be a set of centers (also called {\em sites}) such that $\cup_{r=1}^K \omega_r = \R^n, \cap_{r=1}^K \omega_r = \emptyset$. In addition, each center is associated with a weight $\nu_r \in \{ \nu_1,...,\nu_K \} \subseteq \R^+$. Then, the set of pairs $\{ (\omega_1, \vc_1, \nu_1),...,(\omega_K, \vc_L, \nu_K) \}$ is a Power Diagram (PD), where each cell is obtained via $\omega_r = \{ \vz \in \R^n : r(\vz) = r\}, r \in \{1,..,K\}$, with $r(\vz) = \argmin_{k \in \{ 1,...,K \}} d(\vz, \vc_k)^2 - \nu_k$. If the weights are equal for all $k$, i.e. $\nu_k = \nu_{k'}, \forall k,k' \in \{1,...,K\}$, then a PD collapses to a Voronoi Diagram (VD).
\end{definition}
\noindent\textbf{Prototypes.} As a baseline model, the class centers for iVoro are simply chosen to be the prototypes (feature mean of one class): $\vc_{\tau, k} = \frac{1}{N_{\tau, k}}\ {\textstyle\sum}_{i \in \{1,...,N_{\tau, k}\}, y = k}\ \phi(\vx_{\tau, i}), \nu_{\tau, k} = 0, \tau \in \{1,...,t\}, k \in \{1,...,K_{\tau}\}$. We name those centers prototypical centers. Note that this set of centers $\{\vc_{\tau, k}\}$ carries prototypes for all classes, old and new, up to time $t$. In test-time, a query sample $\vx$ is assigned to the nearest class $\hat{y} = \gC_{\tau', k'}$ s.t. $d(\vz,\vc_{\tau', k'}) = \min_{\tau, k} d(\vz, \vc_{\tau, k})$ in which $d(\vz, \vc_{\tau, k}) = ||\vz - \vc_{\tau, k}||_2^2$.

\noindent\textbf{Parameterized Feature Transformation.} Although PASS uses Gaussian noise to augment the data, the real features are not necessarily normally distributed. To encourage the normality of feature distribution here we adopt compositional feature transformation commonly used in FSL~\citep{ma2022fewshot}: 
(1) \emph{$L_2$ normalization} projects the feature onto the unit sphere: $ f(\vz) = \frac{\vz}{||\vz||_2} $;
(2) \emph{linear transformation} does the scaling and shifting: $g_{w,\eta}(\vz) = w\vz + \eta$; and (3) \emph{Tukey’s ladder of powers transformation} further improves the Gaussianity:
$h_\lambda(\vz) =\left\{
    \begin{aligned}
        &\vz^\lambda \quad &\text{if}\ \lambda \neq 0 \\
        &\log(\vz)   \quad &\text{if}\ \lambda = 0
    \end{aligned}
\right.
$.
Finally, the feature transformation is the composition of three: $(h_\lambda \circ g_{w,\eta} \circ f)(\vz)$, parameterized by ${w,\eta,\lambda}$. If all features (for both training and testing set) go through this normalization function, then iVoro becomes iVoro-N.

% ==================== Contribution 2 ====================
\subsection{Divide and Conquer: Progressive Voronoi Diagrams for Class-Incremental Learning} \label{sec:ivoro-d}
As mentioned earlier, iVoro (and iVoro-N) treats all classes equally and separates them all by bisectors, regardless of at which phase they appear. However, for two classes $\gC_{\tau, k_1}, \gC_{\tau, k_2}$ appear in the same phase $\tau$, we can in fact draw better boundary by training a linear probing model parametrized by $\mW, \vb$ in the fixed feature space. After the training, the locating of new Voronoi center requires an explicit relationship between the probing model and VD. More formally, at phase $t$ a linear classifier with cross-entropy loss is optimized on the \emph{local} data $\gD_t$:
\begin{equation}\label{eq:linear}
    \gL(\mW_t, \vb_t) = \sum_{(\vx, y) \in \gD_t} -\log p(y|\phi(\vx);\mW_t,\vb_t) = \sum_{(\vx, y) \in \gD_t} -\log \frac{\exp(\mW_{t,y}^T \phi(\vx) + \evb_{t,y})}{\sum_k \exp(\mW_{t,k}^T \phi(\vx) + \evb_{t,k})}
\end{equation}
in which $\mW_{t,k}, \evb_{t,k}$ are the linear weight and bias for class $\gC_{t,k}$. As a parameterized model, this linear probing can ideally improve the discrimination within $\gC_t$. However, it is still non-trivial to merge all $\{\mW_{\tau,k}, \evb_{\tau,k}\}_{\tau=1}^{t}$, since the task identity is not assumed to be known like in TIL. To solve this, we get geometric insight from~\citep{ma2022fewshot} which directly connects linear probing model and VD by the theorem shown as follows:
\begin{theorem}[Voronoi Diagram Reduction~\citep{ma2022fewshot}] \label{thm:thm}
    The linear classifier parameterized by $\mW, \vb$ partitions the input space $\R^n$ to a Voronoi Diagram with centers $\{ \tilde{\vc}_1,...,\tilde{\vc}_K \}$ given by $\tilde{\vc}_k = \frac{1}{2} \mW_k$ if $\evb_k = -\frac{1}{4} ||\mW_k||_2^2, k = 1,...,K$.
\end{theorem}
%\begin{proof}
%    See Appendix \ref{supp:power} for details.
%\end{proof}
For completeness, we also include the proof in Appendix~\ref{supp:power}. During linear probing, if Thm.~\ref{thm:thm} is satisfied, then it is guaranteed that the resulting centers $\{\tilde{\vc}_{t, k}\}_{k=1}^{K_t}$ will also induces a VD (locally in phase $t$). Now given that we have two sets of centers $\{\vc_{\tau, k}\}$ and $\{\tilde{\vc}_{\tau, k}\}$, with the latter being better locally but are not transferable across phases, we devise a divide-and-conquer (D\&C) algorithm that progressively construct the decision boundaries from the two sets of centers, boosting iVoro to iVoro-D.

\noindent\textbf{Divide.} Fortunately, the total classes $\{\gC_\tau\}_{\tau = 1}^{t}$ have been split into already disjoint $t$ cliques. %\mg{In section 2.1, the phases are t=1:T, is it better to make it consistent? <added to 2.1: T for total phase e.g. 20, t for current phase e.g. 4, and tao for historical phase e.g. 0-3.>}

\noindent\textbf{Conquer.} Within each clique (i.e. phase) $\tau$, the boundary for any two classes $\gC_{\tau,k_1}, \gC_{\tau,k_2}$ is the bisector separating the probing-induced centers $\tilde{\vc}_{\tau,k_1}, \tilde{\vc}_{\tau,k_2}$, denoted as $\Gamma_{\tau, k_1, \tau, k_2} = \{\vz \in \R^n | \vv^T \vz' - q = 0 \}$ where $\vv = \frac{\tilde{\vc}_{\tau,k_1} - \tilde{\vc}_{\tau,k_2}}{||\tilde{\vc}_{\tau,k_1} - \tilde{\vc}_{\tau,k_2}||_2}$ and $q = \frac{||\tilde{\vc}_{\tau,k_1}||_2^2 - ||\tilde{\vc}_{\tau,k_2}||_2^2}{2||\tilde{\vc}_{\tau,k_1} - \tilde{\vc}_{\tau,k_2}||_2}$. When merging cliques $\tau_1, \tau_2$, we instead resort to the prototypes for space partition: for any $\vc_{\tau_1,k}$ in clique $\tau_1$ and any $\vc_{\tau_2,k'}$ in clique $\tau_2$, their bisector is $\Gamma_{\tau_1, k, \tau_2, k'} = \{\vz \in \R^n | \vv^T \vz' - q = 0 \}$ where $\vv = \frac{\vc_{\tau_1,k} - \vc_{\tau_2,k'}}{||\vc_{\tau_1,k} - \vc_{\tau_2,k'}||_2}$ and $q = \frac{||\vc_{\tau_1,k}||_2^2 - ||\vc_{\tau_2,k'}||_2^2}{2||\vc_{\tau_1,k} - \vc_{\tau_2,k'}||_2}$. In this way, the overall space partition would benefit from both locally probing-induced VD and globally prototype-based VD. See Appendix~\ref{supp:alg} for the time complexity of iVoro-D in details. %\mg{The formulas are the same for the two cases? If yes, can we merge that to improve readability? <different>} \mg{What is the complexity of this part? For CIFAR100 with 100 classes? Do you need $100^2 $ bisector? <moved to appendix>}

\noindent\textbf{Querying the VD.} In test-time, one can find the assigned Voronoi cell for query example $\vx$ by eliminating one class in each round according to $\mathrm{sign}(\vv^T \vz' - q)$, starting from a randomly selected boundary, so the time complexity is $\mathcal{O}({\textstyle\sum}_{\tau=1}^{t}K_\tau)$. 

% ==================== Contribution 3 ====================
\subsection{Voronoi Residual Prototypical Networks: Seeking for Better Prototypes} \label{sec:ivoro-r}
In our geometric modeling of the CIL problem, we first collect all the prototypes that appear so far: $\{\vc_{\tau, k}\}_{\tau \in \{1,...,t\},k \in \{1,...,K_{\tau}\}}$, but do not make use of at which phase that they come (iVoro). Next, iVoro-D divides this collection into cliques according to their time stamp $\tau$ and uses probing-induced centers $\{\tilde{\vc}_{\tau, k}\}_{k \in \{1,...,K_{\tau}\}}$ for the space partition within classes $\gC_\tau$ at phase $t$. However, one may notice that the cross-phase decision boundaries are still determined by bisecting the vanilla prototypes (feature means), which are not necessarily optimal. Most existing methods, e.g. PASS~\citep{zhu2021prototype}, uses classifier and prototypes in parallel and stores both. But with the direct connection revealed by Thm~\ref{thm:thm}, it becomes possible to unify the prototypical centers and the classifier-induced centers. Here, we show that VD could assist the optimization of the classifier $\theta$ by setting the prototype as a well-educated initialization for the classifier. Specifically, at each phase $t$, the prototypes $\{\vc_{t, k}\}$ are firstly calculated and then assigned to the classifier, following~\ref{thm:thm}, i.e.:
\[\mW_{t,k}^{(0)} = 2\vc_{t, k}, \evb_{t,k}^{(0)} = -\sfrac{1}{4} ||\mW_{t,k}^{(0)}||_2^2.\]
Meanwhile, during the optimization guided by Eq.~\ref{eq:linear}, we do not want $\mW_{t,k}$ moving too far away from $\mW_{t,k}^{(0)}$. To do so, we let the probing parameters (now denoted as $\Delta \mW_{t,k}, \Delta \vb_{t,k}$) indicate the moving from $\mW_{t,k}^{(0)}$, and the probing loss is retrofitted to: 
\begin{equation}\label{eq:residue}
    \tilde{\gL}(\Delta \mW_t, \Delta \vb_t) = \gL(\mW_t + \Delta \mW_t, \vb_t + \Delta \vb_t) + \beta ||\Delta \mW_t||_2^2
\end{equation}
in which $\beta$ is the hyper-parameter for weight decay, controlling the magnitude of the movement. Since here the optimization is only applied to the "residue" from Voronoi center to the original prototypes, we call the method Voronoi residual prototypical network (iVoro-R), and the resulting Voronoi centers are $\{\tilde{\vc}_{\tau, k}'\}_{k \in \{1,...,K_{\tau}\}}$ for phase $\tau$. %\mg{phase t or $\tau$? <added above: T total, t current, and tao historical>}

When used in conjunction with divide-and-conquer, the method becomes iVoro-DR, in which the within-phase boundaries are drew as before, but the cross-phase boundaries are determined by $\{\tilde{\vc}_{\tau, k}'\}$.

% ==================== Contribution 4 ====================
\subsection{Augmentation Integration: Uncertainty-aware Test-time Voronoi Cell Assignment} \label{sec:ivoro-a}
\noindent\textbf{Self-supervised Label Augmentation.} To enhance the discriminative power of CIL method, SSL-based label augmentation has been used to expand the original $K_t$ classes to $4K_t$ by rotating the original image $\vx$. Specifically, for image $\vx$, the rotated image $\vx^{(\alpha)} = rotate(\vx, \frac{\pi}{2}\alpha), \alpha \in \{0, 1, 2, 3\}$ will be assigned to a new class $y = k^{(\alpha)}, k \in \{1,...,K\}, K \in {\textstyle\sum}_{\tau = 1}^{t} K_{\tau}$. In training time, the model is trained upon the expanded dataset; however, in testing time, each of the duplicated images $\{\vx^{(\alpha)}\}_{\alpha \in \{0, 1, 2, 3\}}$ could possibly be assigned to each of the expanded classes $\{k^{(\alpha)}\}_{\alpha \in \{0, 1, 2, 3\}}$, so this ambiguity has to be resolved.

\noindent\textbf{Augmentation Consensus.} Let $\vd^{(\alpha, \alpha)} \in \R^K$ be a vector, each component of which denotes the distance from $\phi(\vx)$ to a class that $\phi$ has learned, i.e. $\evd_k^{(\alpha, \alpha)} = d(\phi(\vx^{(\alpha)}), \vc_{k^{(\alpha)}}) = ||\phi(\vx^{(\alpha)}) - \vc_{k^{(\alpha)}}||_2^2, k \in \{1,...,K\}, \alpha \in \{0, 1, 2, 3\}$. Then we want to find a consensus $\hat{k}$, with the maximum occurrence among the $\alpha \times \alpha$ predictions $\{\argmin_k \evd_k^{(\alpha, \alpha)}\}_{\alpha,\alpha \in \{0,1,2,3\}}$. Using augmentation consensus in test-time, then iVoro is then denoted as iVoro-AC.

\noindent\textbf{Augmentation Integration.} Using the consensus from the augmented samples should be more robust than the individual prediction $\argmin_k \evd_k^{(0, 0)}$ itself, but it has not considered the accumulated distance, so alternatively, we propose to integral over all predictions from augmented samples:
\[\hat{k} = {\textstyle\argmin}_k {\textstyle\sum}_{\alpha} {\textstyle\sum}_{\alpha} \evd_k^{(\alpha, \alpha)}.\]
If augmentation integration is applied, then iVoro becomes iVoro-AI.

\noindent\textbf{Uncertainty Quantification.} Since in iVoro-AC and iVoro-AI, the augmented samples collaboratively contribute to the final prediction, the quantitative uncertainty becomes neglected, this is because for some rotation-invariant classes, e.g. balls, the rotation operation makes less sense. Hence, when assigning a query sample $\vx$ to the augmented $4\times$ Voronoi cells, an uncertainty quantification method is needed.

Truth Discovery Ensemble (TDE)~\citep{ma2021improving} is the state-of-the-art uncertainty calibration method for DNNs, which finds the consensus among ensemble members by the minimization of entropy-based geometric variance (HV). Here, we only borrow HV as an indicator for the uncertainty of the $\alpha \times \alpha$ predictions, and refer the readers to~\citep{ma2021improving} for more details about TDE. Given the mean vector of the augmented predictions $\vd^* = \frac{1}{\alpha^2} {\textstyle\sum}_{\alpha,\alpha} \vd^{(\alpha, \alpha)}\in \R^K$, let $V$ denote the total squared distance to $\vd^*$ (i.e., $V = \sum_{\alpha}\sum_{\alpha} ||\vd^* - \vd^{(\alpha, \alpha)}||^2$) and $q^{(\alpha, \alpha)}$ denotes the contribution of each $\vd^{(\alpha, \alpha)}$ to $V$ (i.e., $q^{(\alpha, \alpha)} = ||\vd^* - \vd^{(\alpha, \alpha)}||^2/V$). Then the entropy induced by $\{q^{(\alpha, \alpha)}\}$ is:
\[
  H = -{\textstyle\sum}_{\alpha}{\textstyle\sum}_{\alpha} q^{(\alpha, \alpha)} \log q^{(\alpha, \alpha)} = \sfrac{1}{V} {\textstyle\sum}_{\alpha}{\textstyle\sum}_{\alpha} ||\vd^* - \vd^{(\alpha, \alpha)}||^2 \log (\sfrac{V}{||\vd^* - \vd^{(\alpha, \alpha)}||^2}).
\]
Based on these, we can define the HV as follows:

\begin{definition}[Entropy-based Geometric Variance~\citep{ding2020learning}]
  Given the point set $\{\vd^{(\alpha, \alpha)}\} \subseteq \R^K$ and a point $\vd^*$, the entropy based geometric variance (HV) is $H \times V$ where $H$ and $V$ are defined as shown above. 
\end{definition}

For every query example $\vx$, we calculate $\textrm{HV}(\vx)$ based on its $\{\vd^{(\alpha, \alpha)}\}_{\alpha,\alpha \in \{0,1,2,3\}}$. Later we will show how HV could favorably indicate the uncertainty of the augmented prediction, and tell us when augmentation integration is useful.
\begin{figure}
    \centering
    \subfloat{\includegraphics[height=1.3in]{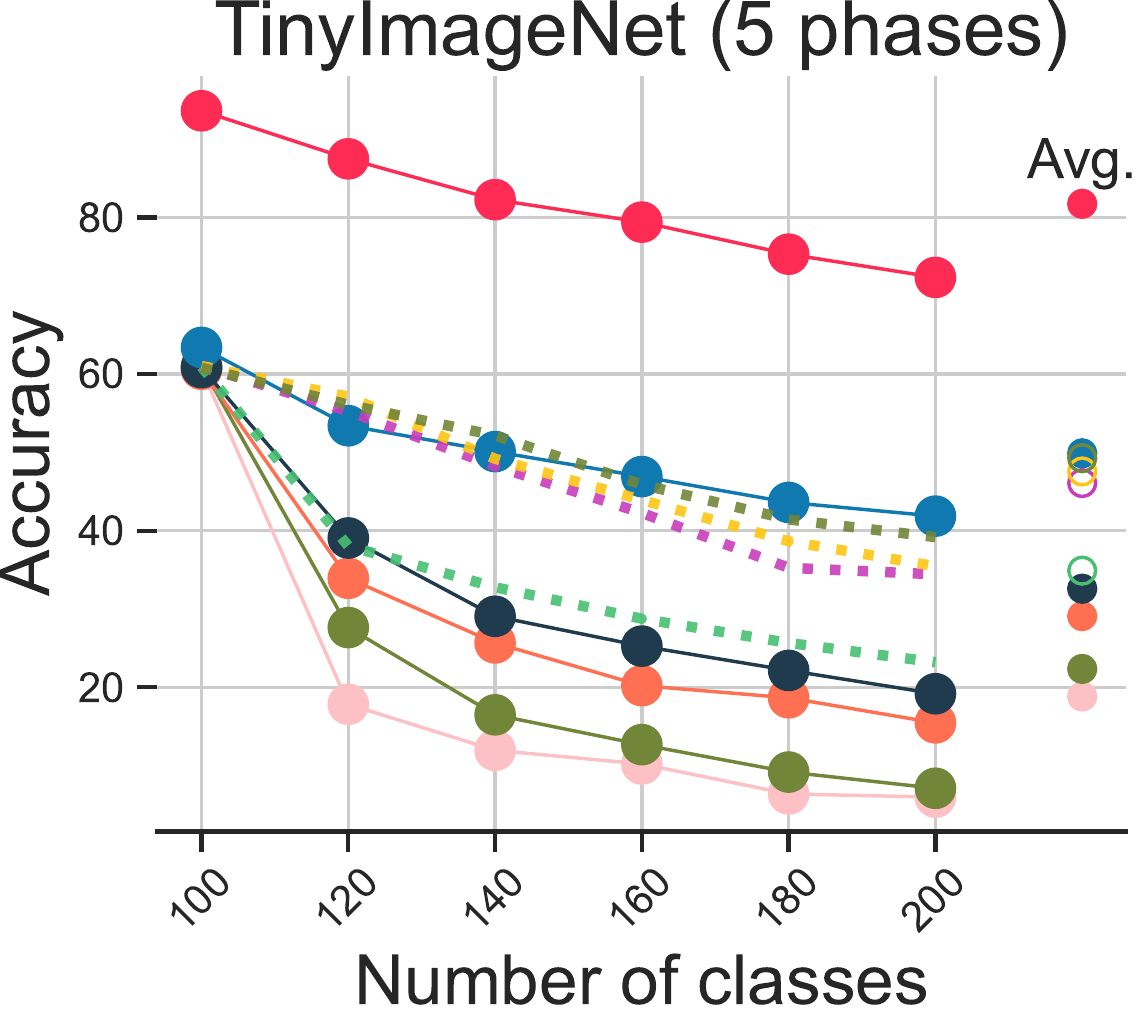}} \hspace{1mm}
    \subfloat{\includegraphics[height=1.3in]{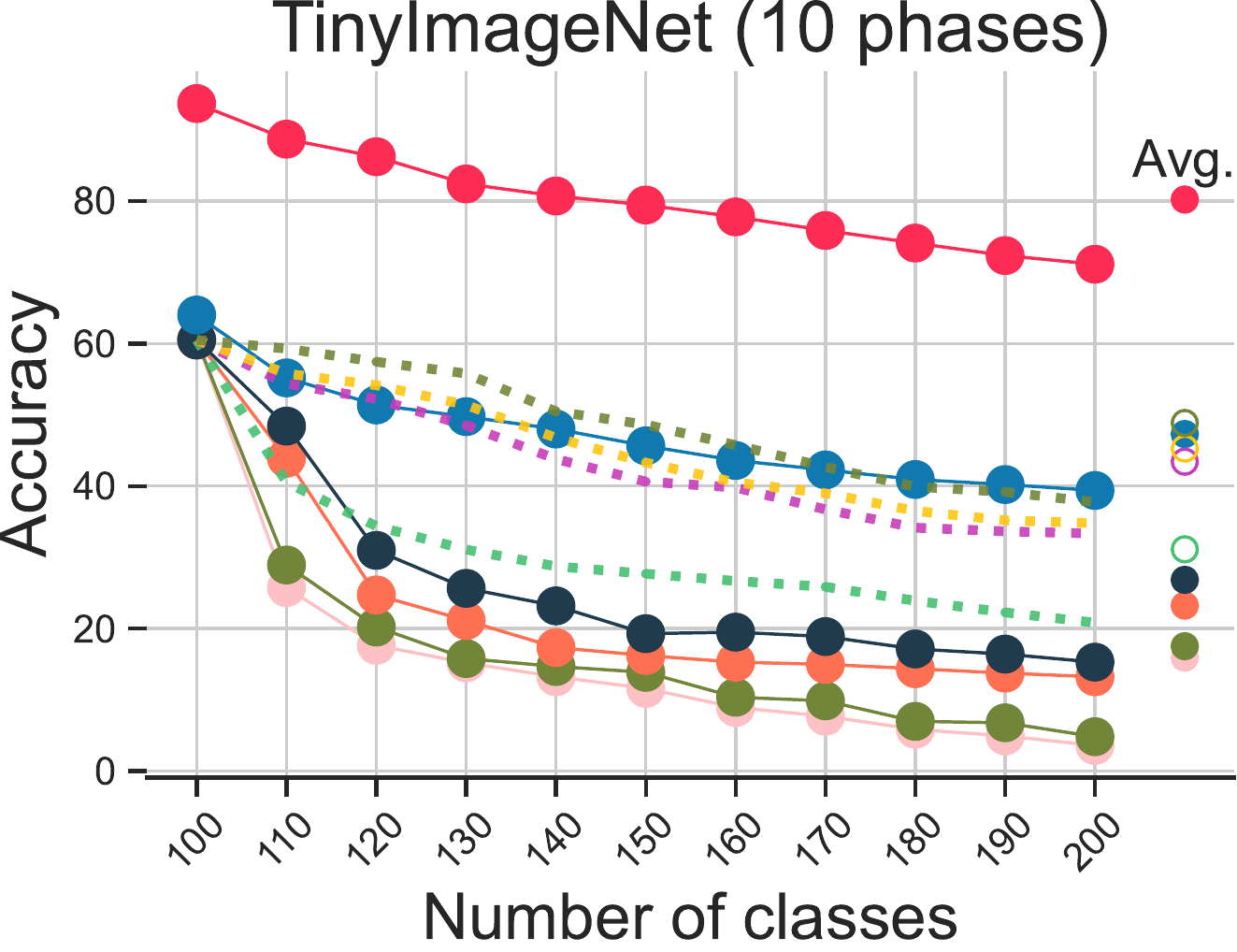}} \hspace{1mm}
    \subfloat{\includegraphics[height=1.3in]{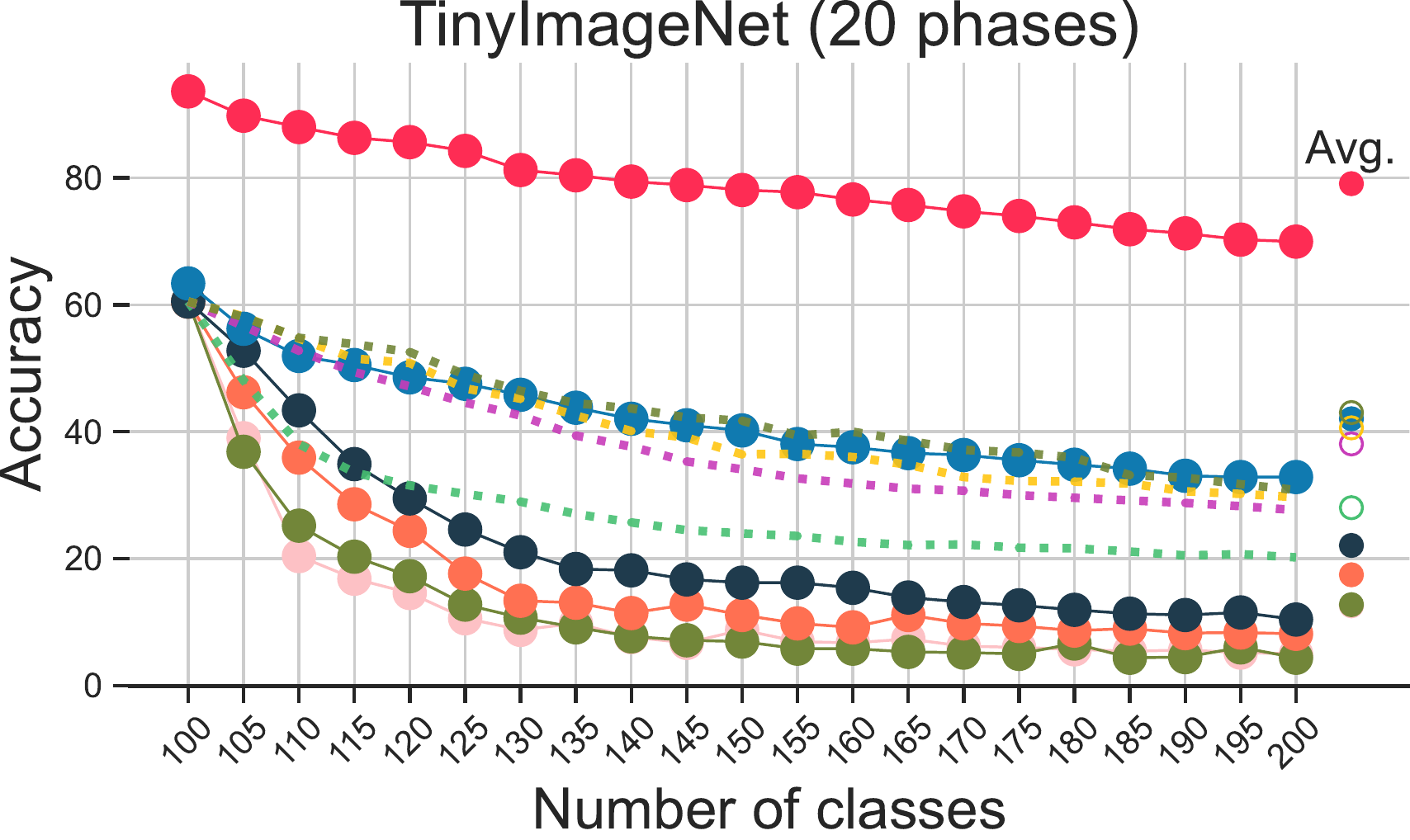}} \\ [-0.0ex]
    \subfloat{\includegraphics[width=0.7\linewidth]{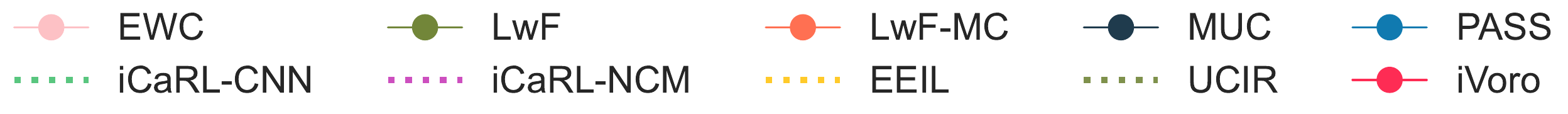}}
    \vspace{-2mm}
    \caption{Top-1 classification accuracy on TinyImagenet during 5/10/20 phases of CIL. See Appendix~\ref{supp:fig} for results on CIFAR-100 and ImageNet-Subset.)}\label{fig:tiny}
    \vspace{-3mm}
\end{figure}
% \begin{figure}
%     \centering
%     \subfloat{\includegraphics[height=1.3in]{figs/CIFAR100_5.pdf}} \hspace{1mm}
%     \subfloat{\includegraphics[height=1.3in]{figs/CIFAR100_10.pdf}} \hspace{1mm}
%     \subfloat{\includegraphics[height=1.3in]{figs/CIFAR100_20.pdf}} \\ [-0.0ex]
%     \subfloat{\includegraphics[width=0.7\linewidth]{figs/CIFAR100.pdf}}
%     \caption{Top-1 classification accuracy on CIFAR-100 during 5/10/20 phases of CIL.)}\label{fig:cifar}
% \end{figure}
% ==================== Contribution 5 ====================
\subsection{Layered Voronoi Diagrams as CIVD: Voronoi Diagram Meets Deep Neural Network} \label{sec:ivoro-l}
Until now, our VD construction is restricted to the deep feature space, i.e., $\vx \mapsto \phi(\vx) \in \R^n$. However, the intermediate layers are likely to contain information supplementary to the final layer and useful to VD construction, which will concern the integration of multiple VDs. Recently, Cluster-induced Voronoi Diagram (CIVD)~\citep{ChenHL017, HuangCX21} and Cluster-to-cluster Voronoi Diagram (CCVD)~\citep{ma2022fewshot}, two advanced VD structures, have shown remarkable ability to integrate multiple sets of centers for VD construction and achieve state-of-the-art performance in metric-based FSL. In this paper, we utilize the concept of CCVD for the integration of multiple VDs induced by multiple layers. Readers can refer to~\citep{ma2022fewshot} for more details about CIVD/CCVD.
\begin{definition}[Cluster-to-cluster Voronoi Diagram] \label{def:ccvd}
    Let $\Omega = \{\omega_1,...,\omega_K\}$ be a partition of the space $\R^n$, and $\gC = \{\gC_1,...,\gC_K\}$ be a set of \emph{totally ordered sets} with the same cardinality $L$ (i.e. $|\gC_1| = |\gC_2| = ... = |\gC_K| = L$). The set of pairs $\{ (\omega_1, \gC_1),..., (\omega_K, \gC_K)\}$ is a Cluster-to-cluster Voronoi Diagram (CCVD) with respect to an \emph{influence function} $F(\gC_k, \gC(\vz))$, and each cell is obtained via $\omega_r = \{ \vz \in \R^n : r(\vz) = r\}, r \in \{1,..,K\}$, with $r(\vz) = \argmax_{k \in \{ 1,...,K \}} F(\gC_k, \gC(\vz))$ where $\gC(\vz)$ is the cluster (also a totally ordered set with cardinality $L$) that query point $\vz$ belongs to, meaning that, all points in this cluster (query cluster) will be assigned to the same cell. The Influence Function is defined upon two totally ordered sets $\gC_k = \{ \vc_k^{(i)} \}_{i=1}^L$ and $\gC(\vz) = \{ \vz^{(i)} \}_{i=1}^L$: $F(\gC_k,\gC(\vz)) = - \sign(\gamma)\ {\textstyle\sum}_{i=0}^{L}\ d(\vc_k^{(i)},\vz^{(i)})^\gamma$.
\end{definition}
As CCVD is a flexible framework and can be applied to iVoro/iVoro-D/iVoro-R/iVoro-AC/iVoro-AI, here, as an example, we show how CCVD can be use to boost iVoro. In iVoro, the VD is induced by $\{\vc_{\tau, k}\}_{\tau \in \{1,...,t\},k \in \{1,...,K_{\tau}\}}$ that are feature means from the last layer $\phi$. Now, we arbitrarily extract $L$ layers $\{\phi^{(l)}\}_{l=1}^L$ and generate the $K$ totally ordered clusters $\{\{\vc_{\tau, k}^{(l)}\}_{l=1}^L\}_{\tau \in \{1,...,t\},k \in \{1,...,K_{\tau}\}}$ to construct CCVD and generate the query cluster $\{\phi^{(l)}(\vx)\}_{l=1}^L$ for the query example $\vx$ for Voronoi cell assignment. See Appendix \ref{supp:notation} for a summary of the notations and acronyms. % used throughout the Method section.

%Please add the following packages if necessary:
%\usepackage{booktabs, multirow} % for borders and merged ranges
%\usepackage{soul}% for underlines
%\usepackage[table]{xcolor} % for cell colors
%\usepackage{changepage,threeparttable} % for wide tables
%If the table is too wide, replace \begin{table}[!htp]...\end{table} with
%\begin{adjustwidth}{-2.5 cm}{-2.5 cm}\centering\begin{threeparttable}[!htb]...\end{threeparttable}\end{adjustwidth}
\setlength\intextsep{0pt} % change this value
\begin{table}[!tp]\centering
    \caption{Comparison between the fully-fledged iVoro with state-of-the-art non-exemplar (marked by \xmark) and exemplar-based (marked by \cmark) CIL methods in terms of the accuracy (in \%) in the last phase and the average accuracy (Avg., in \%) across all phases. \emph{imp.}$\uparrow$ indicates the relative improvement upon the next best \emph{non-exemplar} CIL method. \ddag The best version of iVoro is shown here. See Tab.~\ref{tab:abl} for different versions of iVoro. Note that RMM uses a 100-class subset of ImageNet that is different from others (shown in \textcolor{blue}{blue}). }\label{tab:main}
    \resizebox{\textwidth}{!}{%
    \Large
    \begin{tabular}{lrr|rr|rr|rr|rr|rr|rr}\toprule
    &\multicolumn{6}{c}{\textbf{CIFAR-100}} &\multicolumn{6}{c}{\textbf{TinyImageNet }} &\multicolumn{2}{c}{\textbf{ImageNet-Subset}} \\\cmidrule{2-15}
    &\multicolumn{2}{c}{5 phases} &\multicolumn{2}{c}{10 phases} &\multicolumn{2}{c}{20 phases} &\multicolumn{2}{c}{5 phases} &\multicolumn{2}{c}{10 phases} &\multicolumn{2}{c}{20 phases} &\multicolumn{2}{c}{10 phases} \\\cmidrule{2-15}
    Methods &Avg. &Last &Avg. &Last &Avg. &Last &Avg. &Last &Avg. &Last &Avg. &Last &Avg. &Last \\\midrule
    \cmark{ }iCaRL\textsubscript{CNN}~\citep{rebuffi2017icarl} &51.25 &40.50 &48.52 &39.13 &44.85 &34.38 &34.90 &23.20 &31.12 &20.82 &28.03 &20.20 &50.61 &38.40 \\
    \cmark{ }iCaRL\textsubscript{NCM}~\citep{rebuffi2017icarl} &58.13 &48.00 &53.91 &45.38 &50.79 &40.88 &46.08 &34.43 &43.42 &33.33 &38.08 &27.65 &60.89 &50.06 \\
    \cmark{ }EEIL~\citep{castro2018end} &60.15 &50.13 &55.91 &47.63 &52.79 &42.63 &47.56 &35.46 &45.26 &34.77 &40.61 &29.69 &63.40 &52.91 \\
    \cmark{ }UCIR~\citep{hou2019learning} &63.83 &54.75 &60.94 &50.75 &59.46 &47.00 &49.26 &39.18 &48.85 &37.74 &43.02 &30.82 &67.59 &55.89 \\
    \cmark{ }RMM~\citep{liu2021rmm} &68.86 &59.00 &67.61 &59.03 &$-$ &$-$ &$-$ &$-$ &$-$ &$-$ &$-$ &$-$ &\textcolor{blue}{78.47} &\textcolor{blue}{71.40} \\
    \midrule
    \xmark{ }EWC~\citep{kirkpatrick2017overcoming} &24.23 &9.00 &21.15 &8.50 &16.26 &7.75 &18.83 &5.98 &15.90 &3.59 &12.57 &5.00 &20.26 &9.03 \\
    \xmark{ }LwF~\citep{li2017learning} &32.54 &14.25 &17.91 &5.88 &14.95 &5.50 &22.35 &7.11 &17.52 &4.82 &12.75 &4.39 &23.57 &11.54 \\
    \xmark{ }LwF-MC~\citep{li2017learning} &46.06 &33.38 &27.31 &15.75 &19.99 &11.88 &29.09 &15.46 &23.22 &13.23 &17.46 &8.16 &31.22 &20.69 \\
    \xmark{ }MUC~\citep{liu2020more} &49.56 &36.00 &32.35 &20.63 &22.68 &9.50 &32.59 &19.18 &26.83 &15.28 &22.08 &10.41 &35.03 &24.46 \\
    \xmark{ }PASS~\citep{zhu2021prototype} &63.88 &55.75 &60.07 &49.13 &58.21 &48.75 &49.88 &41.86 &47.30 &39.38 &42.04 &32.86 &62.26 &50.63 \\
    \midrule
    \cellcolor[HTML]{f2f2f2}\textbf{\xmark{ }iVoro (Best)\ddag} &\cellcolor[HTML]{f2f2f2}\textbf{83.57} &\cellcolor[HTML]{f2f2f2}\textbf{74.40} &\cellcolor[HTML]{f2f2f2}\textbf{83.52} &\cellcolor[HTML]{f2f2f2}\textbf{74.39} &\cellcolor[HTML]{f2f2f2}\textbf{81.24} &\cellcolor[HTML]{f2f2f2}\textbf{71.45} &\cellcolor[HTML]{f2f2f2}\textbf{81.74} &\cellcolor[HTML]{f2f2f2}\textbf{72.34} &\cellcolor[HTML]{f2f2f2}\textbf{80.22} &\cellcolor[HTML]{f2f2f2}\textbf{71.13} &\cellcolor[HTML]{f2f2f2}\textbf{79.08} &\cellcolor[HTML]{f2f2f2}\textbf{69.95} &\cellcolor[HTML]{f2f2f2}\textbf{90.04} &\cellcolor[HTML]{f2f2f2}\textbf{83.84} \\
    \cellcolor[HTML]{f2f2f2}\emph{imp.}$\uparrow$ &\cellcolor[HTML]{f2f2f2}\textcolor{red}{+19.69} &\cellcolor[HTML]{f2f2f2}\textcolor{red}{+18.65} &\cellcolor[HTML]{f2f2f2}\textcolor{red}{+23.45} &\cellcolor[HTML]{f2f2f2}\textcolor{red}{+25.26} &\cellcolor[HTML]{f2f2f2}\textcolor{red}{+23.03} &\cellcolor[HTML]{f2f2f2}\textcolor{red}{+22.70} &\cellcolor[HTML]{f2f2f2}\textcolor{red}{+31.86} &\cellcolor[HTML]{f2f2f2}\textcolor{red}{+30.48} &\cellcolor[HTML]{f2f2f2}\textcolor{red}{+32.92} &\cellcolor[HTML]{f2f2f2}\textcolor{red}{+31.75} &\cellcolor[HTML]{f2f2f2}\textcolor{red}{+37.04} &\cellcolor[HTML]{f2f2f2}\textcolor{red}{+37.09} &\cellcolor[HTML]{f2f2f2}\textcolor{red}{+27.78} &\cellcolor[HTML]{f2f2f2}\textcolor{red}{+33.21} \\
    \bottomrule
    \end{tabular}}
    \vspace{-3mm}
\end{table}

\section{Experiments} \label{sec:experiments}
In our geometric framework, starting from iVoro, the simplest prototype-induced VD model, we gradually add five components: <\textbf{1}> parameterized normalization (iVoro-N), <\textbf{2}> divide-and-conquer for progressive VD construction (iVoro-D), <\textbf{3}> Voronoi residual prototypes (iVoro-R), <\textbf{4}> augmentation consensus and integration (iVoro-AC and iVoro-AI), <\textbf{5}> multi-centered VD for multi-layer network (iVoro-L). In this section, our main goals are to: (1) validate the strength of every single component; (2) exhaust as many combinations of components as possible to see how different combinations collaboratively contribute to the overall result; and (3) investigate at which circumstances a method does or does not work, by analyzing data size, number of layers, and quantitative uncertainty.
%Please add the following packages if necessary:
%\usepackage{booktabs, multirow} % for borders and merged ranges
%\usepackage{soul}% for underlines
%\usepackage[table]{xcolor} % for cell colors
%\usepackage{changepage,threeparttable} % for wide tables
%If the table is too wide, replace \begin{table}[!htp]...\end{table} with
%\begin{adjustwidth}{-2.5 cm}{-2.5 cm}\centering\begin{threeparttable}[!htb]...\end{threeparttable}\end{adjustwidth}
\setlength\intextsep{0pt} % change this value
\begin{table}[!tp]\centering
    \caption{Comprehensive ablation experiments by testing with different combinations of 6 components: parameterized feature normalization ($\bigstar$), progressive VD via D\&C ($\spadesuit$), Voronoi residual prototypes ($\blacktriangle$), augmentation consensus ($\clubsuit$) or integration ($\blacklozenge$), and layered VD for multiple feature spaces ($\blacktriangledown$). The solid symbol ($\bigstar \spadesuit \blacktriangle \clubsuit \blacklozenge \blacktriangledown$) indicates a corresponding component is applied while the grayed symbol ($\textcolor{gray}{\bigstar \spadesuit \blacktriangle \clubsuit \blacklozenge \blacktriangledown}$) means the component is ablated. To clearly show the improvement contributed to by a certain component, the colored numbers represent the relative improvements compared to a certain row indicated by the triangle (\tikztriangleright[red,fill=red]\tikztriangleright[teal,fill=teal]\tikztriangleright[orange,fill=orange]\tikztriangleright[OliveGreen,fill=OliveGreen]\tikztriangleright[blue,fill=blue]\tikztriangleright[cyan,fill=cyan]\tikztriangleright[violet,fill=violet]\tikztriangleright[brown,fill=brown]\tikztriangleright[magenta,fill=magenta]\tikztriangleright[YellowOrange,fill=YellowOrange]) with the same color.}\label{tab:abl}
    \resizebox{\textwidth}{!}{%
    \LARGE
    \begin{tabular}{lrr|rr|rr|rr|rr|rr|rr}\toprule
    &\multicolumn{6}{c}{\textbf{CIFAR-100}} &\multicolumn{6}{c}{\textbf{TinyImageNet }} &\multicolumn{2}{c}{\textbf{ImageNet-Subset}} \\\cmidrule{2-15}
    &\multicolumn{2}{c}{5 phases} &\multicolumn{2}{c}{10 phases} &\multicolumn{2}{c}{20 phases} &\multicolumn{2}{c}{5 phases} &\multicolumn{2}{c}{10 phases} &\multicolumn{2}{c}{20 phases} &\multicolumn{2}{c}{10 phases} \\\cmidrule{2-15}
    Methods &Avg. &Last &Avg. &Last &Avg. &Last &Avg. &Last &Avg. &Last &Avg. &Last &Avg. &Last \\\midrule
    \tikztriangleright[red,fill=red] iVoro &66.39 &56.05 &66.09 &56.05 &62.33 &52.38 &45.12 &38.27 &45.09 &38.29 &45.04 &38.29 &66.50 &55.40 \\
    \midrule
    \tikztriangleright[teal,fill=teal] \cellcolor[HTML]{ffffff}iVoro-N &\cellcolor[HTML]{ffffff}66.80 &\cellcolor[HTML]{ffffff}56.39 &\cellcolor[HTML]{ffffff}66.51 &\cellcolor[HTML]{ffffff}56.40 &\cellcolor[HTML]{ffffff}63.50 &\cellcolor[HTML]{ffffff}54.03 &\cellcolor[HTML]{ffffff}47.40 &\cellcolor[HTML]{ffffff}40.48 &\cellcolor[HTML]{ffffff}47.35 &\cellcolor[HTML]{ffffff}40.48 &\cellcolor[HTML]{ffffff}47.29 &\cellcolor[HTML]{ffffff}40.48 &\cellcolor[HTML]{ffffff}68.19 &\cellcolor[HTML]{ffffff}57.80 \\
    $\bigstar \textcolor{gray}{\spadesuit \blacktriangle \clubsuit \blacklozenge \blacktriangledown}$ \cellcolor[HTML]{ffffff} &\cellcolor[HTML]{ffffff}\textcolor{red}{+0.42} &\cellcolor[HTML]{ffffff}\textcolor{red}{+0.34} &\cellcolor[HTML]{ffffff}\textcolor{red}{+0.43} &\cellcolor[HTML]{ffffff}\textcolor{red}{+0.35} &\cellcolor[HTML]{ffffff}\textcolor{red}{+1.17} &\cellcolor[HTML]{ffffff}\textcolor{red}{+1.65} &\cellcolor[HTML]{ffffff}\textcolor{red}{+2.29} &\cellcolor[HTML]{ffffff}\textcolor{red}{+2.21} &\cellcolor[HTML]{ffffff}\textcolor{red}{+2.26} &\cellcolor[HTML]{ffffff}\textcolor{red}{+2.19} &\cellcolor[HTML]{ffffff}\textcolor{red}{+2.25} &\cellcolor[HTML]{ffffff}\textcolor{red}{+2.19} &\cellcolor[HTML]{ffffff}\textcolor{red}{+1.69} &\cellcolor[HTML]{ffffff}\textcolor{red}{+2.40} \\
    \midrule
    \tikztriangleright[white,fill=white] iVoro-D &67.24 &56.94 &66.56 &56.53 &63.84 &53.34 &50.46 &41.71 &49.05 &40.53 &48.33 &40.04 &67.24 &56.12 \\
    $\textcolor{gray}{\bigstar} \spadesuit \textcolor{gray}{\blacktriangle \clubsuit \blacklozenge \blacktriangledown}$ &\textcolor{red}{+0.85} &\textcolor{red}{+0.89} &\textcolor{red}{+0.47} &\textcolor{red}{+0.48} &\textcolor{red}{+1.51} &\textcolor{red}{+0.96} &\textcolor{red}{+5.34} &\textcolor{red}{+3.44} &\textcolor{red}{+3.96} &\textcolor{red}{+2.24} &\textcolor{red}{+3.28} &\textcolor{red}{+1.75} &\textcolor{red}{+0.74} &\textcolor{red}{+0.72} \\
    \midrule
    \tikztriangleright[white,fill=white] \cellcolor[HTML]{ffffff}iVoro-ND &\cellcolor[HTML]{ffffff}67.55 &\cellcolor[HTML]{ffffff}57.25 &\cellcolor[HTML]{ffffff}66.89 &\cellcolor[HTML]{ffffff}56.75 &\cellcolor[HTML]{ffffff}64.65 &\cellcolor[HTML]{ffffff}54.72 &\cellcolor[HTML]{ffffff}51.83 &\cellcolor[HTML]{ffffff}43.43 &\cellcolor[HTML]{ffffff}50.71 &\cellcolor[HTML]{ffffff}42.48 &\cellcolor[HTML]{ffffff}50.17 &\cellcolor[HTML]{ffffff}42.10 &\cellcolor[HTML]{ffffff}69.07 &\cellcolor[HTML]{ffffff}58.52 \\
    $\bigstar \spadesuit \textcolor{gray}{\blacktriangle \clubsuit \blacklozenge \blacktriangledown}$ \cellcolor[HTML]{ffffff} &\cellcolor[HTML]{ffffff}\textcolor{teal}{+0.74} &\cellcolor[HTML]{ffffff}\textcolor{teal}{+0.86} &\cellcolor[HTML]{ffffff}\textcolor{teal}{+0.38} &\cellcolor[HTML]{ffffff}\textcolor{teal}{+0.35} &\cellcolor[HTML]{ffffff}\textcolor{teal}{+1.15} &\cellcolor[HTML]{ffffff}\textcolor{teal}{+0.69} &\cellcolor[HTML]{ffffff}\textcolor{teal}{+4.42} &\cellcolor[HTML]{ffffff}\textcolor{teal}{+2.95} &\cellcolor[HTML]{ffffff}\textcolor{teal}{+3.36} &\cellcolor[HTML]{ffffff}\textcolor{teal}{+2.00} &\cellcolor[HTML]{ffffff}\textcolor{teal}{+2.88} &\cellcolor[HTML]{ffffff}\textcolor{teal}{+1.62} &\cellcolor[HTML]{ffffff}\textcolor{teal}{+0.88} &\cellcolor[HTML]{ffffff}\textcolor{teal}{+0.72} \\
    \midrule
    \tikztriangleright[orange,fill=orange] iVoro-R &66.56 &56.08 &66.25 &56.03 &63.31 &53.09 &48.50 &40.16 &46.78 &39.51 &46.39 &39.31 &66.91 &55.90 \\
    $\textcolor{gray}{\bigstar \spadesuit} \blacktriangle \textcolor{gray}{\clubsuit \blacklozenge \blacktriangledown}$ &\textcolor{red}{+0.17} &\textcolor{red}{+0.03} &\textcolor{red}{+0.17} &\textcolor{red}{+0.02} &\textcolor{red}{+0.98} &\textcolor{red}{+0.71} &\textcolor{red}{+3.38} &\textcolor{red}{+1.89} &\textcolor{red}{+1.69} &\textcolor{red}{+1.22} &\textcolor{red}{+1.35} &\textcolor{red}{+1.02} &\textcolor{red}{+0.41} &\textcolor{red}{+0.50} \\
    \midrule
    \tikztriangleright[OliveGreen,fill=OliveGreen] \cellcolor[HTML]{ffffff}iVoro-DR &\cellcolor[HTML]{ffffff}67.06 &\cellcolor[HTML]{ffffff}56.76 &\cellcolor[HTML]{ffffff}66.45 &\cellcolor[HTML]{ffffff}56.34 &\cellcolor[HTML]{ffffff}64.12 &\cellcolor[HTML]{ffffff}53.66 &\cellcolor[HTML]{ffffff}51.39 &\cellcolor[HTML]{ffffff}42.80 &\cellcolor[HTML]{ffffff}49.50 &\cellcolor[HTML]{ffffff}40.98 &\cellcolor[HTML]{ffffff}48.67 &\cellcolor[HTML]{ffffff}40.41 &\cellcolor[HTML]{ffffff}67.66 &\cellcolor[HTML]{ffffff}56.62 \\
    $\textcolor{gray}{\bigstar} \spadesuit \blacktriangle \textcolor{gray}{\clubsuit \blacklozenge \blacktriangledown}$ \cellcolor[HTML]{ffffff} &\cellcolor[HTML]{ffffff}\textcolor{orange}{+0.50} &\cellcolor[HTML]{ffffff}\textcolor{orange}{+0.68} &\cellcolor[HTML]{ffffff}\textcolor{orange}{+0.20} &\cellcolor[HTML]{ffffff}\textcolor{orange}{+0.31} &\cellcolor[HTML]{ffffff}\textcolor{orange}{+0.80} &\cellcolor[HTML]{ffffff}\textcolor{orange}{+0.57} &\cellcolor[HTML]{ffffff}\textcolor{orange}{+2.89} &\cellcolor[HTML]{ffffff}\textcolor{orange}{+2.64} &\cellcolor[HTML]{ffffff}\textcolor{orange}{+2.72} &\cellcolor[HTML]{ffffff}\textcolor{orange}{+1.47} &\cellcolor[HTML]{ffffff}\textcolor{orange}{+2.28} &\cellcolor[HTML]{ffffff}\textcolor{orange}{+1.10} &\cellcolor[HTML]{ffffff}\textcolor{orange}{+0.75} &\cellcolor[HTML]{ffffff}\textcolor{orange}{+0.72} \\
    \midrule
    \tikztriangleright[white,fill=white] iVoro-NDR &67.50 &57.31 &66.80 &56.94 &64.91 &54.54 &52.62 &44.54 &50.69 &42.62 &49.84 &41.52 &68.72 &57.60 \\
    $\bigstar \spadesuit \blacktriangle \textcolor{gray}{\clubsuit \blacklozenge \blacktriangledown}$ &\textcolor{OliveGreen}{+0.44} &\textcolor{OliveGreen}{+0.55} &\textcolor{OliveGreen}{+0.35} &\textcolor{OliveGreen}{+0.60} &\textcolor{OliveGreen}{+0.79} &\textcolor{OliveGreen}{+0.88} &\textcolor{OliveGreen}{+1.23} &\textcolor{OliveGreen}{+1.74} &\textcolor{OliveGreen}{+1.19} &\textcolor{OliveGreen}{+1.64} &\textcolor{OliveGreen}{+1.17} &\textcolor{OliveGreen}{+1.11} &\textcolor{OliveGreen}{+1.06} &\textcolor{OliveGreen}{+0.98} \\
    \midrule
    \tikztriangleright[blue,fill=blue] \cellcolor[HTML]{ffffff}iVoro-AC &\cellcolor[HTML]{ffffff}81.38 &\cellcolor[HTML]{ffffff}70.63 &\cellcolor[HTML]{ffffff}81.25 &\cellcolor[HTML]{ffffff}70.63 &\cellcolor[HTML]{ffffff}78.16 &\cellcolor[HTML]{ffffff}66.14 &\cellcolor[HTML]{ffffff}64.01 &\cellcolor[HTML]{ffffff}55.26 &\cellcolor[HTML]{ffffff}64.01 &\cellcolor[HTML]{ffffff}55.28 &\cellcolor[HTML]{ffffff}64.00 &\cellcolor[HTML]{ffffff}55.29 &\cellcolor[HTML]{ffffff}83.41 &\cellcolor[HTML]{ffffff}71.90 \\
    $\textcolor{gray}{\bigstar \spadesuit \blacktriangle} \clubsuit \textcolor{gray}{\blacklozenge \blacktriangledown}$ \cellcolor[HTML]{ffffff} &\cellcolor[HTML]{ffffff}\textcolor{red}{+15.00} &\cellcolor[HTML]{ffffff}\textcolor{red}{+14.58} &\cellcolor[HTML]{ffffff}\textcolor{red}{+15.16} &\cellcolor[HTML]{ffffff}\textcolor{red}{+14.58} &\cellcolor[HTML]{ffffff}\textcolor{red}{+15.84} &\cellcolor[HTML]{ffffff}\textcolor{red}{+13.76} &\cellcolor[HTML]{ffffff}\textcolor{red}{+18.90} &\cellcolor[HTML]{ffffff}\textcolor{red}{+16.99} &\cellcolor[HTML]{ffffff}\textcolor{red}{+18.92} &\cellcolor[HTML]{ffffff}\textcolor{red}{+16.99} &\cellcolor[HTML]{ffffff}\textcolor{red}{+18.96} &\cellcolor[HTML]{ffffff}\textcolor{red}{+17.00} &\cellcolor[HTML]{ffffff}\textcolor{red}{+16.90} &\cellcolor[HTML]{ffffff}\textcolor{red}{+16.50} \\
    \midrule
    \tikztriangleright[cyan,fill=cyan] iVoro-AI &62.33 &50.18 &60.37 &50.18 &65.39 &59.11 &55.60 &48.11 &55.63 &48.12 &55.50 &48.11 &72.47 &60.66 \\
    $\textcolor{gray}{\bigstar \spadesuit \blacktriangle \clubsuit} \blacklozenge \textcolor{gray}{\blacktriangledown}$ &\textcolor{red}{-4.05} &\textcolor{red}{-5.87} &\textcolor{red}{-5.72} &\textcolor{red}{-5.87} &\textcolor{red}{+3.07} &\textcolor{red}{+6.73} &\textcolor{red}{+10.48} &\textcolor{red}{+9.84} &\textcolor{red}{+10.54} &\textcolor{red}{+9.83} &\textcolor{red}{+10.46} &\textcolor{red}{+9.82} &\textcolor{red}{+5.97} &\textcolor{red}{+5.26} \\
    \midrule
    \tikztriangleright[violet,fill=violet] \cellcolor[HTML]{ffffff}iVoro-NAC &\cellcolor[HTML]{ffffff}82.31 &\cellcolor[HTML]{ffffff}72.04 &\cellcolor[HTML]{ffffff}82.29 &\cellcolor[HTML]{ffffff}72.19 &\cellcolor[HTML]{ffffff}80.53 &\cellcolor[HTML]{ffffff}70.01 &\cellcolor[HTML]{ffffff}59.75 &\cellcolor[HTML]{ffffff}49.78 &\cellcolor[HTML]{ffffff}59.75 &\cellcolor[HTML]{ffffff}49.80 &\cellcolor[HTML]{ffffff}59.74 &\cellcolor[HTML]{ffffff}49.78 &\cellcolor[HTML]{ffffff}84.31 &\cellcolor[HTML]{ffffff}73.72 \\
    $\bigstar \textcolor{gray}{\spadesuit \blacktriangle} \clubsuit \textcolor{gray}{\blacklozenge \blacktriangledown}$ \cellcolor[HTML]{ffffff} &\cellcolor[HTML]{ffffff}\textcolor{blue}{+0.93} &\cellcolor[HTML]{ffffff}\textcolor{blue}{+1.41} &\cellcolor[HTML]{ffffff}\textcolor{blue}{+1.04} &\cellcolor[HTML]{ffffff}\textcolor{blue}{+1.56} &\cellcolor[HTML]{ffffff}\textcolor{blue}{+2.36} &\cellcolor[HTML]{ffffff}\textcolor{blue}{+3.87} &\cellcolor[HTML]{ffffff}\textcolor{blue}{-4.26} &\cellcolor[HTML]{ffffff}\textcolor{blue}{-5.48} &\cellcolor[HTML]{ffffff}\textcolor{blue}{-4.26} &\cellcolor[HTML]{ffffff}\textcolor{blue}{-5.48} &\cellcolor[HTML]{ffffff}\textcolor{blue}{-4.26} &\cellcolor[HTML]{ffffff}\textcolor{blue}{-5.51} &\cellcolor[HTML]{ffffff}\textcolor{blue}{+0.90} &\cellcolor[HTML]{ffffff}\textcolor{blue}{+1.82} \\
    \midrule
    \tikztriangleright[brown,fill=brown] iVoro-NAI &62.53 &50.42 &60.77 &50.62 &62.43 &56.54 &73.21 &65.84 &73.14 &65.90 &73.13 &65.83 &86.04 &77.12 \\
    $\bigstar \textcolor{gray}{\spadesuit \blacktriangle \clubsuit} \blacklozenge \textcolor{gray}{\blacktriangledown}$ &\textcolor{cyan}{+0.19} &\textcolor{cyan}{+0.24} &\textcolor{cyan}{+0.41} &\textcolor{cyan}{+0.44} &\textcolor{cyan}{-2.97} &\textcolor{cyan}{-2.57} &\textcolor{cyan}{+17.61} &\textcolor{cyan}{+17.73} &\textcolor{cyan}{+17.51} &\textcolor{cyan}{+17.78} &\textcolor{cyan}{+17.63} &\textcolor{cyan}{+17.72} &\textcolor{cyan}{+13.57} &\textcolor{cyan}{+16.46} \\
    \midrule
    \tikztriangleright[white,fill=white] \cellcolor[HTML]{ffffff}iVoro-NDAI &\cellcolor[HTML]{ffffff}69.00 &\cellcolor[HTML]{ffffff}56.35 &\cellcolor[HTML]{ffffff}63.76 &\cellcolor[HTML]{ffffff}52.87 &\cellcolor[HTML]{ffffff}63.94 &\cellcolor[HTML]{ffffff}57.52 &\cellcolor[HTML]{ffffff}81.00 &\cellcolor[HTML]{ffffff}71.29 &\cellcolor[HTML]{ffffff}79.64 &\cellcolor[HTML]{ffffff}70.10 &\cellcolor[HTML]{ffffff}78.17 &\cellcolor[HTML]{ffffff}68.70 &\cellcolor[HTML]{ffffff}86.92 &\cellcolor[HTML]{ffffff}78.64 \\
    $\bigstar \spadesuit \textcolor{gray}{\blacktriangle \clubsuit} \blacklozenge \textcolor{gray}{\blacktriangledown}$ \cellcolor[HTML]{ffffff} &\cellcolor[HTML]{ffffff}\textcolor{brown}{+6.47} &\cellcolor[HTML]{ffffff}\textcolor{brown}{+5.93} &\cellcolor[HTML]{ffffff}\textcolor{brown}{+2.99} &\cellcolor[HTML]{ffffff}\textcolor{brown}{+2.25} &\cellcolor[HTML]{ffffff}\textcolor{brown}{+1.51} &\cellcolor[HTML]{ffffff}\textcolor{brown}{+0.98} &\cellcolor[HTML]{ffffff}\textcolor{brown}{+7.79} &\cellcolor[HTML]{ffffff}\textcolor{brown}{+5.45} &\cellcolor[HTML]{ffffff}\textcolor{brown}{+6.50} &\cellcolor[HTML]{ffffff}\textcolor{brown}{+4.20} &\cellcolor[HTML]{ffffff}\textcolor{brown}{+5.04} &\cellcolor[HTML]{ffffff}\textcolor{brown}{+2.87} &\cellcolor[HTML]{ffffff}\textcolor{brown}{+0.89} &\cellcolor[HTML]{ffffff}\textcolor{brown}{+1.52} \\
    \midrule
    \tikztriangleright[white,fill=white] iVoro-RAC &81.00 &69.78 &81.26 &70.64 &79.68 &68.08 &66.45 &55.55 &65.97 &57.08 &66.28 &57.68 &83.32 &71.48 \\
    $\textcolor{gray}{\bigstar \spadesuit} \blacktriangle \clubsuit \textcolor{gray}{\blacklozenge \blacktriangledown}$ &\textcolor{blue}{-0.39} &\textcolor{blue}{-0.85} &\textcolor{blue}{+0.01} &\textcolor{blue}{+0.01} &\textcolor{blue}{+1.52} &\textcolor{blue}{+1.95} &\textcolor{blue}{+2.44} &\textcolor{blue}{+0.29} &\textcolor{blue}{+1.96} &\textcolor{blue}{+1.80} &\textcolor{blue}{+2.28} &\textcolor{blue}{+2.39} &\textcolor{blue}{-0.09} &\textcolor{blue}{-0.42} \\
    \midrule
    \tikztriangleright[magenta,fill=magenta] \cellcolor[HTML]{ffffff}iVoro-RAI &\cellcolor[HTML]{ffffff}62.95 &\cellcolor[HTML]{ffffff}50.99 &\cellcolor[HTML]{ffffff}60.04 &\cellcolor[HTML]{ffffff}49.49 &\cellcolor[HTML]{ffffff}63.69 &\cellcolor[HTML]{ffffff}57.58 &\cellcolor[HTML]{ffffff}58.63 &\cellcolor[HTML]{ffffff}50.01 &\cellcolor[HTML]{ffffff}58.96 &\cellcolor[HTML]{ffffff}50.40 &\cellcolor[HTML]{ffffff}58.00 &\cellcolor[HTML]{ffffff}49.40 &\cellcolor[HTML]{ffffff}73.59 &\cellcolor[HTML]{ffffff}61.90 \\
    $\textcolor{gray}{\bigstar \spadesuit} \blacktriangle \textcolor{gray}{\clubsuit} \blacklozenge \textcolor{gray}{\blacktriangledown}$ \cellcolor[HTML]{ffffff} &\cellcolor[HTML]{ffffff}\textcolor{cyan}{+0.61} &\cellcolor[HTML]{ffffff}\textcolor{cyan}{+0.81} &\cellcolor[HTML]{ffffff}\textcolor{cyan}{-0.32} &\cellcolor[HTML]{ffffff}\textcolor{cyan}{-0.69} &\cellcolor[HTML]{ffffff}\textcolor{cyan}{-1.70} &\cellcolor[HTML]{ffffff}\textcolor{cyan}{-1.53} &\cellcolor[HTML]{ffffff}\textcolor{cyan}{+3.03} &\cellcolor[HTML]{ffffff}\textcolor{cyan}{+1.90} &\cellcolor[HTML]{ffffff}\textcolor{cyan}{+3.34} &\cellcolor[HTML]{ffffff}\textcolor{cyan}{+2.28} &\cellcolor[HTML]{ffffff}\textcolor{cyan}{+2.50} &\cellcolor[HTML]{ffffff}\textcolor{cyan}{+1.29} &\cellcolor[HTML]{ffffff}\textcolor{cyan}{+1.12} &\cellcolor[HTML]{ffffff}\textcolor{cyan}{+1.24} \\
    \midrule
    \tikztriangleright[white,fill=white] iVoro-DRAI &68.91 &56.44 &62.43 &52.01 &64.84 &58.15 &66.90 &55.55 &67.52 &56.13 &62.86 &51.51 &75.36 &64.68 \\
    $\textcolor{gray}{\bigstar} \spadesuit \blacktriangle \textcolor{gray}{\clubsuit} \blacklozenge \textcolor{gray}{\blacktriangledown}$ &\textcolor{magenta}{+5.96} &\textcolor{magenta}{+5.45} &\textcolor{magenta}{+2.39} &\textcolor{magenta}{+2.52} &\textcolor{magenta}{+1.15} &\textcolor{magenta}{+0.57} &\textcolor{magenta}{+8.27} &\textcolor{magenta}{+5.54} &\textcolor{magenta}{+8.56} &\textcolor{magenta}{+5.73} &\textcolor{magenta}{+4.86} &\textcolor{magenta}{+2.11} &\textcolor{magenta}{+1.77} &\textcolor{magenta}{+2.78} \\
    \midrule
    \tikztriangleright[white,fill=white] \cellcolor[HTML]{ffffff}iVoro-NACL &\cellcolor[HTML]{ffffff}\textbf{83.57} &\cellcolor[HTML]{ffffff}\textbf{74.40} &\cellcolor[HTML]{ffffff}\textbf{83.52} &\cellcolor[HTML]{ffffff}\textbf{74.39} &\cellcolor[HTML]{ffffff}72.64 &\cellcolor[HTML]{ffffff}61.14 &\cellcolor[HTML]{ffffff}59.49 &\cellcolor[HTML]{ffffff}50.09 &\cellcolor[HTML]{ffffff}59.51 &\cellcolor[HTML]{ffffff}50.10 &\cellcolor[HTML]{ffffff}59.52 &\cellcolor[HTML]{ffffff}50.13 &\cellcolor[HTML]{ffffff}84.83 &\cellcolor[HTML]{ffffff}76.24 \\
    $\bigstar \textcolor{gray}{\spadesuit \blacktriangle} \clubsuit \textcolor{gray}{\blacklozenge} \blacktriangledown$ \cellcolor[HTML]{ffffff} &\cellcolor[HTML]{ffffff}\textcolor{violet}{+1.25} &\cellcolor[HTML]{ffffff}\textcolor{violet}{+2.36} &\cellcolor[HTML]{ffffff}\textcolor{violet}{+1.23} &\cellcolor[HTML]{ffffff}\textcolor{violet}{+2.20} &\cellcolor[HTML]{ffffff}\textcolor{violet}{-7.89} &\cellcolor[HTML]{ffffff}\textcolor{violet}{-8.87} &\cellcolor[HTML]{ffffff}\textcolor{violet}{-0.26} &\cellcolor[HTML]{ffffff}\textcolor{violet}{+0.31} &\cellcolor[HTML]{ffffff}\textcolor{violet}{-0.24} &\cellcolor[HTML]{ffffff}\textcolor{violet}{+0.30} &\cellcolor[HTML]{ffffff}\textcolor{violet}{-0.22} &\cellcolor[HTML]{ffffff}\textcolor{violet}{+0.35} &\cellcolor[HTML]{ffffff}\textcolor{violet}{+0.51} &\cellcolor[HTML]{ffffff}\textcolor{violet}{+2.52} \\
    \midrule
    \tikztriangleright[YellowOrange,fill=YellowOrange] iVoro-NAIL &71.63 &61.14 &69.87 &60.37 &78.19 &70.35 &74.42 &67.34 &74.35 &67.37 &74.39 &67.37 &89.38 &83.18 \\
    $\bigstar \textcolor{gray}{\spadesuit \blacktriangle \clubsuit} \blacklozenge \blacktriangledown$ &\textcolor{brown}{+9.10} &\textcolor{brown}{+10.72} &\textcolor{brown}{+9.10} &\textcolor{brown}{+9.75} &\textcolor{brown}{+15.76} &\textcolor{brown}{+13.81} &\textcolor{brown}{+1.21} &\textcolor{brown}{+1.50} &\textcolor{brown}{+1.22} &\textcolor{brown}{+1.47} &\textcolor{brown}{+1.26} &\textcolor{brown}{+1.54} &\textcolor{brown}{+3.34} &\textcolor{brown}{+6.06} \\
    \midrule
    \tikztriangleright[white,fill=white] \cellcolor[HTML]{ffffff}iVoro-NDAIL &\cellcolor[HTML]{ffffff}77.57 &\cellcolor[HTML]{ffffff}66.54 &\cellcolor[HTML]{ffffff}72.50 &\cellcolor[HTML]{ffffff}62.28 &\cellcolor[HTML]{ffffff}\textbf{81.24} &\cellcolor[HTML]{ffffff}\textbf{71.45} &\cellcolor[HTML]{ffffff}\textbf{81.74} &\cellcolor[HTML]{ffffff}\textbf{72.34} &\cellcolor[HTML]{ffffff}\textbf{80.22} &\cellcolor[HTML]{ffffff}\textbf{71.13} &\cellcolor[HTML]{ffffff}\textbf{79.08} &\cellcolor[HTML]{ffffff}\textbf{69.95} &\cellcolor[HTML]{ffffff}\textbf{90.04} &\cellcolor[HTML]{ffffff}\textbf{83.84} \\
    $\bigstar \spadesuit \textcolor{gray}{\blacktriangle \clubsuit} \blacklozenge \blacktriangledown$ \cellcolor[HTML]{ffffff} &\cellcolor[HTML]{ffffff}\textcolor{YellowOrange}{+5.95} &\cellcolor[HTML]{ffffff}\textcolor{YellowOrange}{+5.40} &\cellcolor[HTML]{ffffff}\textcolor{YellowOrange}{+2.63} &\cellcolor[HTML]{ffffff}\textcolor{YellowOrange}{+1.91} &\cellcolor[HTML]{ffffff}\textcolor{YellowOrange}{+3.05} &\cellcolor[HTML]{ffffff}\textcolor{YellowOrange}{+1.10} &\cellcolor[HTML]{ffffff}\textcolor{YellowOrange}{+7.31} &\cellcolor[HTML]{ffffff}\textcolor{YellowOrange}{+5.00} &\cellcolor[HTML]{ffffff}\textcolor{YellowOrange}{+5.86} &\cellcolor[HTML]{ffffff}\textcolor{YellowOrange}{+3.76} &\cellcolor[HTML]{ffffff}\textcolor{YellowOrange}{+4.69} &\cellcolor[HTML]{ffffff}\textcolor{YellowOrange}{+2.58} &\cellcolor[HTML]{ffffff}\textcolor{YellowOrange}{+0.66} &\cellcolor[HTML]{ffffff}\textcolor{YellowOrange}{+0.66} \\
    \bottomrule
    \end{tabular}}
    \vspace{-3mm}
\end{table}

\noindent\textbf{Datasets, Benchmarks, and Implementation Details.} 
Three standard datasets, CIFAR-100~\citep{krizhevsky2009learning}, TinyImageNet~\citep{le2015tiny} and ImageNet-Subset~\citep{deng2009imagenet} for CIL are used for method evaluation. We follow the popular benchmarking protocol used by~\citep{liu2021rmm,zhu2021prototype,douillard2020podnet,hou2019learning} in which the inital phase contains a half of the classes while the subsequent phases each has $\frac{1}{5}$ of the remaining classes ($\frac{1}{10}$ or $\frac{1}{20}$ for longer sequence of tasks). We mainly compare our method to non-exemplar methods including EWC~\citep{kirkpatrick2017overcoming}, LwF~\citep{li2017learning}, LwF-MC~\citep{li2017learning}, LwM~\citep{dhar2019learning}, and MUC~\citep{liu2020more}, but we also compare with several recent exemplar-based methods iCaRL~\citep{rebuffi2017icarl}, EEIL~\citep{castro2018end}, UCIR~\citep{hou2019learning}, and RMM~\citep{liu2021rmm} for reference. A ResNet-18~\citep{he2016deep} model is used for all experiments. We follow PASS~\citep{zhu2021prototype} to train the feature extractor on the first phase data which is frozen afterwards for all subsequent phases. All classes are expanded via rotating the original image by 90\textdegree, 180\textdegree, and 270\textdegree. See Appendix~\ref{supp:impl} for more details about the implementations of all the 18 ablation methods in Tab.~\ref{tab:abl}

\noindent\textbf{iVoro: Simple VD is A Strong Baseline.} Surprisingly, by only using prototypes for VD construction, our baseline method iVoro can achieve competitive performance for short phases and much better results for long phases, compared to the state-of-the-art non-exemplar CIL method. For example, the difference in accuracy in comparison to PASS is 0.29\%/6.91\%/3.63\% for 5/10/20-phase CIFAR-100, -3.58\%/-1.09\%/5.43\% for 5/10/20-phase TinyImageNet, and 4.76\% for ImageNet-Subset. We suspect that this is because the features generated by the frozen feature extractor can be satisfactorily separable by linear bisectors. To verify this, we use t-SNE to visualize the features (see~\ref{supp:tsne} for details and Fig.~\ref{fig:mnist-vd} for 2D visualization without t-SNE). As we can see, the features for other methods are all dramatically changing during the phases, but those for iVoro are all fixed, making incremental VD construction possible. Moreover, the accuracy of the last phase usually drops significantly with longer task sequence (e.g. 20 phases vs. 5 phases), but iVoro is highly robust at the last phase,
%able to maintain the accuracy for the last phase, 
because the final VDs are the same no matter how many phases it goes through. These results show that iVoro works favorably with long phases. When parameterized normalization is applied, iVoro-N further consistently improves upon iVoro by up to 2.40\% (10-phase ImageNet-Subset) (see Tab.~\ref{tab:abl}), by encouraging the compactness of feature distribution (Appendix~\ref{supp:tsne}). See Appendix~\ref{supp:norm} about the detailed analysis of iVoro-N.

\noindent\textbf{Normalization, D\&C, Residues: Synergistic Effects.} These three components can individually improve iVoro, but they also have collective impacts. For example, as shown in Tab.~\ref{tab:abl}, iVoro-N/iVoro-D/iVoro-R > iVoro, iVoro-ND > iVoro-N, and iVoro-NDR > iVoro-DR > iVoro-R, corroborating that every single contribution is useful and necessary for a better result.

\begin{figure}
    \centering
    \subfloat{\includegraphics[height=1.15in]{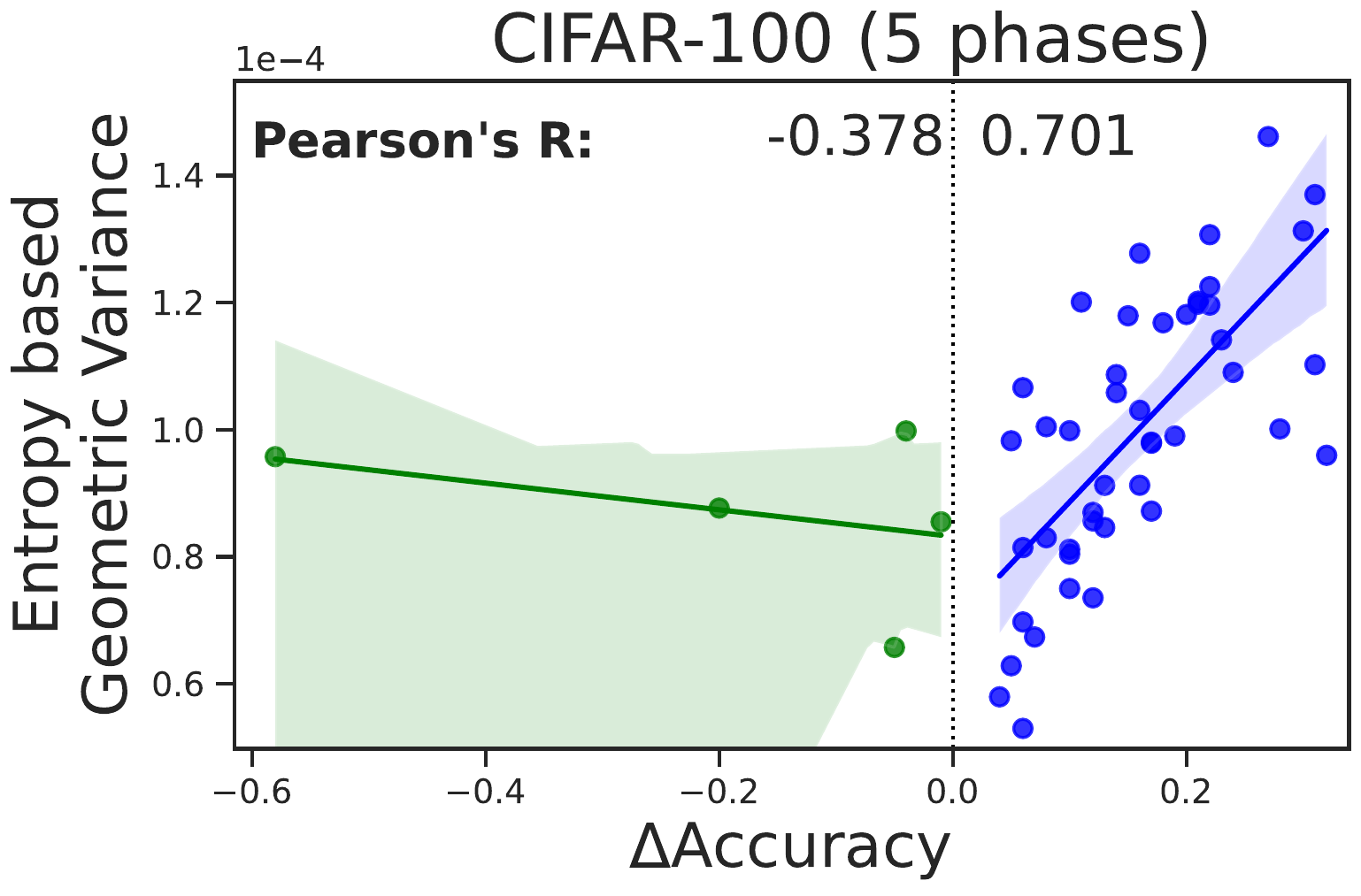}} \hspace{0.5mm}
    \subfloat{\includegraphics[height=1.15in]{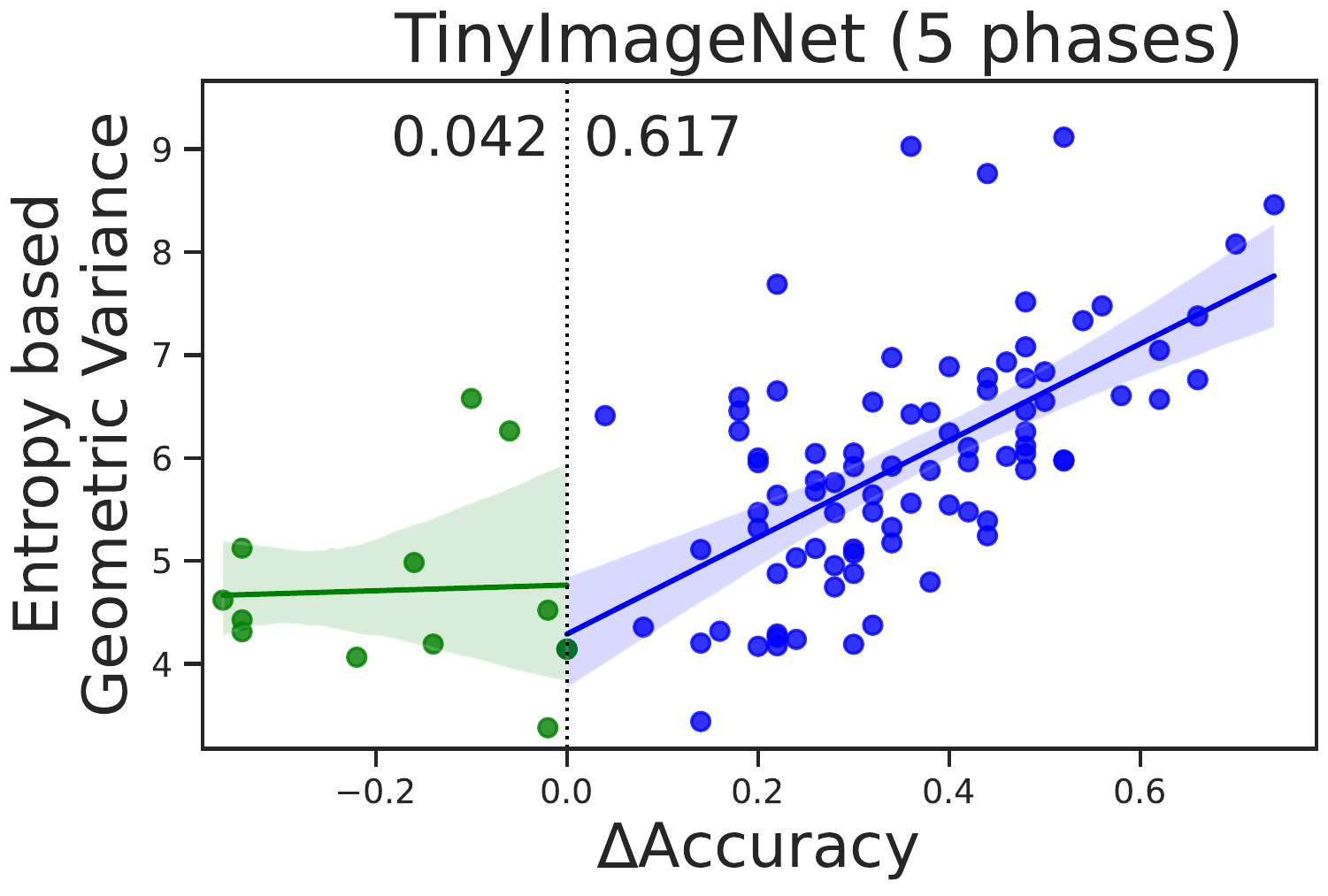}} \hspace{0.5mm}
    \subfloat{\includegraphics[height=1.15in]{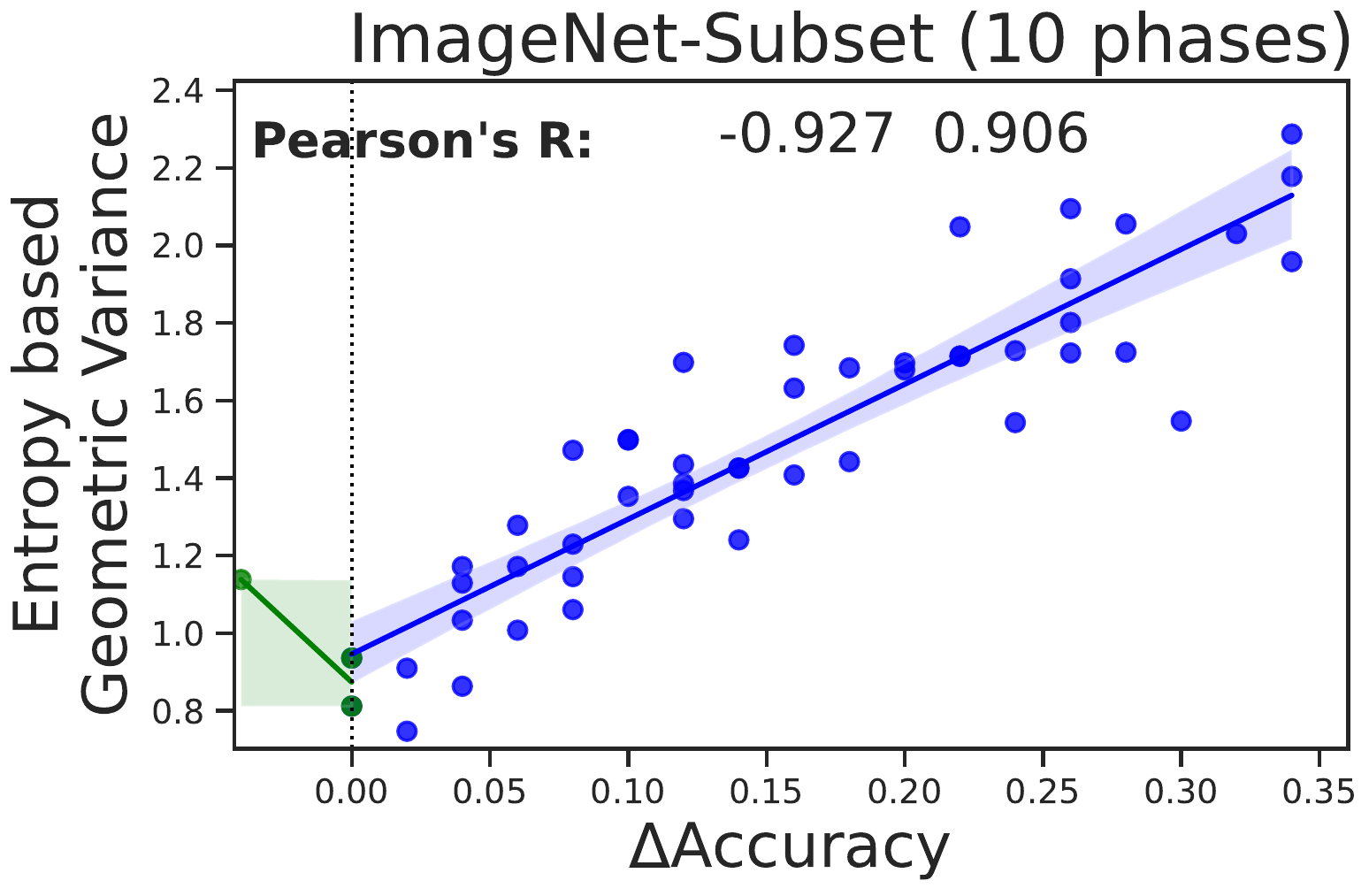}} 
    \vspace{-2mm}
    \caption{Entropy-based geometric variance (HV) in class level as a function of $\Delta$accuracy.}\label{fig:uncertainty}
    \vspace{-5mm}
\end{figure}

\noindent\textbf{Why and When Will Augmentation Integration Help?} When augmentation consensus (iVoro-AC) or integration (iVoro-AI) is applied, the improvement is significant. For example, iVoro-AC obtains 13.76\%, 17.00\%, and 16.50\% improvements upon iVoro on CIFAR-100, TinyImageNet, and ImageNet-Subset, respectively. iVoro-AI itself is worse than iVoro-AC, but if combined with normalization, D\&C, and Voronoi residue, it further elevates the accuracy by a large margin, e.g. as high as 68.70\% (iVoro-NDAI) on 20-phase TinyImageNet and 78.64\% on 10-phase ImageNet-Subset. To investigate the reason of this prominent improvement, we calculate the entropy-based geometric variance in class level and plot them as a function of the $\Delta$accuracy (i.e. the improvement in accuracy after augmentation integration is used), as shown in Fig.~\ref{fig:uncertainty}. Interestingly, there is a clear correlation between HV and $\Delta$accuracy, and this is more notable on ImageNet-Subset (Pearson's R ${\sim}0.9$), probably because of its high resolution ($224 \times 224$). This tendency suggests that the higher the variance within the assignments from augmented images to expanded classes, the better the improvement after using augmentation integration. See Appendix~\ref{supp:unc} for uncertainty analysis in more details.

\noindent\textbf{How Good Should the Feature Extractor Be?} As iVoro is heavily dependent on the feature extractor, which cannot be evolved in any way along the learning process, one may wonder if our method still work with a poorly trained feature extractor. To verify this, we gradually decrease the number of classes used to train the feature extractor $\phi$. As shown in Appendix~\ref{supp:ext}, compared with PASS, the best version of iVoro still has 17.75\%, 13.59\%, 10.89\%, and 1.60\% improvements with 40, 30, 20, and 10 initial classes, respectively. This means that, even if there is no strong feature extractor, our method can still reach acceptable performance higher than the state-of-the-art method.

\noindent\textbf{VD Can Also Go Deeper.} In this section, we extract the feature from the 3\textsuperscript{rd} block and rebuild iVoro. As expected, shown in Appendix~\ref{supp:civd}, the overall performance degenerates %largely
substantially, e.g. iVoro-NAI drops from 65.84\% to 42.03\%. However, and interestingly, if integrated with CCVD and D\&C, iVoro-NDAIL obtains even higher accuracy of 72.34\%. The final results are presented in Tab.~\ref{tab:main}, Fig.~\ref{fig:tiny}, and Appendix~\ref{supp:fig}. Our final model, layered VD, surpasses all previous methods by a large margin of $25.26\%$, $37.09\%$, and $33.21\%$ on CIFAR-100, TinyImageNet, and ImageNet-Subset, respectively, even higher than exemplar-based CIL methods, showing the effectiveness of our geometric framework for combating catastrophic forgetting (Appendix~\ref{supp:forget}). % moved Fig.~\ref{fig:cifar} to appendix

\section{Conclusion} \label{sec:discussion}
In this paper, we use incremental Voronoi Diagram to model the incremental learning problem (iVoro), and propose a number of new techniques that handle various aspects of this VD construction process, including DNN-assisted progressive VD construction (iVoro-D), prototype refinement (iVoro-R), uncertainty-aware label augmentation (iVoro-AC/iVoro-AI), multi-centered VD for multi-layer deep network (iVoro-L) that gradually and greatly improve the CIL performance. Thus, iVoro is shown to be a flexible, scalable, and robust framework that holistically promotes the performance of CIL but still strictly maintain the privacy of previous data. We hope iVoro can open up new avenues for future research about both exemplar-free and exemplar-based CIL. Our code is available at GitHub.

\begin{ack}
  We thank the anonymous reviewers for their constructive comments on this paper.
\end{ack}

%%\section*{References}
\bibliography{ref_fsl,ref_cil,ref_unc}
\bibliographystyle{bibstyle}

%%%%%%%%%%%%%%%%%%%%%%%%%%%%%%%%%%%%%%%%%%%%%%%%%%%%%%%%%%%%
\section*{Checklist}

%%% BEGIN INSTRUCTIONS %%%
% The checklist follows the references.  Please
% read the checklist guidelines carefully for information on how to answer these
% questions.  For each question, change the default \answerTODO{} to \answerYes{},
% \answerNo{}, or \answerNA{}.  You are strongly encouraged to include a {\bf
% justification to your answer}, either by referencing the appropriate section of
% your paper or providing a brief inline description.  For example:
% \begin{itemize}
%   \item Did you include the license to the code and datasets? \answerYes{See Section 4.}
%   \item Did you include the license to the code and datasets? \answerNo{The code and the data are proprietary.}
%   \item Did you include the license to the code and datasets? \answerNA{}
% \end{itemize}
% Please do not modify the questions and only use the provided macros for your
% answers.  Note that the Checklist section does not count towards the page
% limit.  In your paper, please delete this instructions block and only keep the
% Checklist section heading above along with the questions/answers below.
%%% END INSTRUCTIONS %%%

\begin{enumerate}

\item For all authors...
\begin{enumerate}
  \item Do the main claims made in the abstract and introduction accurately reflect the paper's contributions and scope?
    \answerYes{Our scope is exemplar-free class-incremental learning (CIL) and our contribution is a holistic geometric framework including 5 novel components that jointly improve CIL by 25.26\% to 37.09\% on various datasets.}
  \item Did you describe the limitations of your work?
    \answerYes{The limitation is that the feature extractor is frozen all the time, but even though our method still achieves consistent state-of-the-art performance. See Sec.~\ref{sec:experiments}.}
  \item Did you discuss any potential negative societal impacts of your work?
    \answerNA{We discuss the societal impacts but they are positive since our method ensures the privacy of historical data.}
  \item Have you read the ethics review guidelines and ensured that your paper conforms to them?
    \answerYes{We will obey the ethics review guidelines.}
\end{enumerate}

\item If you are including theoretical results...
\begin{enumerate}
  \item Did you state the full set of assumptions of all theoretical results?
    \answerYes{}
  \item Did you include complete proofs of all theoretical results?
    \answerYes{See Appendix~\ref{supp:power}}
\end{enumerate}

\item If you ran experiments...
\begin{enumerate}
  \item Did you include the code, data, and instructions needed to reproduce the main experimental results (either in the supplemental material or as a URL)?
    \answerYes{We provide code link, data links, and implementation details in Appendix.}
  \item Did you specify all the training details (e.g., data splits, hyperparameters, how they were chosen)?
    \answerYes{They are all mentioned in Appendix.}
  \item Did you report error bars (e.g., with respect to the random seed after running experiments multiple times)?
    \answerYes{We include a number of box-plots in the Appendix.}
  \item Did you include the total amount of compute and the type of resources used (e.g., type of GPUs, internal cluster, or cloud provider)?
    \answerYes{}
\end{enumerate}

\item If you are using existing assets (e.g., code, data, models) or curating/releasing new assets...
\begin{enumerate}
  \item If your work uses existing assets, did you cite the creators?
    \answerYes{}
  \item Did you mention the license of the assets?
    \answerYes{}
  \item Did you include any new assets either in the supplemental material or as a URL?
    \answerYes{}
  \item Did you discuss whether and how consent was obtained from people whose data you're using/curating?
    \answerNA{All the datasets and code repositories are publicly available.}
  \item Did you discuss whether the data you are using/curating contains personally identifiable information or offensive content?
    \answerNA{All the datasets are well-accepted by the community, e.g. CIFAR and ImageNet.}
\end{enumerate}

\item If you used crowdsourcing or conducted research with human subjects...
\begin{enumerate}
  \item Did you include the full text of instructions given to participants and screenshots, if applicable?
    \answerNA{}
  \item Did you describe any potential participant risks, with links to Institutional Review Board (IRB) approvals, if applicable?
    \answerNA{}
  \item Did you include the estimated hourly wage paid to participants and the total amount spent on participant compensation?
    \answerNA{}
\end{enumerate}

\end{enumerate}

%%%%%%%%%%%%%%%%%%%%%%%%%%%%%%%%%%%%%%%%%%%%%%%%%%%%%%%%%%%%

\appendix
%\input{06_appendix}
% ====================
\pagebreak
\counterwithin{figure}{section}
\counterwithin{table}{section}
\tableofcontents
% ==================== Extended related work ====================
\section{Extended Related Work} \label{supp:related}

\subsection{Recent Progress in Incremental Learning}\label{sec:related-il}
% <IL, catastrophic forgetting>
Incremental Learning~\citep{rebuffi2017icarl,hou2019learning,wu2019large,zhu2021prototype,liu2021rmm} requires continuously updating a model using a sequence of new tasks without forgetting the old knowledge, which is also referred to as continual learning~\citep{parisi2019continual,9349197,chaudhry2019tiny}. The main challenge of incremental learning is catastrophic forgetting~\citep{mccloskey1989catastrophic,french1999catastrophic,goodfellow2013empirical,kemker2018measuring}, where deep neural network is prone to performance deterioration on the previously learned tasks as the model parameters overfit to the current data to optimize the stability-plasticity trade-off. 

%--> Causation of catastrophic forgetting
\noindent\textbf{Causation of Catastrophic Forgetting.} Generally speaking, in deep neural networks, catastrophic forgetting~\citep{mccloskey1989catastrophic,goodfellow2013empirical,kemker2018measuring} comes from two sources: the feature distribution shifting of the old classes in the feature embedding space as well as the confusion and imbalance of the decision boundary of the classifier when learning new task. The former is caused by the excessive plasticity and parameter changing of the feature extractor of the deep model during finetuning on unseen data/classes, thus deteriorates the feature extraction and prediction on previous classes; while the latter is due to the highly overfitting and bias of the classifier on current task as well as the overlapping between the representation of new and old classes in the feature space. 

% <Three Scenarios>
\noindent\textbf{Incremental Learning Scenarios.} Three common incremental learning scenarios are widely explored in recent papers~\citep{van2019three}. 

% <TIL>
\textit{Task-incremental learning} (TIL)~\citep{ostapenko2019learning,shin2017continual,kirkpatrick2017overcoming,zenke2017continual,wu2018incremental,lopez2017gradient,chaudhry2019tiny,buzzega2020dark,cha2021co2l,pham2021dualnet,fernando2017pathnet} incrementally learns a sequence of tasks in multiple phases, where each task contains unseen data of a new set of classes. To mitigate catastrophic forgetting, TIL assumes a simple setting where the task identity is known at inference time. The methods under this scenario keep leaning new task-independent classifiers or growing the model capacity by attaching additional modules (e.g. kernels, layers or branches), each corresponding to a specific task or a subset of classes. Since the task ID is available during inference, the model can directly select proper classifier or module without inferring task identity, which effectively solves the confusion boundary and classifier bias between old and new tasks, and often achieves satisfying performance. However, knowing task identity at test time is normally unrealistic in real-world situation hence restricts practical usage. Moreover, it may incur unbounded memory consumption for super long task sequence if increasing the model capacity for new tasks. 

% <CIL>
Unlike TIL constrained by the availability of task identity, \textit{class-incremental learning} (CIL)~\citep{liu2020more,belouadah2019il2m,chaudhry2018riemannian,zhu2021prototype,douillard2020podnet,rebuffi2017icarl,hou2019learning,liu2020mnemonics,liu2021adaptive,liu2021rmm} updates a unified classifier for all classes learned so far while task identity is no longer required during inference. To compensate the missing task identity and alleviate forgetting issue, a branch of works~\citep{rebuffi2017icarl,hou2019learning,liu2021rmm,douillard2020podnet,castro2018end,wu2019large} alternatively follow a memory-based setting, in which a limited number of samples from old classes (e.g., 20 exemplars per class) is stored and maintained in a memory buffer, which are later replayed to jointly train the model with current data (normally combined with knowledge distillation) in order to constrain the feature distribution shifting of the old classes and the decision boundary bias of the classifier. However, their performance deteriorates with smaller buffer size, and eventually, the storing and sharing of previous data, e.g. medical images, may not be feasible when memory limits and privacy issue are taken into consideration. %Alternatively, other works try to train a deep generative model incrementally to generate fake samples of previous classes~\citep{ostapenko2019learning,shin2017continual,wu2018incremental,kemker2017fearnet}. But the generative model itself is also prone to catastrophic forgetting and not able to handle complex dataset thus the quality of generated image is not reliable. 
Given the potential memory issue, another direction of works~\citep{kirkpatrick2017overcoming,zenke2017continual,li2017learning,dhar2019learning,zhu2021prototype} intend to explore CIL in a much challenging setting without memory rehearsal, mainly based on regularization and knowledge distillation techniques, which is known as exemplar-free CIL. In this paper, we are following this CIL setting. 

% <DIL>
\textit{Domain-incremental learning} (DIL)~\citep{rostami2021lifelong,tang2021gradient,volpi2021continual}, different from the aforementioned two scenarios, incrementally learning new domains of the same classes in each phase. Some domain adaptation techniques, e.g. meta learning, data shifting, domain randomization, are implemented in DIL to increase the model robustness and generalizability to handle various domain distributions. Since this scenario is not quite related to this paper, no detailed discussion will be included. 

% <Categories>
\noindent\textbf{Categories of Incremental Learning Methods.} There are three categories of existing IL methods to overcome catastrophic forgetting~\citep{9349197}. 

% <regularize, KD>
\textit{Regularization-based} methods constrain the plasticity of the model to preserve old knowledge. This can be addressed by directly penalizing the changes of important parameters for previous tasks~\citep{aljundi2018memory,chaudhry2018riemannian,kirkpatrick2017overcoming,zenke2017continual,kumar2021bayesian} or regularizing the gradients when training on unseen data~\citep{lopez2017gradient,chaudhry2018efficient}. Knowledge distillation is another regularization solution, which is widely used in various IL methods to implicitly consolidate previous knowledge by introducing regularization loss term on model representations, including output logits or probabilities~\citep{li2017learning,schwarz2018progress,rebuffi2017icarl,castro2018end} and intermediate features~\citep{hou2019learning,dhar2019learning,douillard2020podnet,zhu2021prototype}. Some other works focus on correcting the classifier bias on new classes~\citep{belouadah2019il2m,wu2019large,belouadah2020scail,zhao2020maintaining}. 

% <rehearsal, memory, GAN>
\textit{Rehearsal-based} methods either store and replay a limited amount of exemplars from old classes as raw images~\citep{rebuffi2017icarl,hou2019learning,liu2021rmm,chaudhry2019tiny,buzzega2020dark} or embedded features~\citep{hayes2020remind,iscen2020memory} to jointly train the model in the incremental phases, or alternatively generate exemplars of previous classes~\citep{ostapenko2019learning,shin2017continual,wu2018incremental,kemker2017fearnet}. The former relies on memory buffer for all learned classes, where the performance is constrained by the buffer size limits and it is impracticable when data privacy is required and storing data is prohibited. The latter requires continuously learning a deep generative model, which is also prone to catastrophic forgetting thus the quality of generated exemplars is not reliable. 

% <architecture, expand, prune>
\textit{Architecture-based} methods aims at dynamically adapting task-specific sub-network architectures, which requires task identity to select proper sub-network. Some works directly expand the network by adding new layers or branches~\citep{rusu2016progressive,li2019learn,wang2017growing,yoon2017lifelong}, which is limited in practice due to unbounded model parameter growth. Others freeze partial network with masks for old tasks~\citep{golkar2019continual,hung2019compacting,mallya2018packnet,serra2018overcoming}, but suffering from running out of model parameters for new knowledge. The architecture-based methods are usually combined with memory buffer and distillation, and can achieve good results. 

% <Our Work>
Our work is focusing on the most challenging but practical non-exemplar class-incremental learning problem, which is a general real-world scenario when no old data can be stored due to memory limits or data privacy and task identity is unavailable during inference, with the constraint of fixed model capacity in the same time. 

\subsection{Computational Geometry for Deep Learning}\label{sec:related-geo} 

Computational Geometry for Deep Learning is an emerging toolset for studying various aspects of deep learning. The geometric structure of deep neural networks is first hinted at by~\citep{raghu2017expressive} who reveals that piecewise linear activations subdivide input space into convex polytopes. Then,~\citep{power2019} points out that the exact structure is a Power Diagram~\citep{aurenhammer1987power} which is subsequently applied upon recurrent neural network~\citep{wang2018max} and generative model~\citep{balestriero2020max}. The Power/Voronoi Diagram subdivision, however, is not necessarily the optimal model for describing feature space. Recently,~\citep{ChenHLX13,ChenHL017,HuangCX21} uses an influence function $F(\gC, \vz)$ to measure the joint influence of all objects in $\gC$ on a query $\vz$ to build a Cluster-induced Voronoi Diagram (CIVD). Voronoi subdivision has also been used for deep learning uncertainty calibration~\citep{ma2021improving}, metric-based few-shot learning~\citep{ma2022fewshoticlr}, adversarial robustness~\citep{sitawarin2021adversarial} and topological data analysis~\citep{polianskii2019voronoi,polianskii2020voronoi,poklukar2022delaunay}, and medical applications~\citep{ma2018improved,ma2019neural}.

% ==================== 2D-MNIST ====================
\section{Demonstrative Illustration on MNIST Dataset in 2D Space} \label{supp:mnist}
MNIST~\citep{lecun1998mnist} was used for the illustration of four methods, fine-tuning, PASS~\citep{zhu2021prototype}, iVoro, and iVoro-AC, because of the convenience of embedding the examples into $\R^2$. The total 10 classes are split into a sequence of 4, 3, and 3 classes. A ResNet-18 model is used as a feature extractor for all four methods. (\textbf{A}) In fine-tuning, the model is firstly trained on the 4 classes in the first phase, and then fine-tuned only on the subsequent 3 and 3 classes in phase 2 and phase 3. (\textbf{B}) In PASS, SSL-based label augmentation is applied on all three phases and expands the classes to be 16, 9, and 9 classes. The default hyper-parameters are used to train PASS (i.e. the weight for knowledge distillation is 10 and the weight for prototype augmentation is 10). To ensure the final subdivision of space is a Voronoi diagram, Thm.~\ref{thm:thm} (i.e. Voronoi diagram reduction in Algorithm~\ref{alg:power}) is applied during the training of fine-tuning and PASS. (\textbf{C}) In iVoro, the feature extractor from the first phase of (B) is frozen and used without fine-tuning for all phases. The feature means are calculated as prototypes and no feature transformation is used. Note that only the features from the original images without rotation are used. (\textbf{D}) The only difference with (C) is that all the expanded classes are also considered as independent cells, allowing for further integration.

\noindent\textbf{Result Analysis.} For (A) fine-tuning and (B) PASS, the model's accuracy for data from individual phases up to now are also shown in shadow. Fine-tuning are able to achieve near-perfect prediction for classes in the current locally, but fails to maintain satisfactory performance on any historical class (accuracy ${\sim}0$). PASS, on the other hand, basically deteriorates slightly on the classes from the first phase, due to the high KD loss and prototypical loss, but it also becomes almost incapable of learning on new classes (accuracy ${\sim}0$). iVoro, surprisingly, obtains superior accuracy (64.84\%, 24.28\% higher than PASS) by only using a fixed feature extractor trained from only 4 classes (16 expanded classes). iVoro-AC achieves comparable result (61.77\%) with iVoro, because the 2D embedding makes it harder to demonstrate the efficacy of our proposed techniques.

% ==================== Notations and acronyms ====================
\section{Notations and Acronyms} \label{supp:notation}
In this section, we list all notations used in Methodology in Tab.~\ref{tab:math}, notations and acronyms for various geometric structures used in the paper in Tab.~\ref{tab:notationI}, and all ablation methods in Tab.~\ref{tab:ivoros}.
%Please add the following packages if necessary:
%\usepackage{booktabs, multirow} % for borders and merged ranges
%\usepackage{soul}% for underlines
%\usepackage[table]{xcolor} % for cell colors
%\usepackage{changepage,threeparttable} % for wide tables
%If the table is too wide, replace \begin{table}[!htp]...\end{table} with
%\begin{adjustwidth}{-2.5 cm}{-2.5 cm}\centering\begin{threeparttable}[!htb]...\end{threeparttable}\end{adjustwidth}
\begin{table}[!htp]\centering
\vspace{3mm}
\caption{Complete list of all notations used in Methodology~\ref{sec:method}.}\label{tab:math}
\resizebox{0.9\textwidth}{!}{%
    \scriptsize
    \begin{tabular}{ll}\toprule
    Notations &Descriptions \\\midrule
    $T$ &total phase, $T \in \R$ \\
    $t$ &current phase, $t \in \{1,...,T\}$ \\
    $\tau$ &historical phase, $\tau \in \{1,...,t\}$ \\
    $\gD_t$ &dataset in phase $t$ \\
    $(\vx_{t,i}, y_{t,i})$ &data (image) and label in phase $t$, $i \in \{1,...,N_t\}, t \in \{1,...,T\}$ \\
    $\gC_t$ &the set of classes in phase $t$ \\
    $\gC_{t,k}$ &the $t$\textsuperscript{th} class in phase $t$ \\
    $N_{t,k}$ &number of examples for class $k$ in phase $t$ \\
    $N_t$ &number of all examples in phase $t$, i.e. $N_t = {\textstyle\sum}_{k=1}^{K_t} N_{t,k} $ \\
    $\phi$ &feature extractor (a deep neural network) \\
    $\phi^{(l)}$ &feature extractor, but only the features from the $l$\textsuperscript{th} layer is used \\
    $\vz$ &feature for $\vx$, i.e. $\vz = \phi(\vx), \vz \in \R^n$ \\
    $\theta$ &classification head, can be either Voronoi diagram, or logistic regression \\
    $\vc_{\tau, k}$ &prototypical Voronoi center for phase $\tau$ and class $\gC_{\tau,k}$ \\
    $\tilde{\vc}_{\tau, k}$ &linear probing-induced Voronoi center for phase $\tau$ and class $\gC_{\tau,k}$ \\
    $\tilde{\vc}_{\tau, k}'$ &Voronoi residual prototypical center for phase $\tau$ and class $\gC_{\tau,k}$ \\
    $f$ &$L_2$ normalization \\
    $g$ &linear transformation $g_{w,\eta}(\vz) = w\vz + \eta$ \\
    $h$ &Tukey’s ladder of powers transformation, parameterized by $\lambda$ \\
    $\mW_{t,k}, \evb_{t,k}$ &linear weight and bias for class $\gC_{t,k}$ \\
    $\vv, q$ &parameters for linear bisector $\vv^T \vz' - q = 0$ \\
    $\mW_{t,k}^{(0)}, \evb_{t,k}^{(0)}$ &the initialization of linear probing educated by prototypes and Thm.~\ref{thm:thm} \\
    $\Delta \mW_t, \Delta \vb_t$ &the residues from $\mW_t^{(0)}, \vb_t^{(0)}$; SGD will only be applied on $\Delta \mW_t, \Delta \vb_t$ \\
    $\alpha$ &index of four rotations, $\alpha \in \{0, 1, 2, 3\}$ \\
    $\vd^{(\alpha, \alpha)} \in \R^K$ &the collection of the distances from $\phi(\vx^{\alpha})$ to $K$ classes with rotation index $\alpha$ \\
    HV &Entropy-based geometric variance \\
    \bottomrule
    \end{tabular}}
    \vspace{3mm}
\end{table}

%Please add the following packages if necessary:
%\usepackage{booktabs, multirow} % for borders and merged ranges
%\usepackage{soul}% for underlines
%\usepackage[table]{xcolor} % for cell colors
%\usepackage{changepage,threeparttable} % for wide tables
%If the table is too wide, replace \begin{table}[!htp]...\end{table} with
%\begin{adjustwidth}{-2.5 cm}{-2.5 cm}\centering\begin{threeparttable}[!htb]...\end{threeparttable}\end{adjustwidth}
\setlength\intextsep{0pt} % change this value
\begin{table}[!hp]\centering
    \caption{\textcolor{black}{Notations and acronyms for VD, PD, and CCVD, three geometric structures used in the paper.}}\label{tab:notationI}
    \resizebox{\textwidth}{!}{%
    %\small
    \begin{tabular}{ccll}\toprule
    \textbf{Geometric Structures} &\textbf{Acronyms} &\textbf{Notations} &\textbf{Description} \\\midrule
    \multirow{2}{*}{Voronoi Diagram} &\multirow{2}{*}{VD} &$\vc_k$ &center for a Voronoi cell $\omega_k, k \in \{1,..,K\}$ \\
    & &$\omega_k$ &dominating region for center $\vc_k, k \in \{1,..,K\}$\\\midrule
    \multirow{3}{*}{Power Diagram~\citep{aurenhammer1987power}} &\multirow{3}{*}{PD} &$\vc_k$ &center for a Power cell $\omega_k, k \in \{1,..,K\}$ \\
    & &$\nu_k$ &weight for center $\vc_k, k \in \{1,..,K\}$ \\
    & &$\omega_k$ &dominating region for center $\vc_k, k \in \{1,..,K\}$\\\midrule
    \multirow{5}{*}{\makecell{Cluster-to-cluster \\ Voronoi Diagram~\citep{ma2022fewshot}}} &\multirow{4}{*}{CCVD} &$\gC_k$ &cluster as the "center" for a CCVD cell $\omega_k, k \in \{1,..,K\}$ \\
    & &$\omega_k$ &dominating region for cluster $\gC_k$\\
    & &$\gC(\vz)$ &the cluster that query point $\vz$ belongs \\
    & &$F$ &influence function $F(\gC_k,\gC(\vz))$ from $\gC_k$ to query cluster $\gC(\vz)$\\
    & &$\alpha$ &magnitude of the influence \\
    \bottomrule
    \end{tabular}}
    \vspace{3mm} % change this value
\end{table}

% \multirow{4}{*}{\makecell{Cluster-induced \\ Voronoi Diagram}} &\multirow{4}{*}{CIVD} &$\gC_k$ &cluster as the "center" for a CIVD cell $\omega_k, k \in \{1,..,K\}$ \\
% & &$\omega_k$ &dominating region for cluster $\gC_k$\\
% & &$F$ &influence function $F(\gC_k,\vz)$ from cluster $\gC_k$ to query point $\vz$\\
% & &$\alpha$ &magnitude of the influence \\\midrule
%Please add the following packages if necessary:
%\usepackage{booktabs, multirow} % for borders and merged ranges
%\usepackage{soul}% for underlines
%\usepackage[table]{xcolor} % for cell colors
%\usepackage{changepage,threeparttable} % for wide tables
%If the table is too wide, replace \begin{table}[!htp]...\end{table} with
%\begin{adjustwidth}{-2.5 cm}{-2.5 cm}\centering\begin{threeparttable}[!htb]...\end{threeparttable}\end{adjustwidth}
\begin{table}[!htp]\centering
\caption{Complete list of all variants of iVoro.}\label{tab:ivoros}
\resizebox{\textwidth}{!}{%
\small
    \begin{tabular}{lccccccc}\toprule
    \textbf{Methods} &\textbf{Prototype} &\textbf{Normalization} &\textbf{D\&C} &\textbf{\makecell{Voronoi \\ Residue}} &\textbf{\makecell{Augmentation \\ Consensus}} &\textbf{\makecell{Augmentation \\ Integration}} &\textbf{Layered VD} \\\midrule
    iVoro &\cmark &\xmark &\xmark &\xmark &\xmark &\xmark &\xmark \\
    iVoro-N &\cmark &\cellcolor[HTML]{f2f2f2}\cmark &\xmark &\xmark &\xmark &\xmark &\xmark \\
    iVoro-D &\cmark &\xmark &\cellcolor[HTML]{f2f2f2}\cmark &\xmark &\xmark &\xmark &\xmark \\
    iVoro-ND &\cmark &\cellcolor[HTML]{f2f2f2}\cmark &\cellcolor[HTML]{f2f2f2}\cmark &\xmark &\xmark &\xmark &\xmark \\
    iVoro-R &\cmark &\xmark &\xmark &\cellcolor[HTML]{f2f2f2}\cmark &\xmark &\xmark &\xmark \\
    iVoro-DR &\cmark &\xmark &\cellcolor[HTML]{f2f2f2}\cmark &\cellcolor[HTML]{f2f2f2}\cmark &\xmark &\xmark &\xmark \\
    iVoro-NDR &\cmark &\cellcolor[HTML]{f2f2f2}\cmark &\cellcolor[HTML]{f2f2f2}\cmark &\cellcolor[HTML]{f2f2f2}\cmark &\xmark &\xmark &\xmark \\
    iVoro-AC &\cmark &\xmark &\xmark &\xmark &\cellcolor[HTML]{f2f2f2}\cmark &\xmark &\xmark \\
    iVoro-AI &\cmark &\xmark &\xmark &\xmark &\xmark &\cellcolor[HTML]{f2f2f2}\cmark &\xmark \\
    iVoro-NAC &\cmark &\cellcolor[HTML]{f2f2f2}\cmark &\xmark &\xmark &\cellcolor[HTML]{f2f2f2}\cmark &\xmark &\xmark \\
    iVoro-NAI &\cmark &\cellcolor[HTML]{f2f2f2}\cmark &\xmark &\xmark &\xmark &\cellcolor[HTML]{f2f2f2}\cmark &\xmark \\
    iVoro-NDAC &\cmark &\cellcolor[HTML]{f2f2f2}\cmark &\cellcolor[HTML]{f2f2f2}\cmark &\xmark &\cellcolor[HTML]{f2f2f2}\cmark &\xmark &\xmark \\
    iVoro-NDAI &\cmark &\cellcolor[HTML]{f2f2f2}\cmark &\cellcolor[HTML]{f2f2f2}\cmark &\xmark &\xmark &\cellcolor[HTML]{f2f2f2}\cmark &\xmark \\
    iVoro-RAC &\cmark &\xmark &\xmark &\cellcolor[HTML]{f2f2f2}\cmark &\cellcolor[HTML]{f2f2f2}\cmark &\xmark &\xmark \\
    iVoro-RAI &\cmark &\xmark &\xmark &\cellcolor[HTML]{f2f2f2}\cmark &\xmark &\cellcolor[HTML]{f2f2f2}\cmark &\xmark \\
    iVoro-DRAC &\cmark &\xmark &\cellcolor[HTML]{f2f2f2}\cmark &\cellcolor[HTML]{f2f2f2}\cmark &\cellcolor[HTML]{f2f2f2}\cmark &\xmark &\xmark \\
    iVoro-DRAI &\cmark &\xmark &\cellcolor[HTML]{f2f2f2}\cmark &\cellcolor[HTML]{f2f2f2}\cmark &\xmark &\cellcolor[HTML]{f2f2f2}\cmark &\xmark \\
    iVoro-NACL &\cmark &\cellcolor[HTML]{f2f2f2}\cmark &\xmark &\xmark &\cellcolor[HTML]{f2f2f2}\cmark &\xmark &\cellcolor[HTML]{f2f2f2}\cmark \\
    iVoro-NAIL &\cmark &\cellcolor[HTML]{f2f2f2}\cmark &\xmark &\xmark &\xmark &\cellcolor[HTML]{f2f2f2}\cmark &\cellcolor[HTML]{f2f2f2}\cmark \\
    iVoro-NDACL &\cmark &\cellcolor[HTML]{f2f2f2}\cmark &\cellcolor[HTML]{f2f2f2}\cmark &\xmark &\cellcolor[HTML]{f2f2f2}\cmark &\xmark &\cellcolor[HTML]{f2f2f2}\cmark \\
    iVoro-NDAIL &\cmark &\cellcolor[HTML]{f2f2f2}\cmark &\cellcolor[HTML]{f2f2f2}\cmark &\xmark &\xmark &\cellcolor[HTML]{f2f2f2}\cmark &\cellcolor[HTML]{f2f2f2}\cmark \\
    \bottomrule
    \end{tabular}}
\end{table}
% ==================== Lemma ====================
\section{Power Diagram Subdivision and Voronoi Reduction} \label{supp:power}
In this section we provide the proof of Theorem~\ref{thm:thm}.
\begin{lemma} \label{lem:power}
    The vertical projection from the lower envelope of the hyperplanes $\{\Pi_k(\vz) : \mW_k^T \vz + \evb_k\}_{k=1}^K$ onto the input space $\R^n$ defines the cells of a PD.
\end{lemma}

\noindent\textbf{Theorem 2.1} (Voronoi Diagram Reduction~\citep{ma2022fewshot}). The linear classifier parameterized by $\mW, \vb$ partitions the input space $\R^n$ to a Voronoi Diagram with centers $\{ \tilde{\vc}_1,...,\tilde{\vc}_K \}$ given by $\tilde{\vc}_k = \frac{1}{2} \mW_k$ if $\evb_k = -\frac{1}{4} ||\mW_k||_2^2, k = 1,...,K$.

\begin{proof}
    We first articulate Lemma \ref{lem:power} and find the exact relationship between the hyperplane $\Pi_k(\vz)$ and the center of its associated cell in $\R^n$. By Definition \ref{def:power}, the cell for a point $\vz \in \R^n$ is found by comparing $d(\vz, \vc_k)^2 - \nu_k$ for different $k$, so we define the power function $p(\vz, S)$ expressing this value
    \begin{equation}\label{eq:powerfun}
        p(\vz, S) = (\vz - \vu)^2 - r^2
    \end{equation}
    in which $S \subseteq \R^n$ is a sphere with center $\vu$ and radius $r$. In fact, the weight $\nu$ associated with a center in Definition \ref{def:power} can be interpreted as the square of the radius $r^2$.
    Next, let $U$ denote a paraboloid $y = \vz^2$, let $\Pi(S)$ be the transform that maps sphere $S$ with center $\vu$ and radius $r$ into hyperplane
    \begin{equation}\label{eq:pifun}
        \Pi(S): y = 2\vz \cdot \vu - \vu \cdot \vu + r^2.
    \end{equation}
    It can be proved that $\Pi$ is a bijective mapping between arbitrary spheres in $\R^n$ and nonvertical hyperplanes in $\R^{n+1}$ that intersect $U$~\citep{aurenhammer1987power}.
    Further, let $\vz'$ denote the vertical projection of $\vz$ onto $U$ and $\vz''$ denote its vertical projection onto $\Pi(S)$, then the power function can be written as
    \begin{equation}\label{eq:pifun2}
        p(\vz, S) = d(\vz, \vz') - d(\vz, \vz''),
    \end{equation}
    which implies the following relationships between a sphere in $\R^n$ and an associated hyperplane in $\R^{n+1}$ (Lemma 4 in~\citep{aurenhammer1987power}): let $S_1$ and $S_2$ be nonco-centeric spheres in $\R^n$, then the bisector of their Power cells is the vertical projection of $\Pi(S_1) \cap \Pi(S_2)$ onto $\R^n$.
    Now, we have a direct relationship between sphere $S$, and hyperplane $\Pi(S)$, and comparing equation (\ref{eq:pifun}) with the hyperplanes used in logistic regression $\{\Pi_k(\vz) : \mW_k^T \vz + \evb_k\}_{k=1}^K$ gives us
    \begin{equation}\label{eq:mapping}
        \begin{aligned}
            \vu &= \frac{1}{2} \mW_k \\
            r^2 &= \evb_k + \frac{1}{4} ||\mW_k||_2^2.
        \end{aligned}
    \end{equation}
    Although there is no guarantee that $\evb_k + \frac{1}{4} ||\mW_k||_2^2$ is always positive for an arbitrary logistic regression model, we can impose a constraint on $r^2$ to keep it be zero during the optimization, which implies
    \begin{equation}\label{eq:b}
        \evb_k = -\frac{1}{4} ||\mW_k||_2^2.
    \end{equation}
    By this way, the radii for all $K$ spheres become identical (all zero). After the optimization of logistic regression model, the centers $\{\frac{1}{2} \mW_k\}_{k=1}^K$ will be used as probing-induced Voronoi centers.
\end{proof}

% ==================== Dataset statistics ====================
\section{Dataset Details}\label{supp:dataset}
Here we give the detailed statistics of the three datasets used in the paper. Augmentation consensus (iVoro-AC) and augmentation integration (iVoro-AI) work more favorably with images with higher resolution, e.g. ImageNet-Subset (see Fig.~\ref{fig:uncertainty}), as the rotation operation makes less sense if the image is blur.
%Please add the following packages if necessary:
%\usepackage{booktabs, multirow} % for borders and merged ranges
%\usepackage{soul}% for underlines
%\usepackage[table]{xcolor} % for cell colors
%\usepackage{changepage,threeparttable} % for wide tables
%If the table is too wide, replace \begin{table}[!htp]...\end{table} with
%\begin{adjustwidth}{-2.5 cm}{-2.5 cm}\centering\begin{threeparttable}[!htb]...\end{threeparttable}\end{adjustwidth}
\begin{table}[!htp]\centering
\vspace{3mm}
\caption{Summarization of the datasets used in the paper.}\label{tab:datas}
\resizebox{\textwidth}{!}{%
    \small
    \begin{tabular}{lcccc}\toprule
    \textbf{Datasets} &\textbf{Image size} &\textbf{Training Images} &\textbf{Total Classes} &\textbf{Number of Phases} \\\midrule
    CIFAR-100~\citep{krizhevsky2009learning} &32 × 32 × 3 &60000 &100 &5, 10, 20 \\
    TinyImageNet~\citep{le2015tiny} &64 × 64 × 3 &100000 &200 &5, 10, 20 \\
    ImageNet-Subset~\citep{deng2009imagenet2} &224 × 224 × 3 &130000 &100 &10 \\
    \bottomrule
    \end{tabular}}
    \vspace{3mm}
\end{table}
% ==================== Ablation ====================
\section{Implementation Details and Result Analysis of Comprehensive Ablation Studies}\label{supp:impl}
\noindent\textbf{iVoro.} We generally follow the protocol of PASS~\citep{zhu2021prototype} to train the feature extractor $\phi$ only on the data from the first phase, i.e. 50 (for 5/10 phases) or 40 (for 20 phases) classes of CIFAR-100, 100 classes of TinyImageNet, and 50 classes of ImageNet-Subset. 
We also reproduce the results of PASS, using the same hyper-parameters, e.g. knowledge distillation loss at 10, and prototype augmentation loss at 10.
The simplest iVoro method (i.e. vanilla Voronoi diagram, or 1-nearest-neighbor) can already achieve comparable or better results than the state-of-the-art non-exemplar CIL method. 
For example, the difference in accuracy in comparison to PASS is 0.29\%/6.91\%/3.63\% for 5/10/20-phase CIFAR-100, -3.58\%/-1.09\%/5.43\% for 5/10/20-phase TinyImageNet, and 4.76\% for 10-phase ImageNet-Subset.
Notably, there is always a significant elevation of accuracy on long-phase data, suggesting the continuous fine-tuning of model, even with improved loss functions, tends to forget seriously on earlier datasets.

\noindent\textbf{iVoro-N.} To inspect the effectiveness of parameterized feature transformation, we apply $L_2$ normalization with/without Tukey’s ladder of powers transformation ($\lambda$ varying from 0.3 to 0.9), and compare with iVoro. The detailed analysis is presented in Sec.~\ref{supp:norm}.

\noindent\textbf{iVoro-D/iVoro-ND.} The detailed algorithm of iVoro-D is presented in Alg.~\ref{alg:ivoro_d}. Specifically, for each phase $\tau \in \{1,...,t\}$, the local dataset $\gD_{\tau}$ is used to train a logistic regression model with weight decay $\beta$ at 0.0001 and initial learning rate at 0.001. The result is also shown in Tab.~\ref{tab:abl}. Aided by D\&C algorithm and local logistic regression, iVoro-D is consistently better than iVoro, e.g. 0.48\%${\sim}$0.96\% higher on CIFAR-100, 1.75\%${\sim}$3.44\% higher on TinyImageNet, and 0.72\% higher on ImageNet-Subset. When further combined with feature normalization, iVoro-ND achieves even higher accuracy, 54.72\% on 20-phase CIFAR-100, 42.10\% on TinyImageNet, and 58.52\% on ImageNet-Subset. 

\noindent\textbf{iVoro-R/iVoro-DR/iVoro-NDR.} iVoro-D can provide better discrimination with one phase, whereas iVoro-R aims at determine better boundaries across phases by seeking for better prototypes. In iVoro-R, all prototypes are calculated and then assigned to the local logistic regression. After the training of linear model is done, the new Voronoi centers derived from the linear model are stored and for all the subsequent space subdivision. Compared to the baseline iVoro method, iVoro-R obtains 0.03\%${\sim}$0.71\% improvement on CIFAR-100, 1.02\%${\sim}$1.89\% improvement on TinyImageNet, and 0.50\% improvement on ImageNet-Subset. 
In iVoro-DR, the within-phase decision boundaries are still determined by probing-induced Voronoi centers (like iVoro-D), but the cross-phase boundaries is instead delineated by the same way as iVoro-R. iVoro-DR is consistently better than iVoro-R, and this is more prominent on more complex dataset e.g. TinyImageNet (2.64\% improvement for 5 phases). 
Additionally, feature normalization works favorably with iVoro-R, and iVoro-DR, and the three (iVoro-NDR) collectively boosts the accuracy to 54.54\% on 20-phase CIFAR-100, 41.52\% on 20-phase TinyImageNet, and 57.60\% on 10-phase ImageNet-Subset. 

\noindent\textbf{iVoro-AC/iVoro-AI/iVoro-NAC/iVoro-NAI.} While the previous variants of iVoro only consider the prediction on the original image/class, here we show that the prediction can be substantially improved by augmentation consensus and augmentation integration proposed in this paper. Specifically, iVoro-AC improves iVoro by 13.76\%${\sim}$14.58\% on CIFAR-100, by 16.99\%${\sim}$17.00\% on TinyImageNet, and by 16.50\% on ImageNet-Subset. iVoro-AI itself generally works worse than iVoro-AC, e.g. improves up to 6.73\% on CIFAR-100, up to 9.82\% on TinyImageNet, and 5.26\% on ImageNet-Subset, but if combined with feature normalization, iVoro-NAI performs much better than iVoro-NAC on TinyImageNet (up to 65.83\%) and ImageNet-Subset (77.12\%). 

\noindent\textbf{iVoro-NDAI/iVoro-RAC/iVoro-RAI/iVoro-DRAI.} When augmentation consensus/integration is used, the previous variants iVoro-N, iVoro-D, and iVoro-R can all be promoted. Generally, adding an additional component will bring in more performance gain, as shown in Tab.~\ref{tab:abl} in details.

\noindent\textbf{iVoro-NACL/iVoro-NAIL/iVoro-NDAIL.} We further validate the layered VD for multiple feature spaces. As a proof of concept, we only extract the feature from the third block to build an additional VD and integrate using CCVD~\ref{def:ccvd} with $\gamma$ set at 1. Compared with iVoro-NAC, layered VD (iVoro-NACL) further improves both average accuracy and last accuracy on CIFAR-100 under 5/10 phases settings (gain 2.36\%/2.20\% on last accuracy respectively), which also achieves the highest performance on the given settings among all ablation settings. The final performance on ImageNet-Subset is also improved by 2.52\%. However, layered VD does not make obvious difference on TinyImageNet, and even deteriorates the performance on CIFAR-100 by 7.89\%/8.87\% when task sequence is 20, which shows that layered VD may not work well on long sequence and large class set when paired with augmentation consensus. 
When adding layered VD to iVoro-NAI, labelled as iVoro-NAIL, significant performance gain is observed on all experiments of CIFAR-100 (9.1\%/9.1\%/15.76\% average accuracy increments on 5/10/20 phases) and large improvement is also achieved on ImageNet-Subset (6.06\% last accuracy growth). On the contrary, only limited gain is presented on TinyImageNet. The above ablation results may demonstrate that layered VD with only augmentation integration works extraordinarily well for small class set (100 classes), but not very effective when class number increases. 
% iVoro-NDAIL, compared with NAIL. 

\noindent\textbf{Robustness Analysis.} We run PASS 5 times on CIFAR-100 with 10 phases setting, and the last accuracies (\%) are distributed as 48.25\%, 49.03\%, 53.03\%, 53.95\%, 54.75\%, with mean and standard deviation (std) as 51.80\%$\pm$2.65\%; meanwhile, we also test PASS for 5 runs on ImageNet-Subset 10 phases setting, and the last accuracies (\%) are 49.85\%, 50.63\%, 51.03\%, 51.88\%, 52.52\%, with mean$\pm$std as 51.18\%$\pm$0.94\%. As shown above, even though PASS achieves relatively good accuracy on average, the training is not very stable on CIFAR-100, where the difference between the highest and lowest accuracy on CIFAR-100 is as large as 6.5\%; while the performance on ImageNet-Subset is slightly better regarding to robustness, PASS still ranges from 49.85\% to 52.52\%. However, compared to PASS, our iVoro method is naturally robust to various datasets with no fluctuation in performance, due to the frozen feature extractor and unbiased classifier based on VD. 

\vspace{1em}
\begin{figure}[ht]
    \centering
    \subfloat{\includegraphics[width=0.7\textwidth]{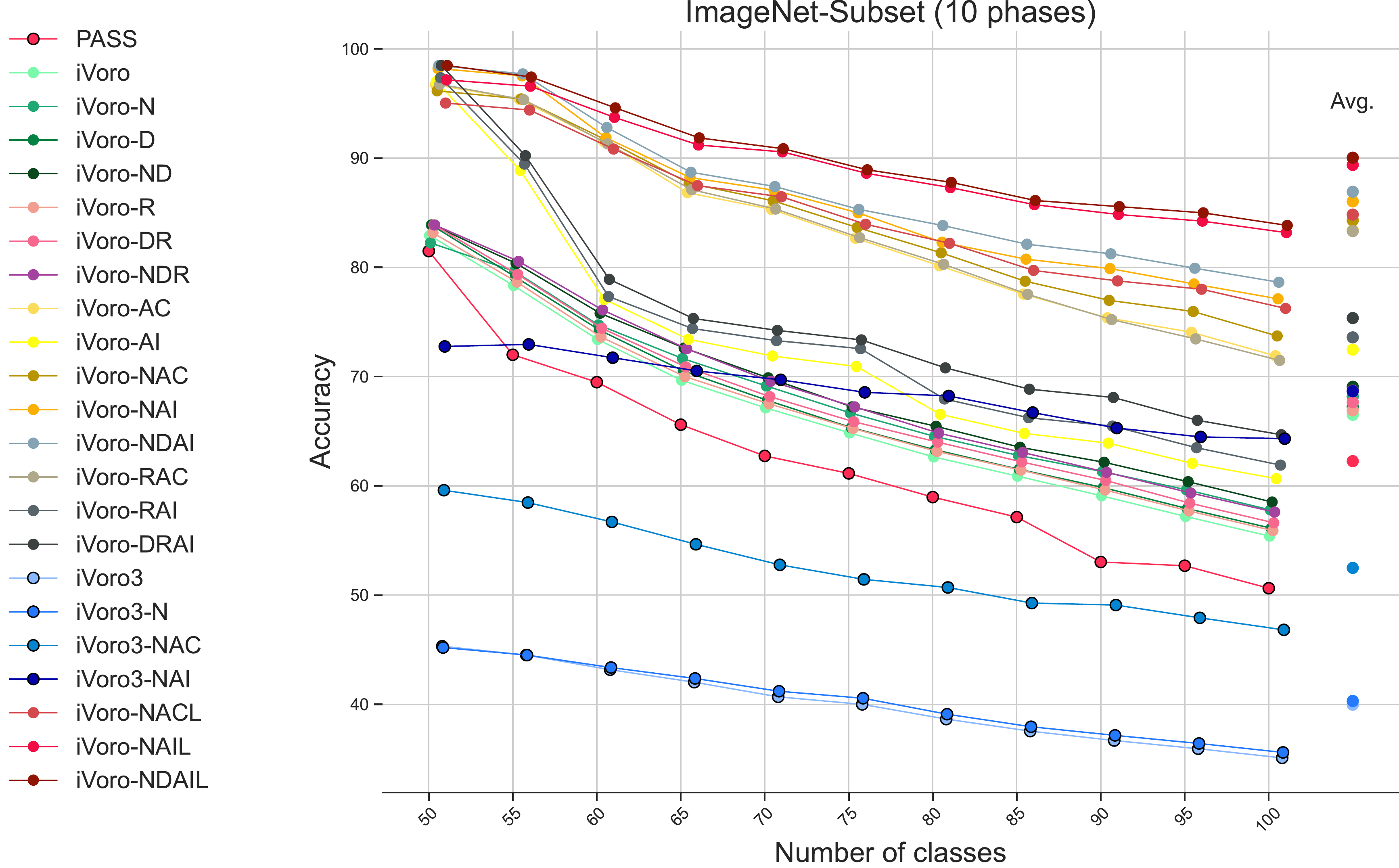}}
    \caption{Top-1 classification accuracy on ImageNet-Subset with all 18 ablation methods during 10 phases of CIL.}\label{fig:abl-subset}
\end{figure}

\begin{figure}[ht]
    \centering
    \subfloat{\includegraphics[width=0.8\textwidth]{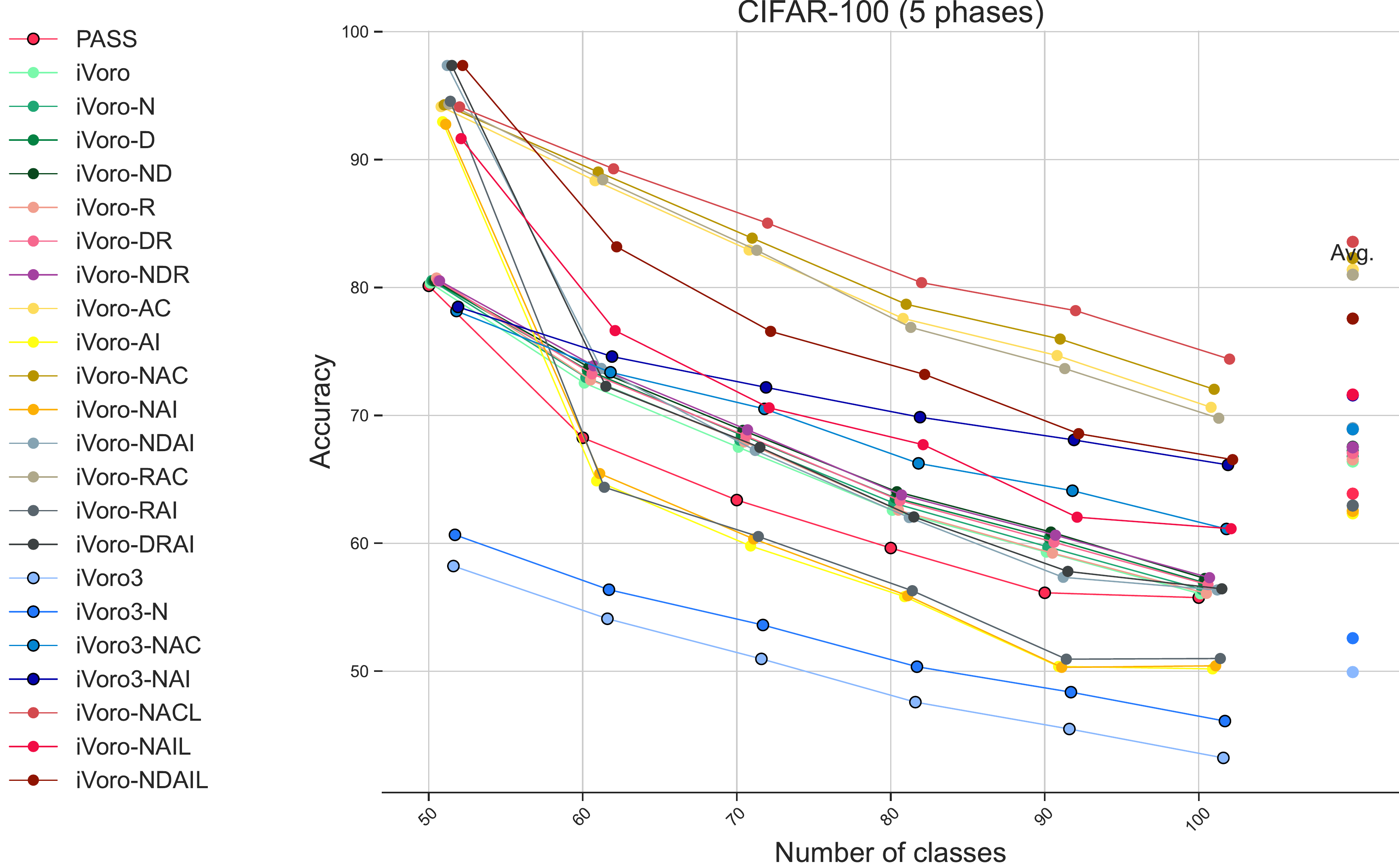}} \\
    \subfloat{\includegraphics[width=0.8\textwidth]{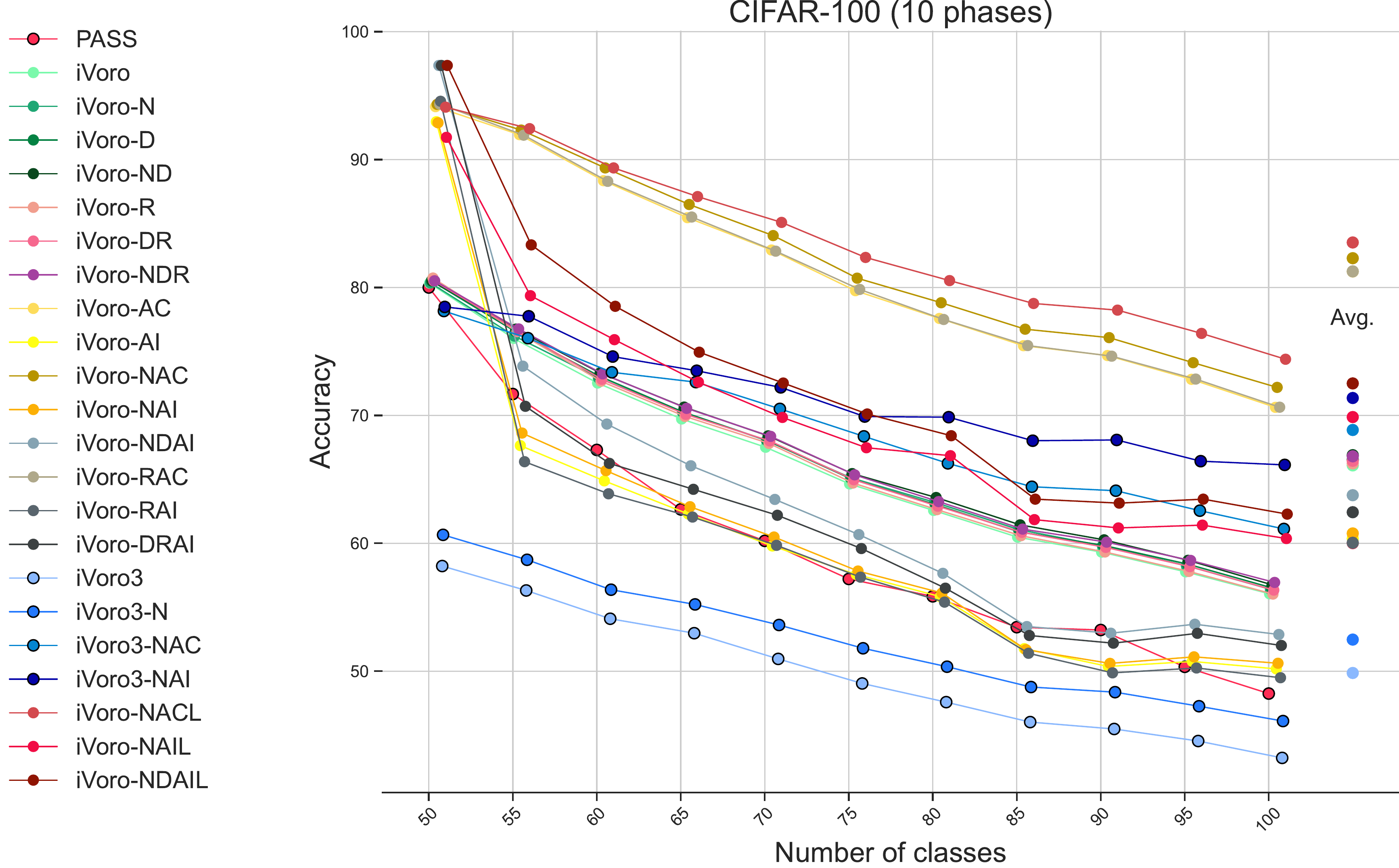}} \\
    \subfloat{\includegraphics[width=0.8\textwidth]{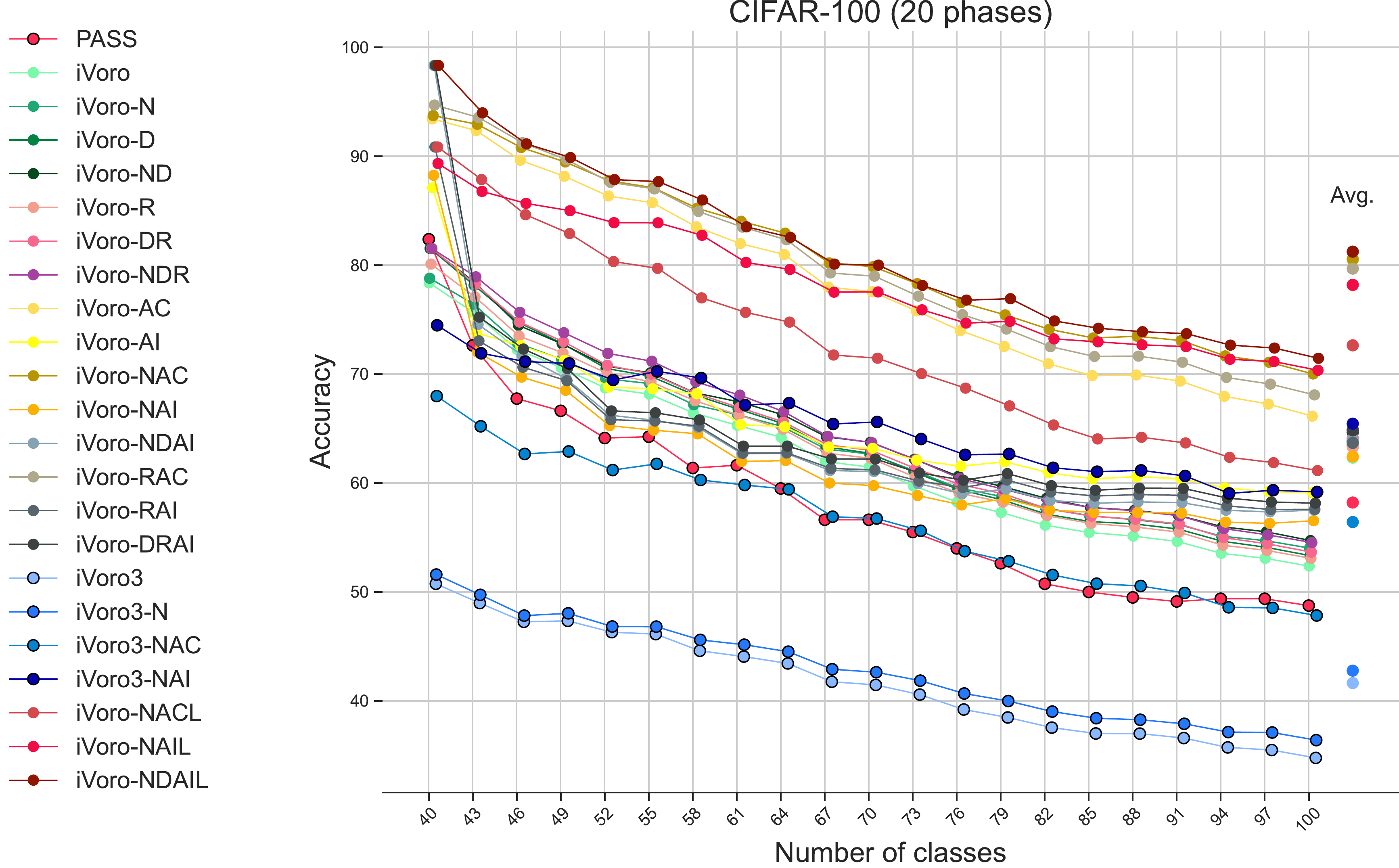}} \\ 
    \caption{Top-1 classification accuracy on CIFAR-100 with all 18 ablation methods during 5/10/20 phases of CIL.}\label{fig:abl-cifar}
\end{figure}
\begin{figure}[ht]
    \centering
    \subfloat{\includegraphics[width=0.8\textwidth]{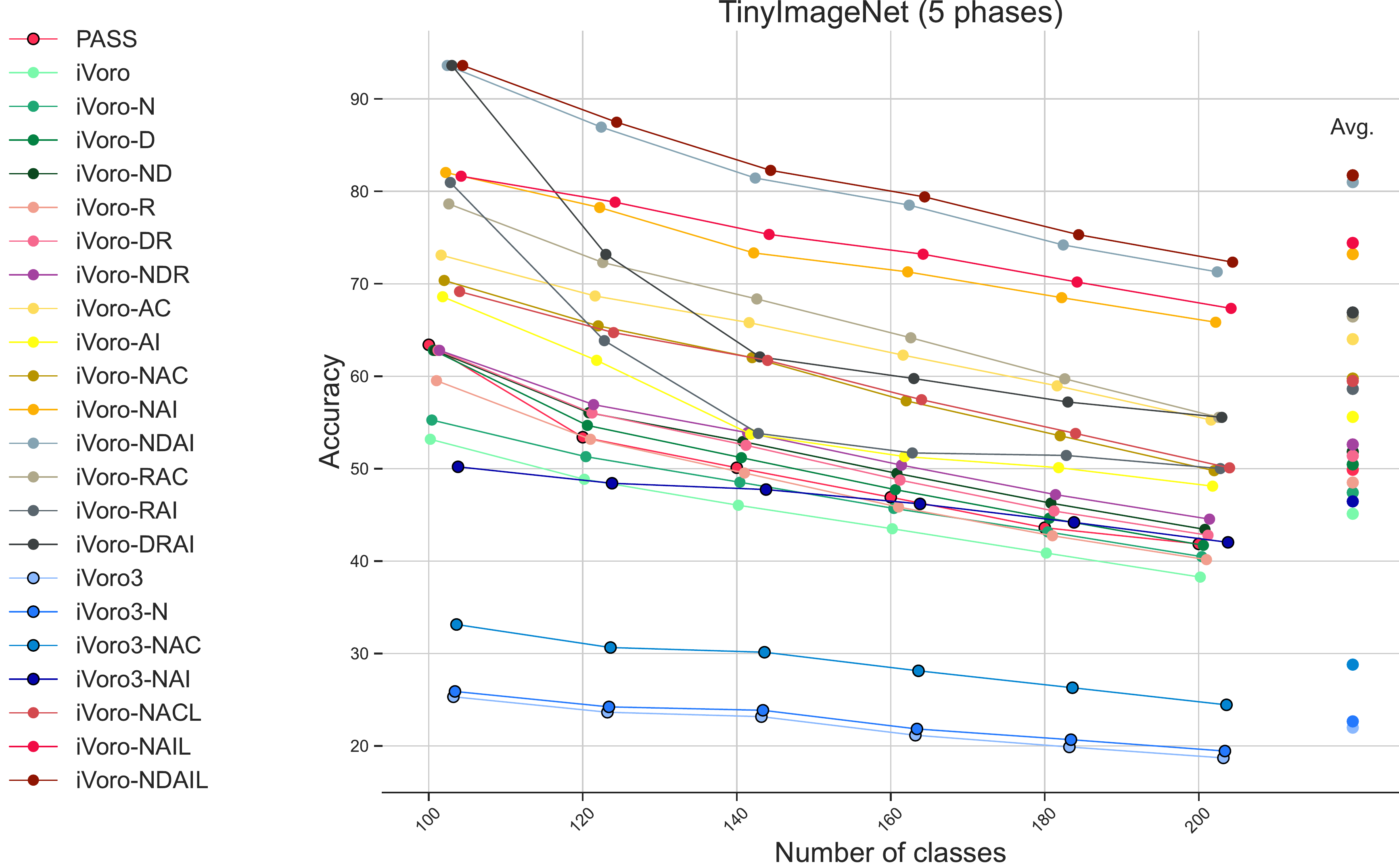}} \\
    \subfloat{\includegraphics[width=0.8\textwidth]{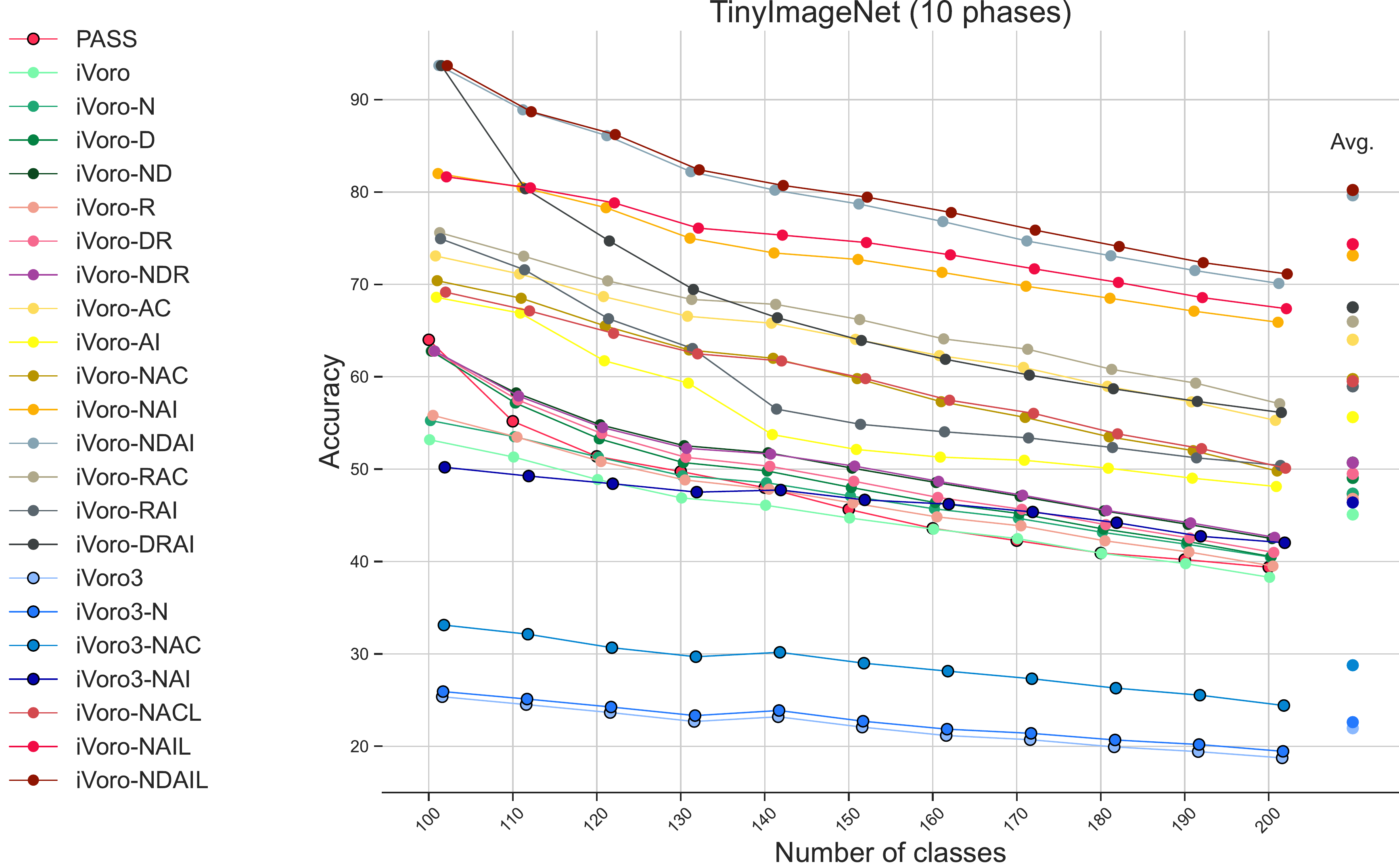}} \\
    \subfloat{\includegraphics[width=0.8\textwidth]{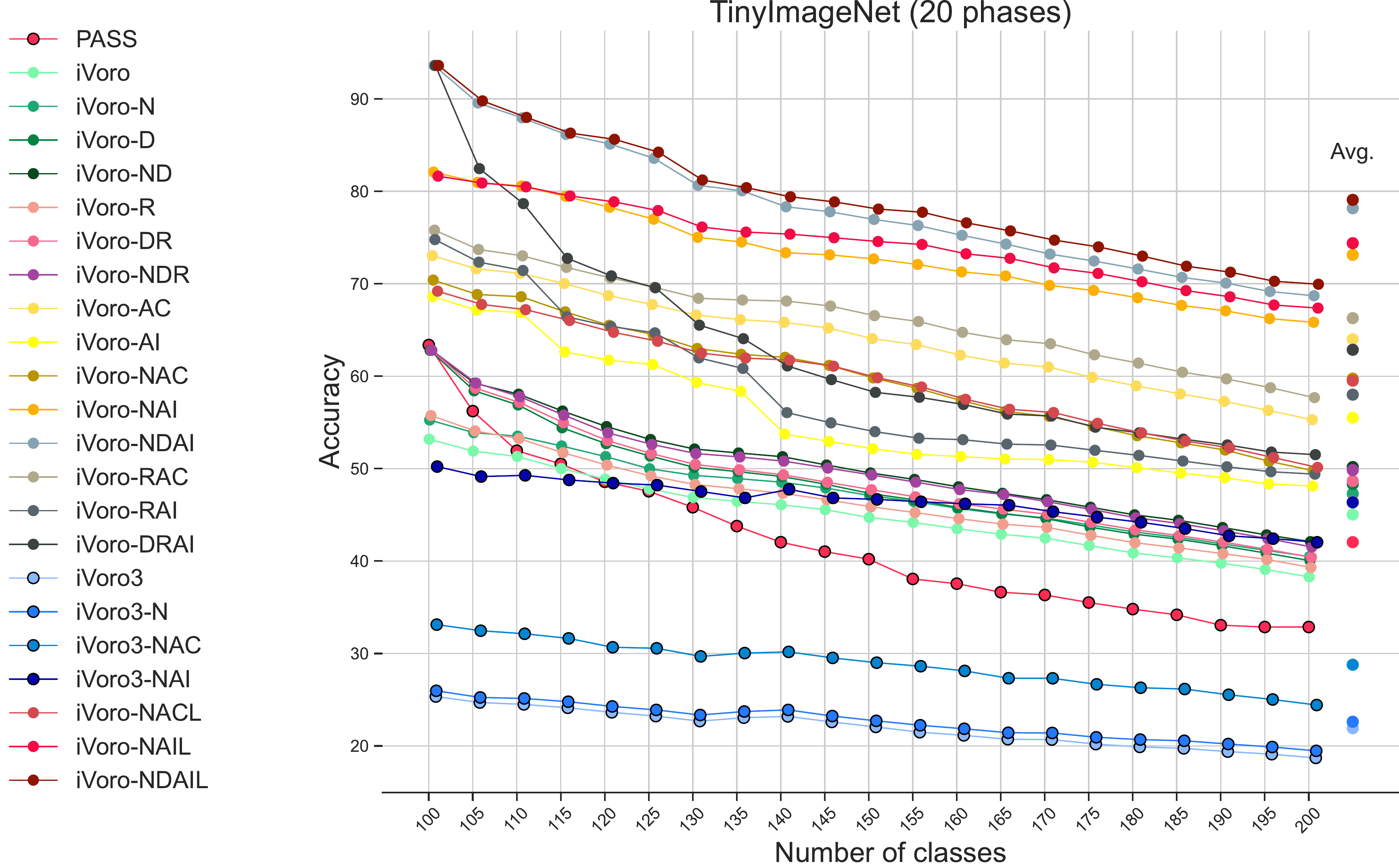}} \\ 
    \caption{Top-1 classification accuracy on TinyImageNet with all 18 ablation methods during 5/10/20 phases of CIL.}\label{fig:abl-tiny}
\end{figure}
% ==================== TSNE ====================
\clearpage
\section{t-SNE Visualization of Features from Various Methods}\label{supp:tsne}
\begin{figure}[!htp]
    \centering
    \subfloat[PASS]{\includegraphics[width=0.5\textwidth]{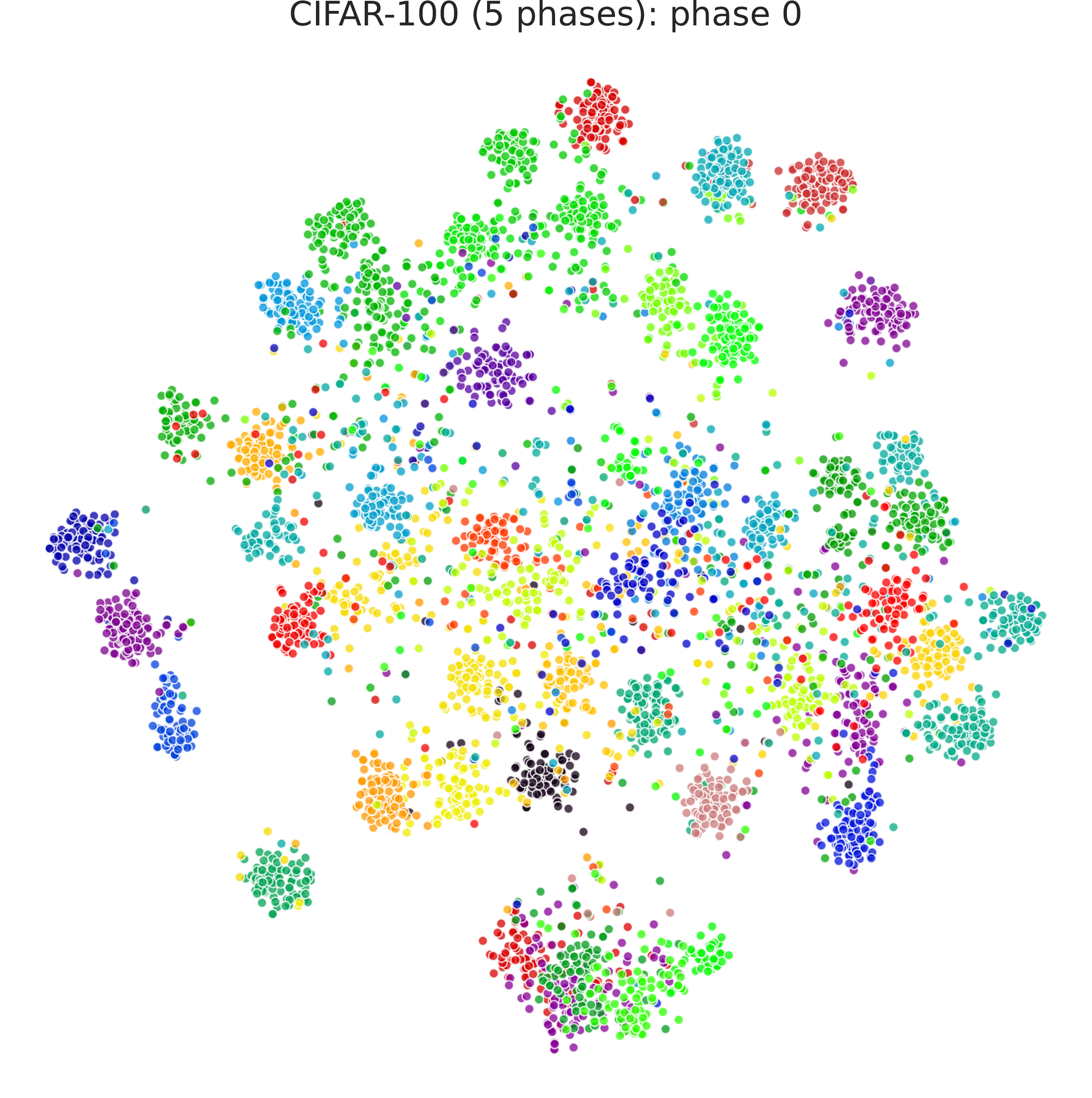}} 
    \subfloat[iVoro]{\includegraphics[width=0.5\textwidth]{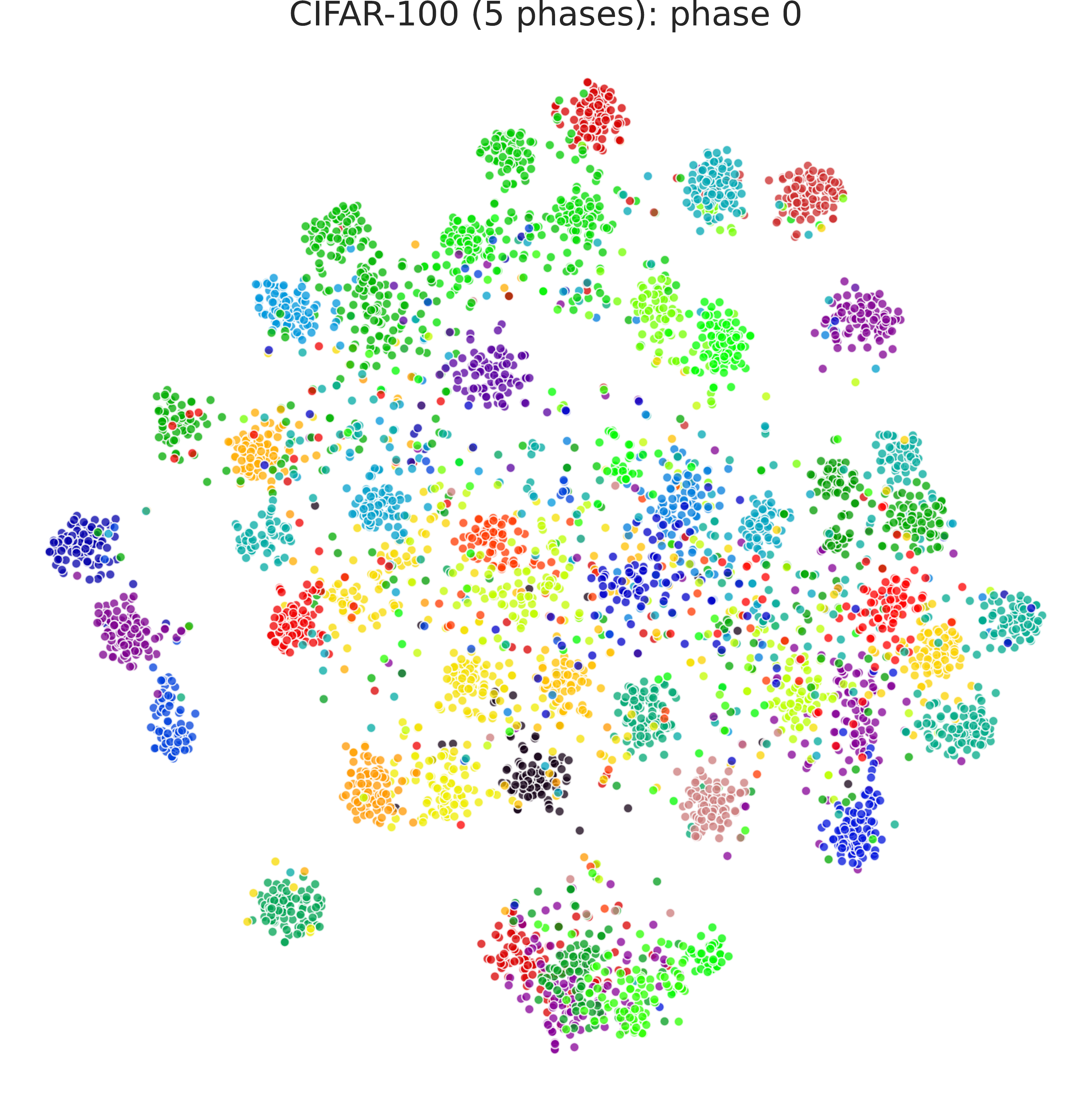}} \\ [-0.0ex]
    \subfloat[iVoro-N]{\includegraphics[width=0.5\textwidth]{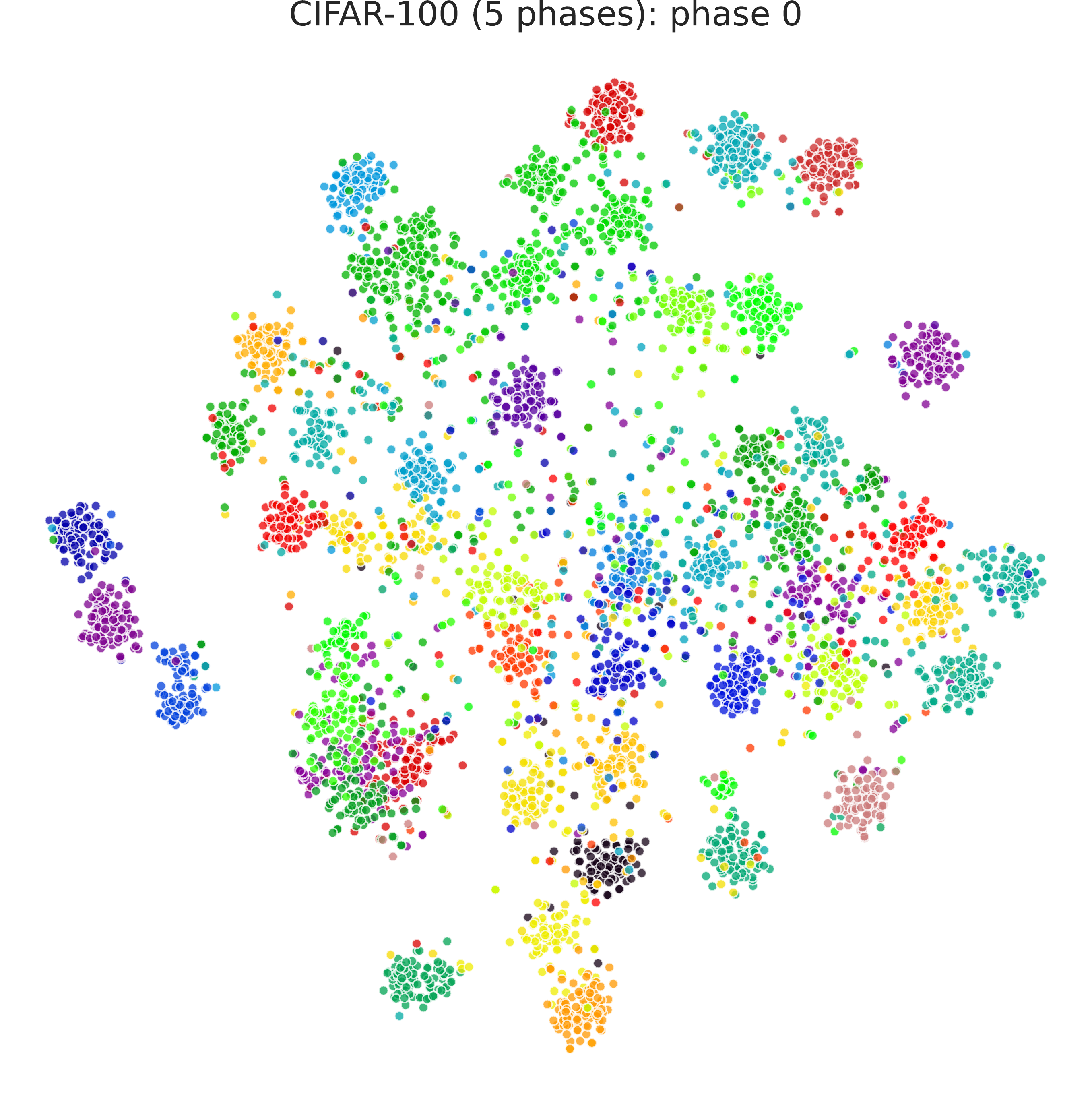}} 
    \subfloat[iVoro-NAC]{\includegraphics[width=0.5\textwidth]{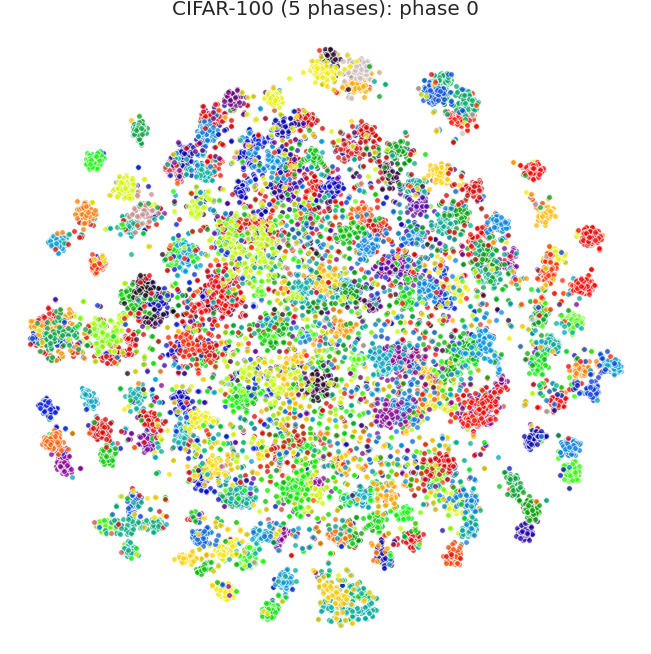}} \\ [-0.0ex]
    \caption{The t-SNE visualization of (a) PASS, (b) iVoro, (c) iVoro-N, and (d) iVoro-NAC on 5-phase CIFAR-100 dataset (phase 0).}\label{fig:tsne-0}
\end{figure}

\begin{figure}[!htp]
    \centering
    \subfloat[PASS]{\includegraphics[width=0.5\textwidth]{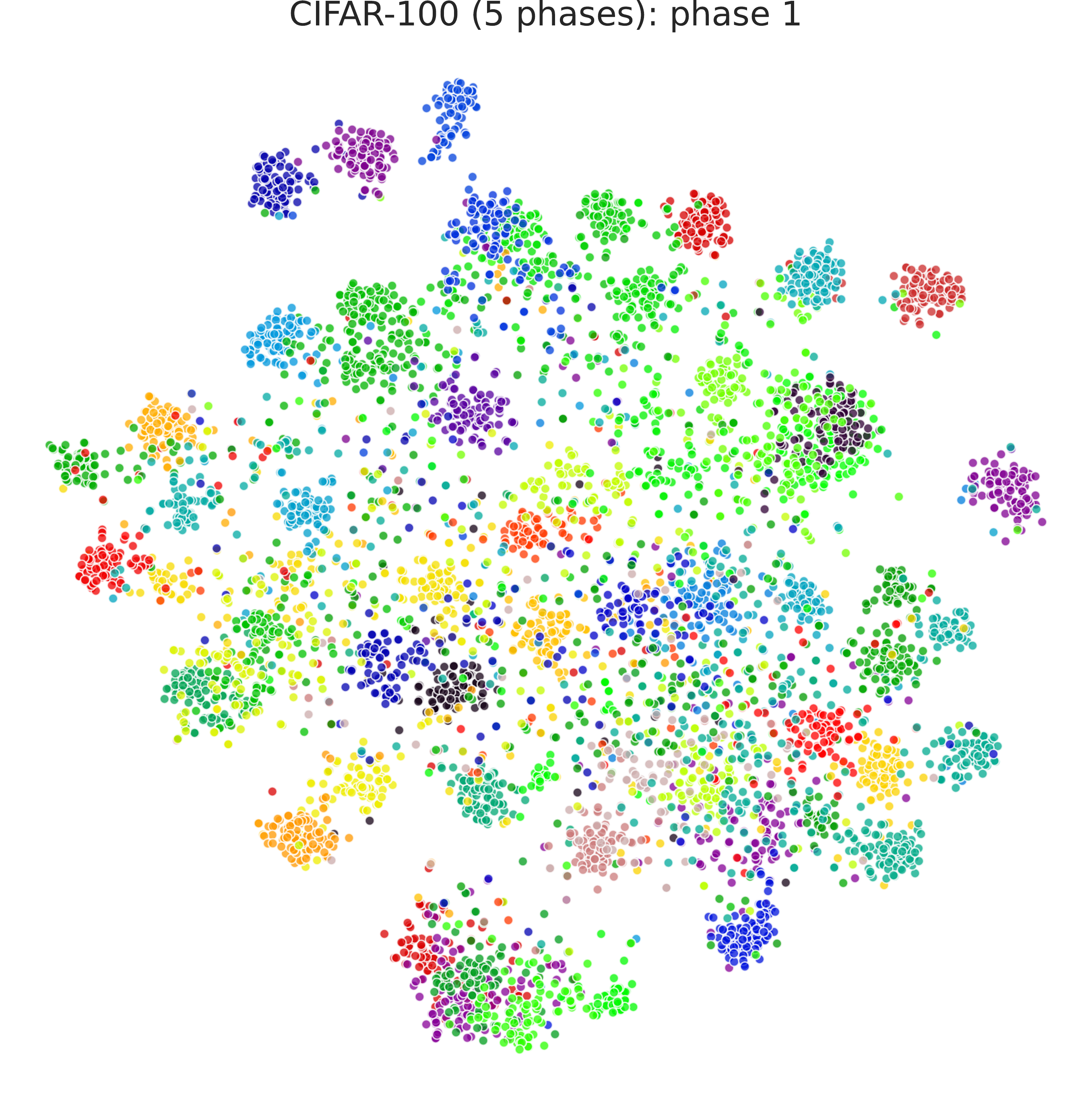}} 
    \subfloat[iVoro]{\includegraphics[width=0.5\textwidth]{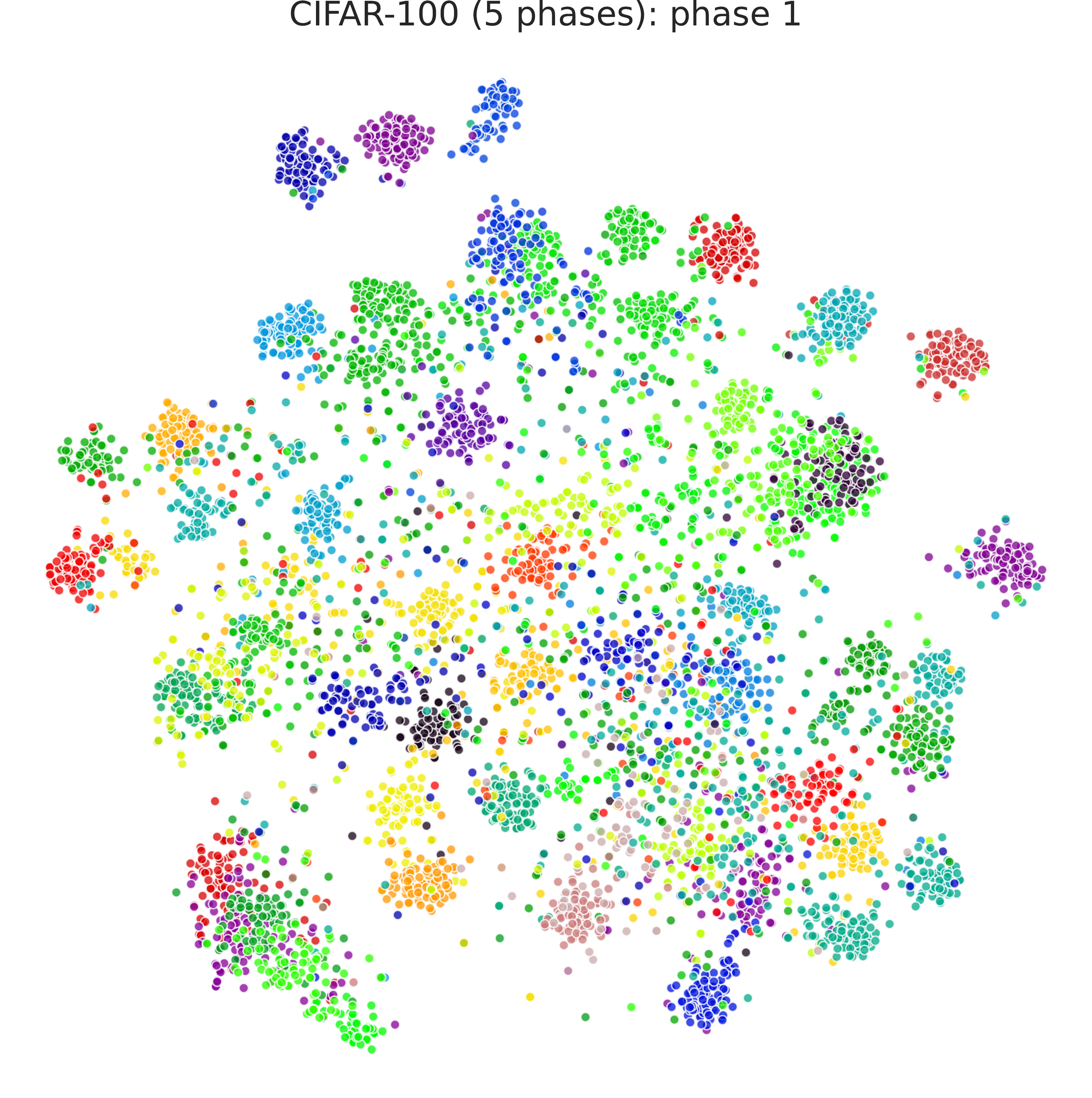}} \\ [-0.0ex]
    \subfloat[iVoro-N]{\includegraphics[width=0.5\textwidth]{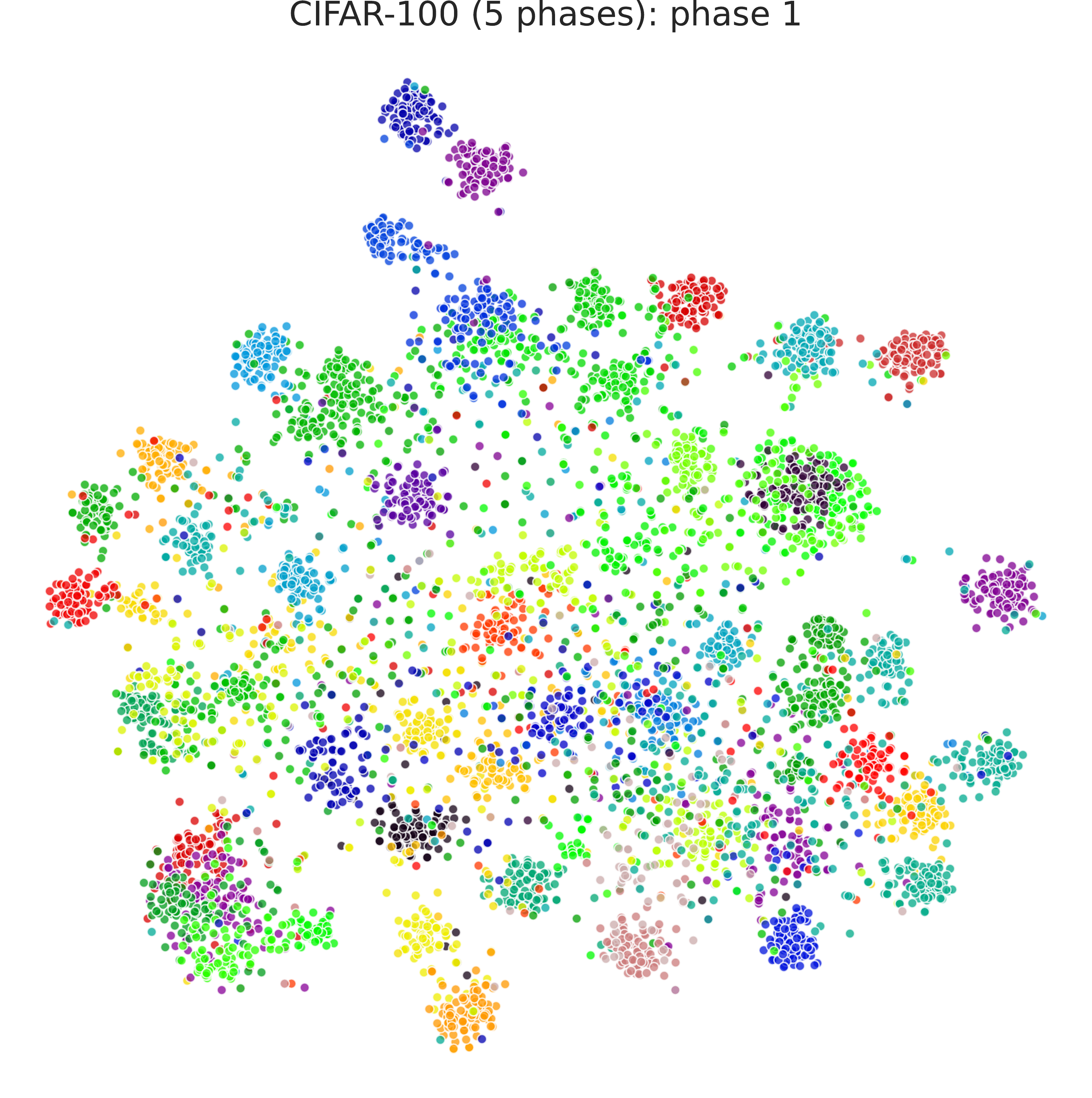}} 
    \subfloat[iVoro-NAC]{\includegraphics[width=0.5\textwidth]{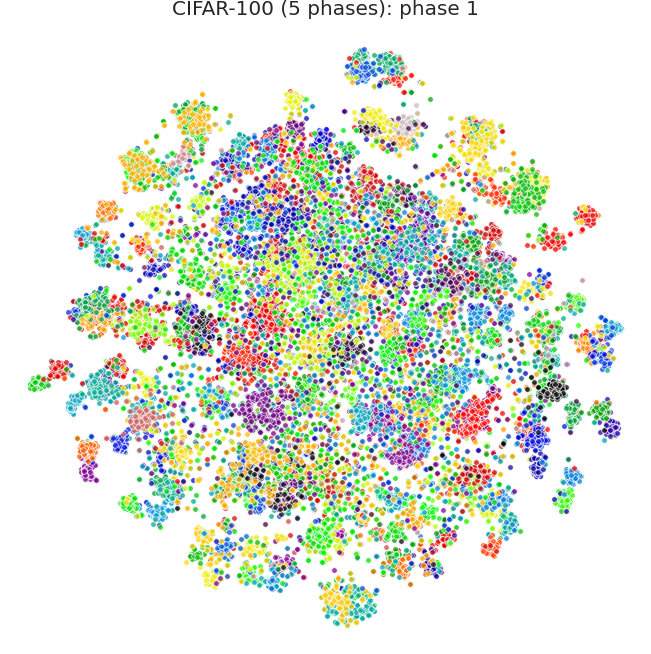}} \\ [-0.0ex]
    \caption{The t-SNE visualization of (a) PASS, (b) iVoro, (c) iVoro-N, and (d) iVoro-NAC on 5-phase CIFAR-100 dataset (phase 1).}\label{fig:tsne-1}
\end{figure}

\begin{figure}[!htp]
    \centering
    \subfloat[PASS]{\includegraphics[width=0.5\textwidth]{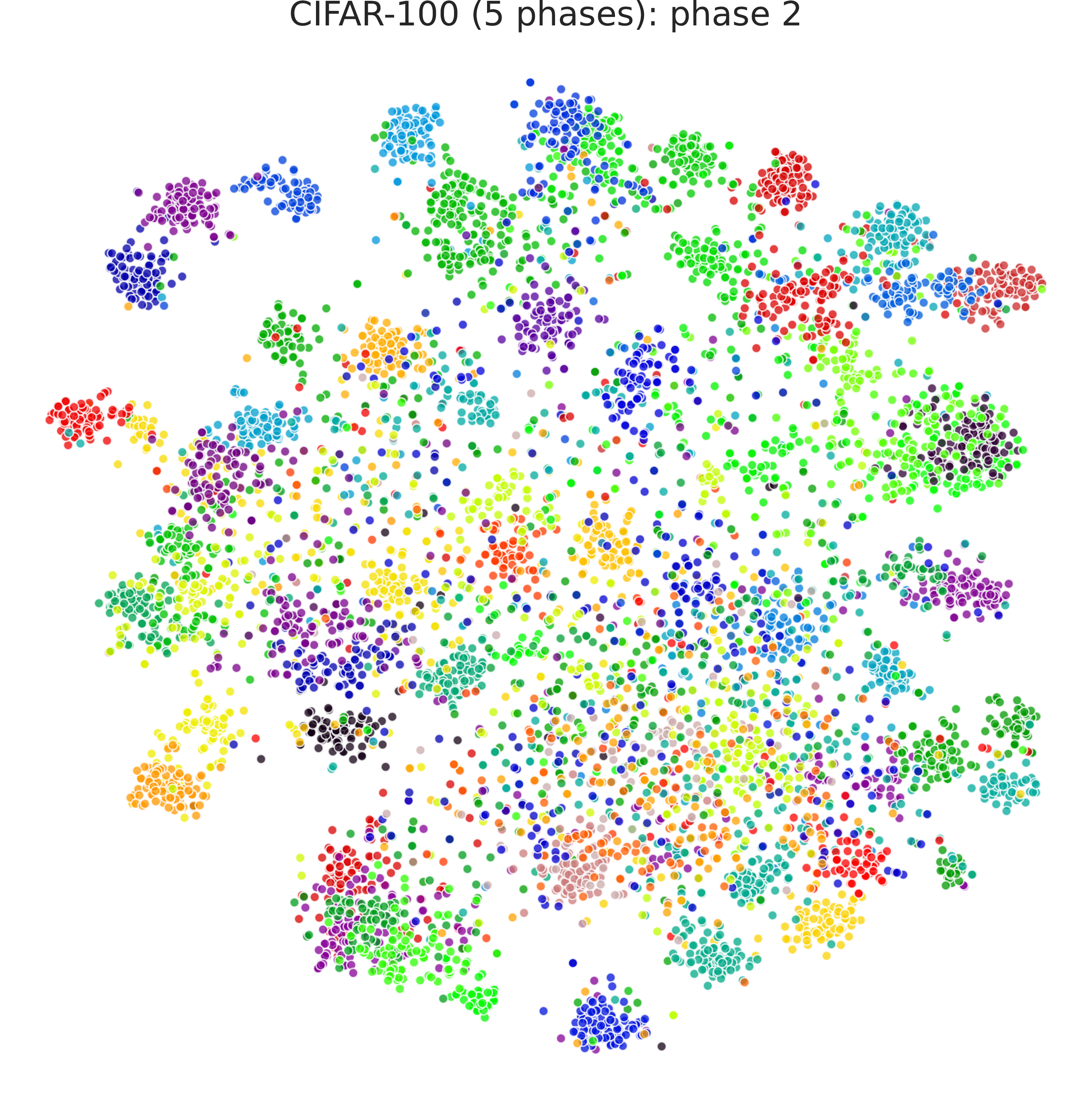}} 
    \subfloat[iVoro]{\includegraphics[width=0.5\textwidth]{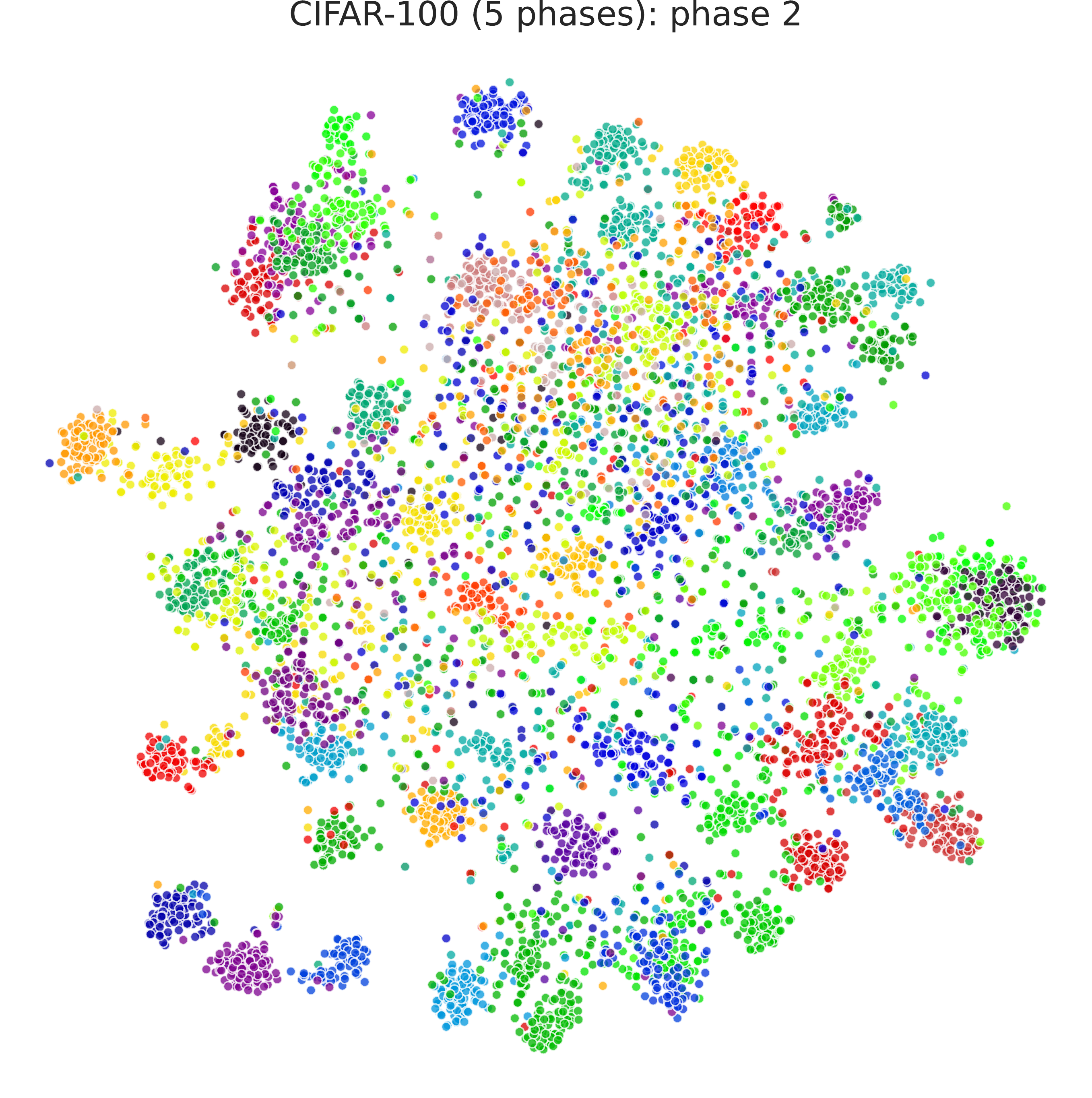}} \\ [-0.0ex]
    \subfloat[iVoro-N]{\includegraphics[width=0.5\textwidth]{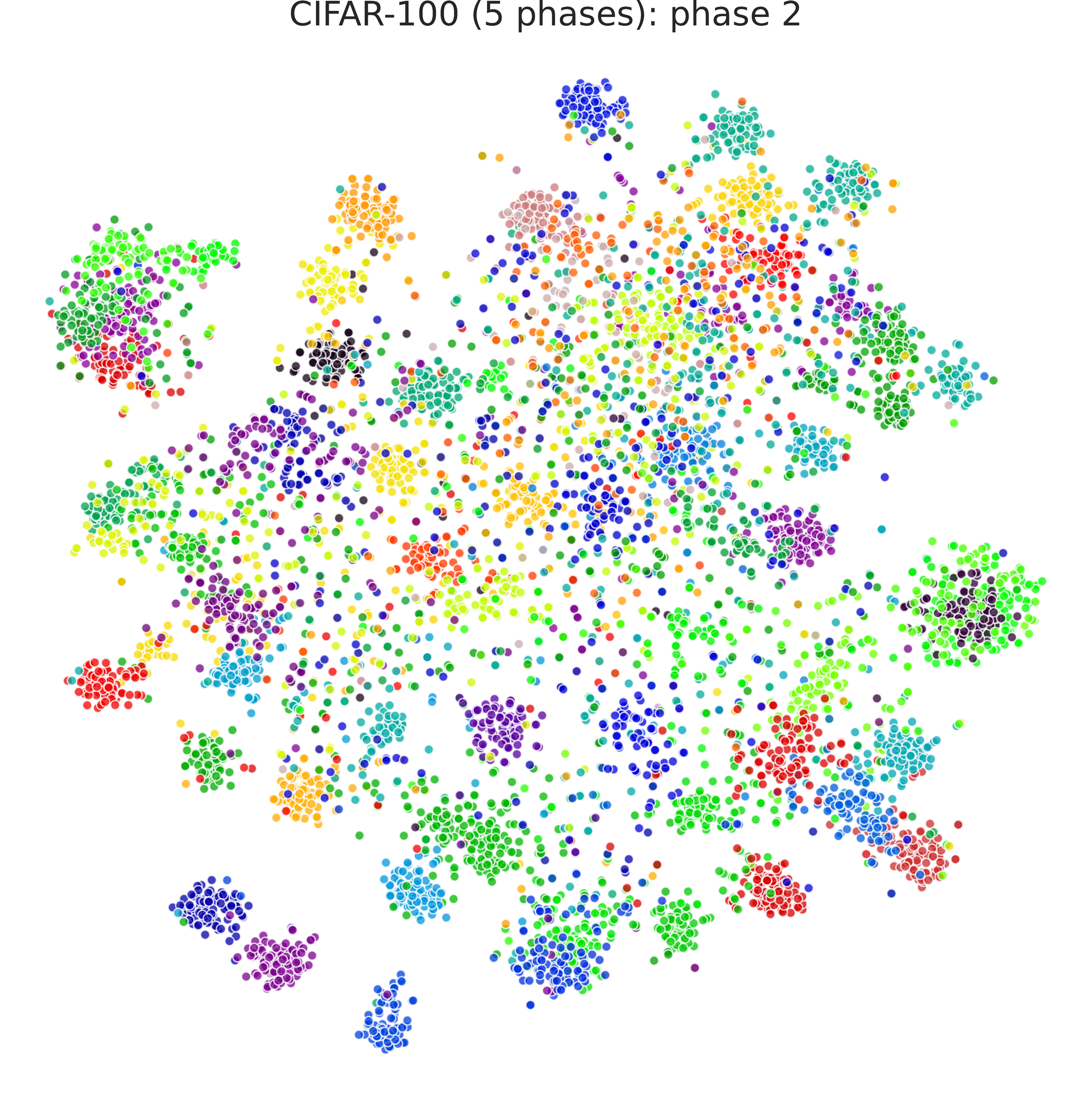}} 
    \subfloat[iVoro-NAC]{\includegraphics[width=0.5\textwidth]{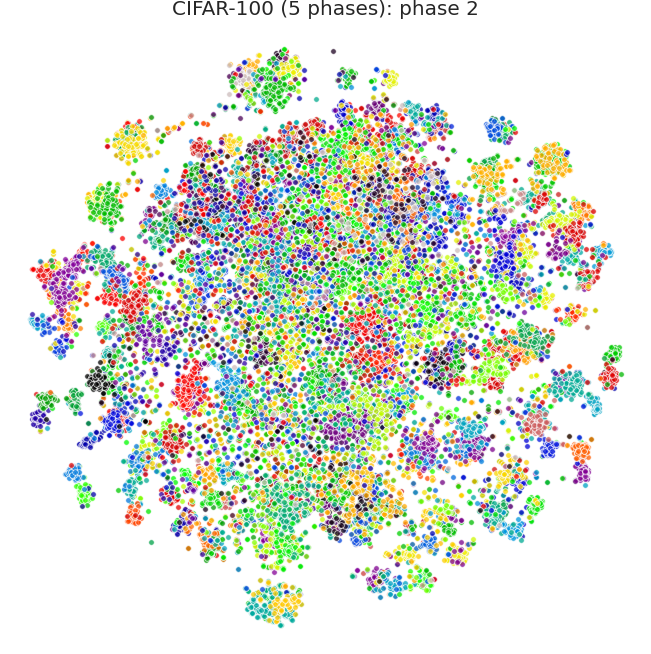}} \\ [-0.0ex]
    \caption{The t-SNE visualization of (a) PASS, (b) iVoro, (c) iVoro-N, and (d) iVoro-NAC on 5-phase CIFAR-100 dataset (phase 2).}\label{fig:tsne-2}
\end{figure}

\begin{figure}[!htp]
    \centering
    \subfloat[PASS]{\includegraphics[width=0.5\textwidth]{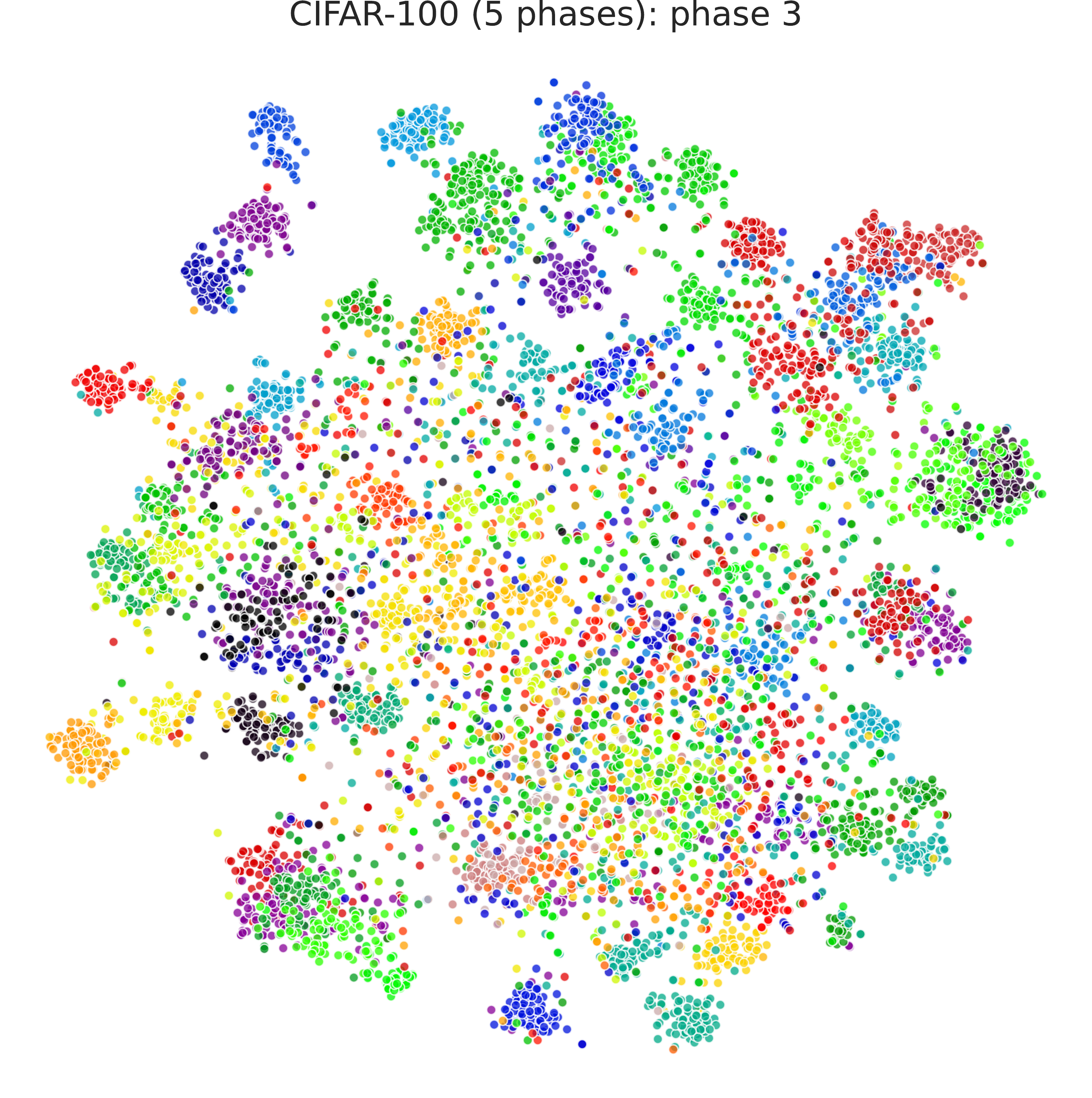}} 
    \subfloat[iVoro]{\includegraphics[width=0.5\textwidth]{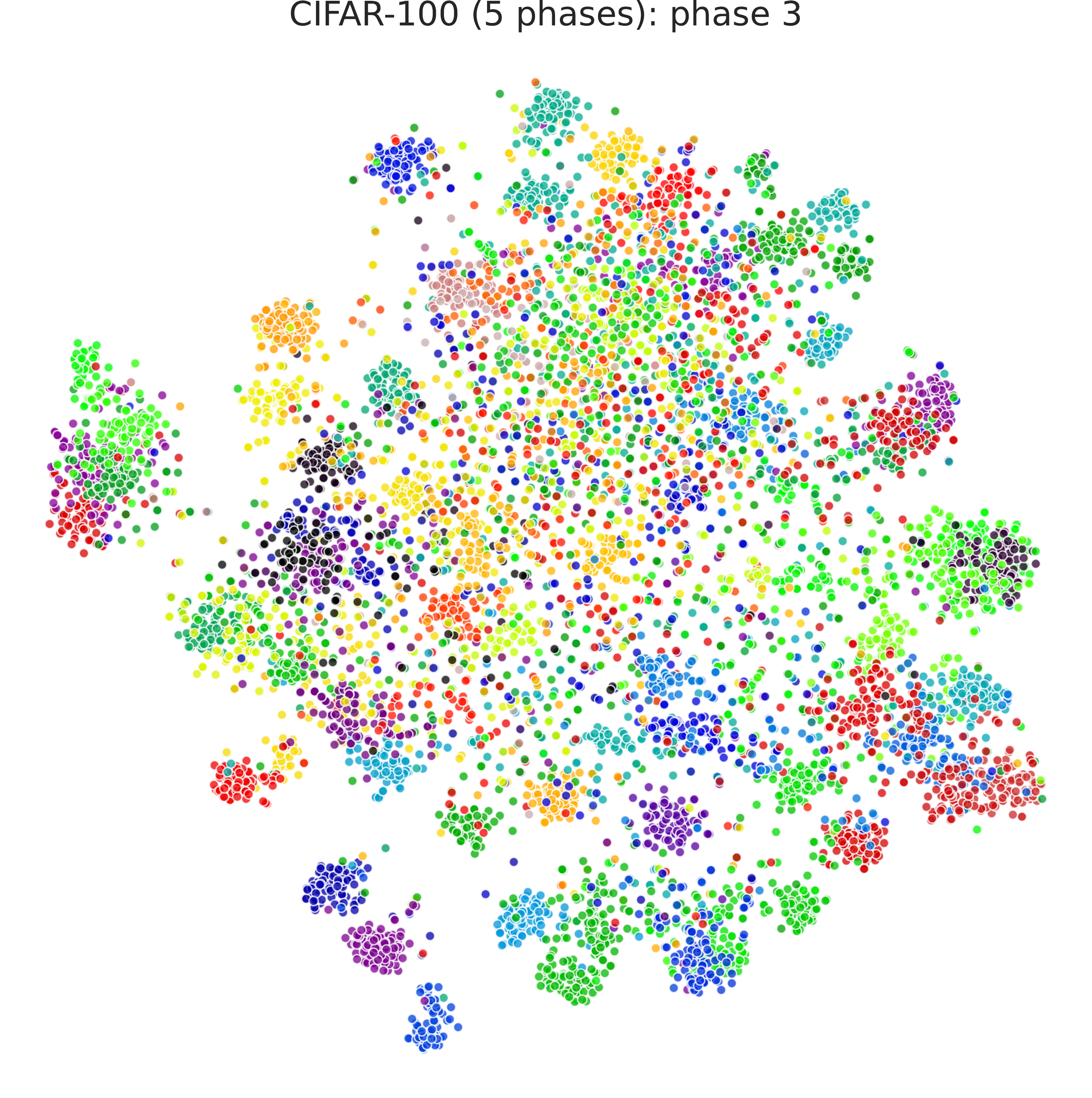}} \\ [-0.0ex]
    \subfloat[iVoro-N]{\includegraphics[width=0.5\textwidth]{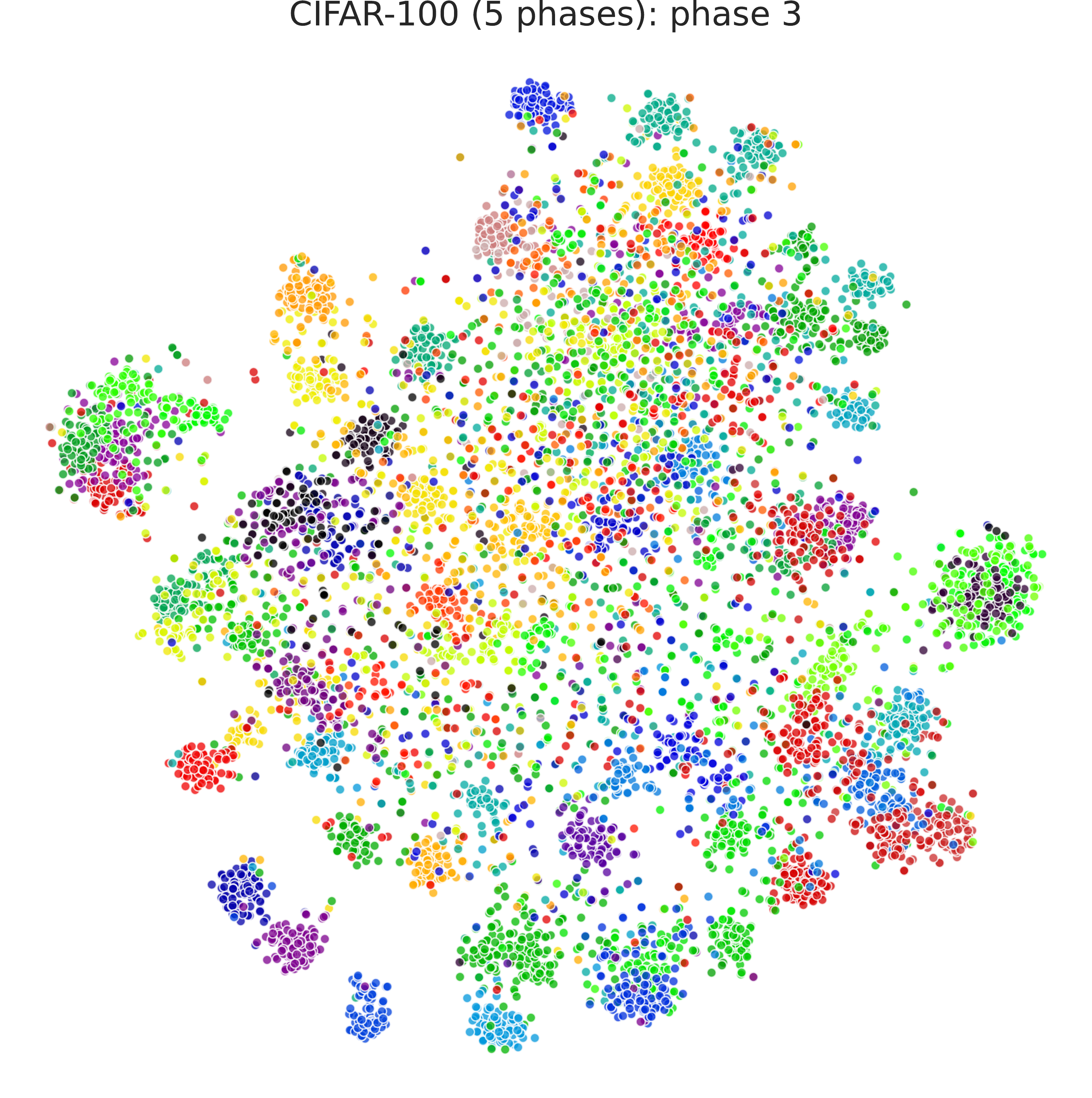}} 
    \subfloat[iVoro-NAC]{\includegraphics[width=0.5\textwidth]{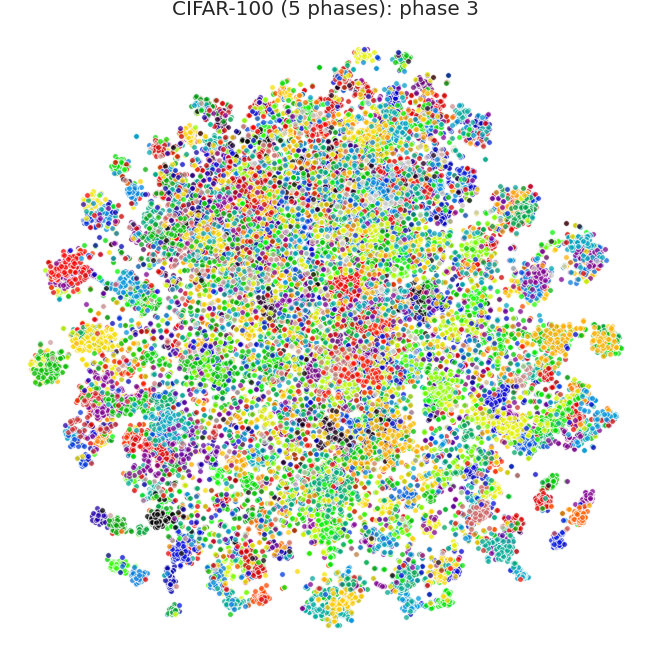}} \\ [-0.0ex]
    \caption{The t-SNE visualization of (a) PASS, (b) iVoro, (c) iVoro-N, and (d) iVoro-NAC on 5-phase CIFAR-100 dataset (phase 3).}\label{fig:tsne-3}
\end{figure}

\begin{figure}[!htp]
    \centering
    \subfloat[PASS]{\includegraphics[width=0.5\textwidth]{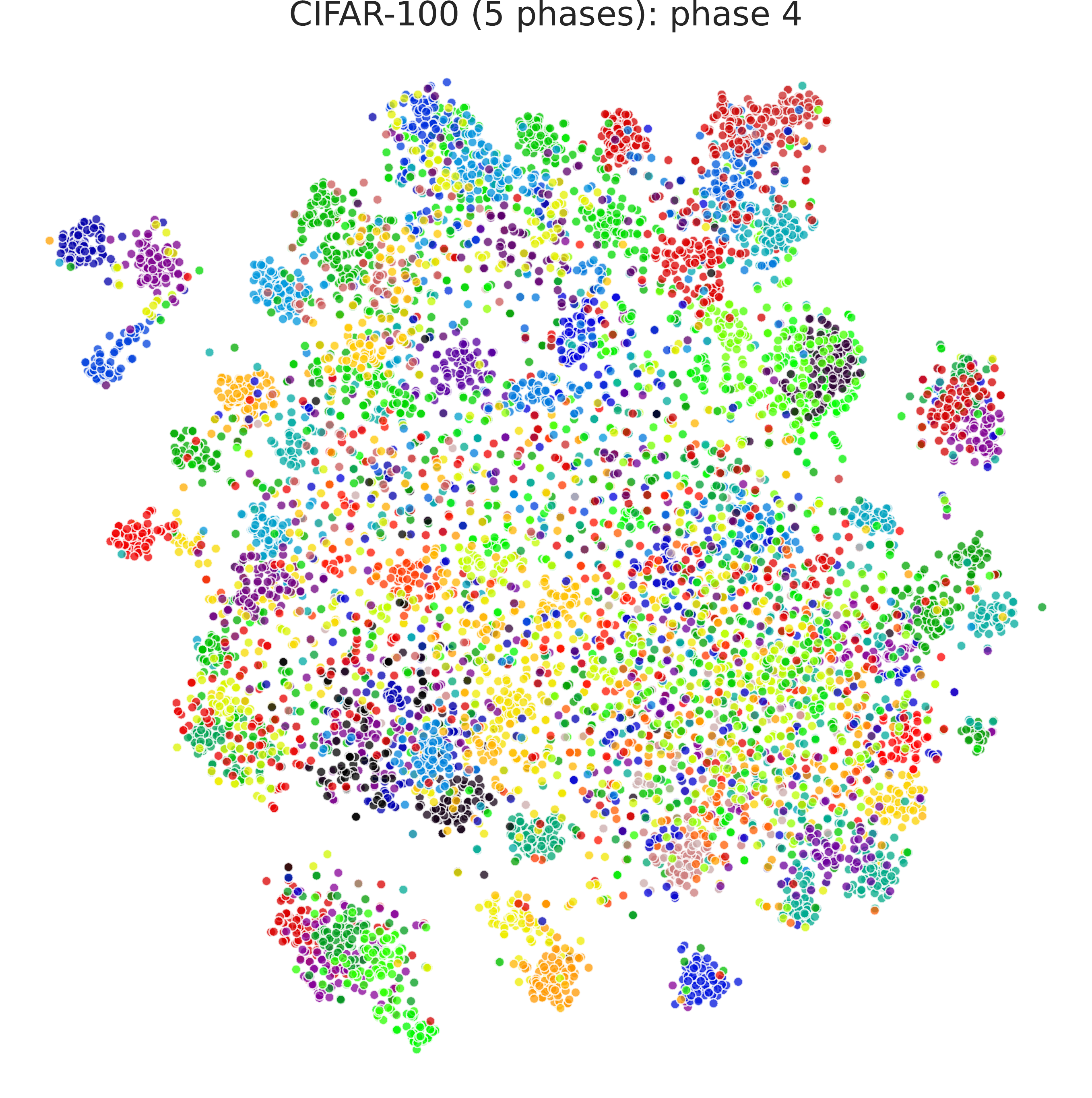}} 
    \subfloat[iVoro]{\includegraphics[width=0.5\textwidth]{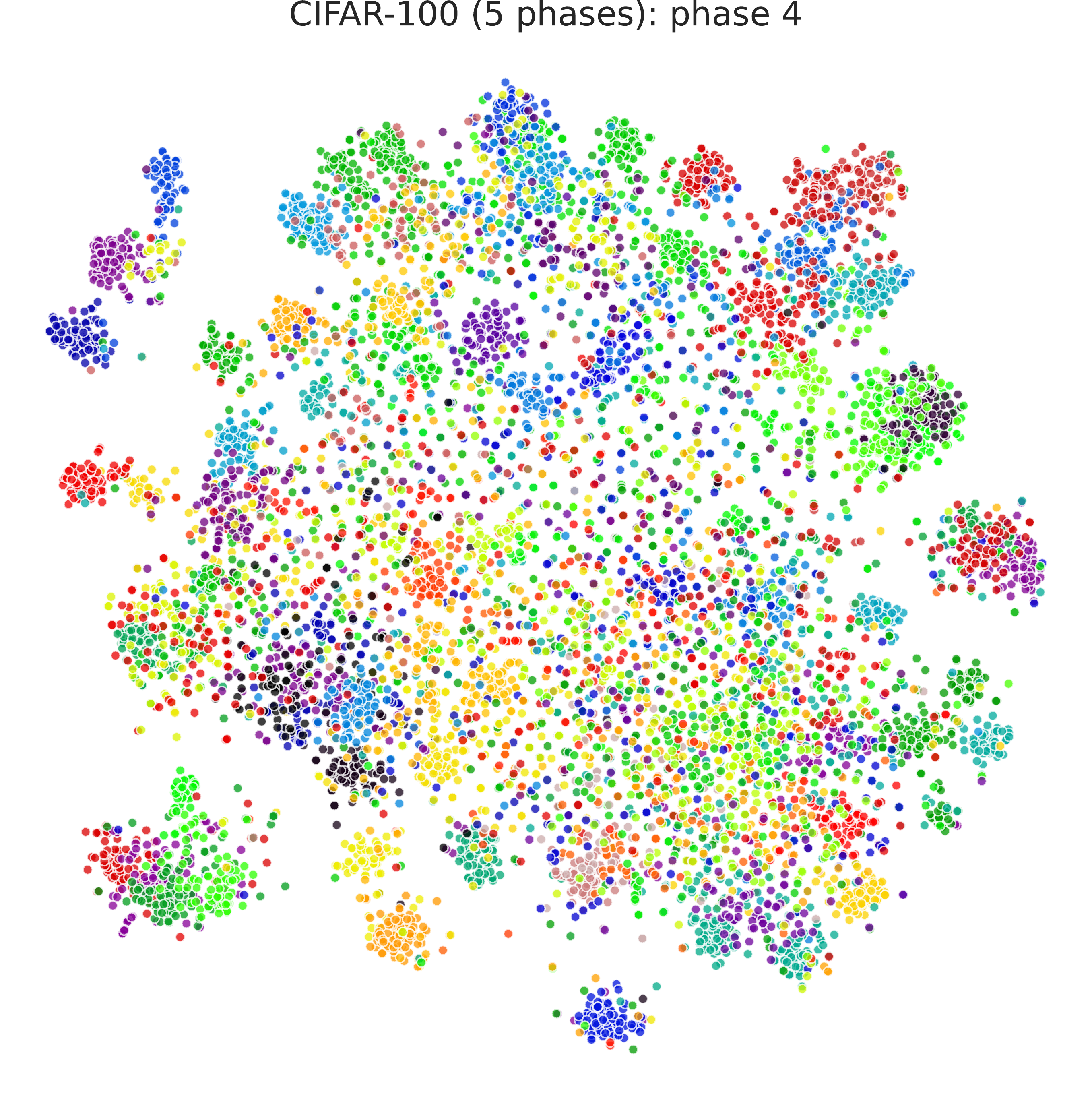}} \\ [-0.0ex]
    \subfloat[iVoro-N]{\includegraphics[width=0.5\textwidth]{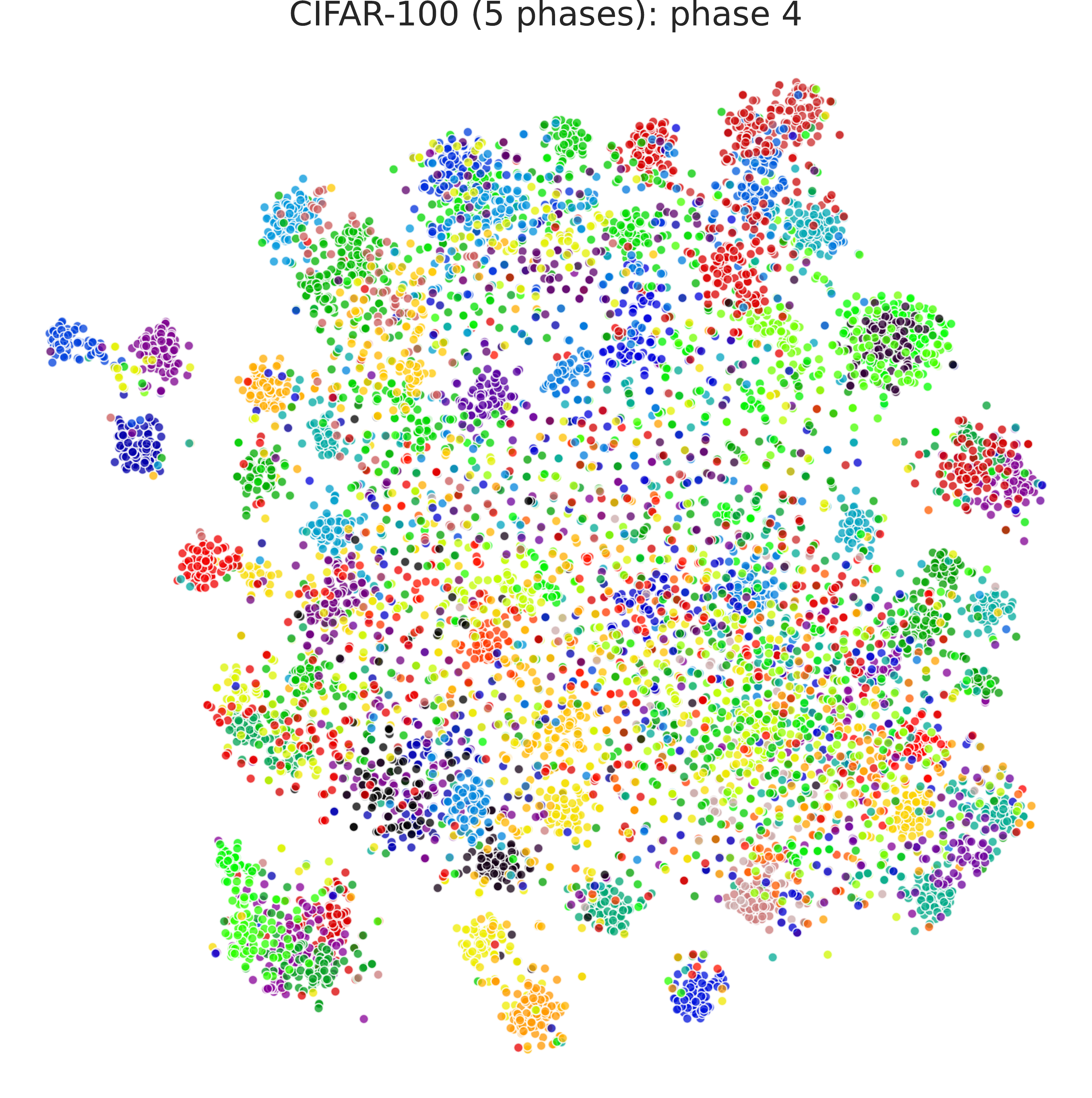}} 
    \subfloat[iVoro-NAC]{\includegraphics[width=0.5\textwidth]{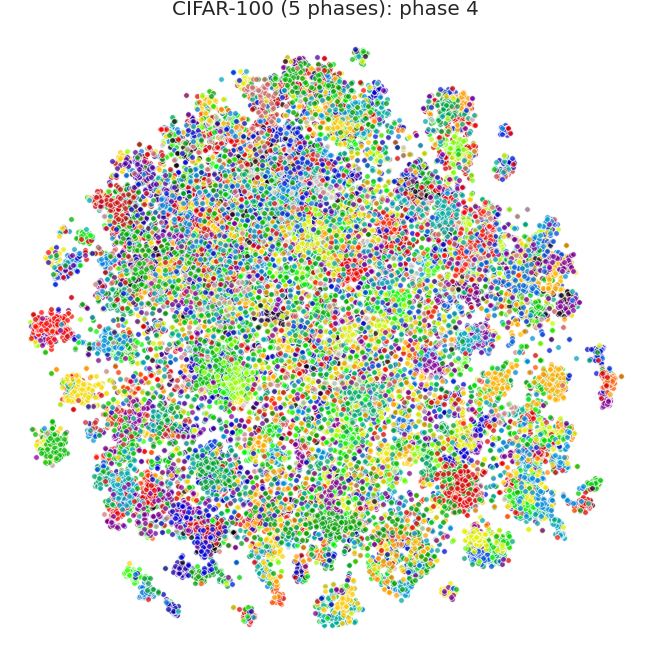}} \\ [-0.0ex]
    \caption{The t-SNE visualization of (a) PASS, (b) iVoro, (c) iVoro-N, and (d) iVoro-NAC on 5-phase CIFAR-100 dataset (phase 4).}\label{fig:tsne-4}
\end{figure}

\begin{figure}[!htp]
    \centering
    \subfloat[PASS]{\includegraphics[width=0.5\textwidth]{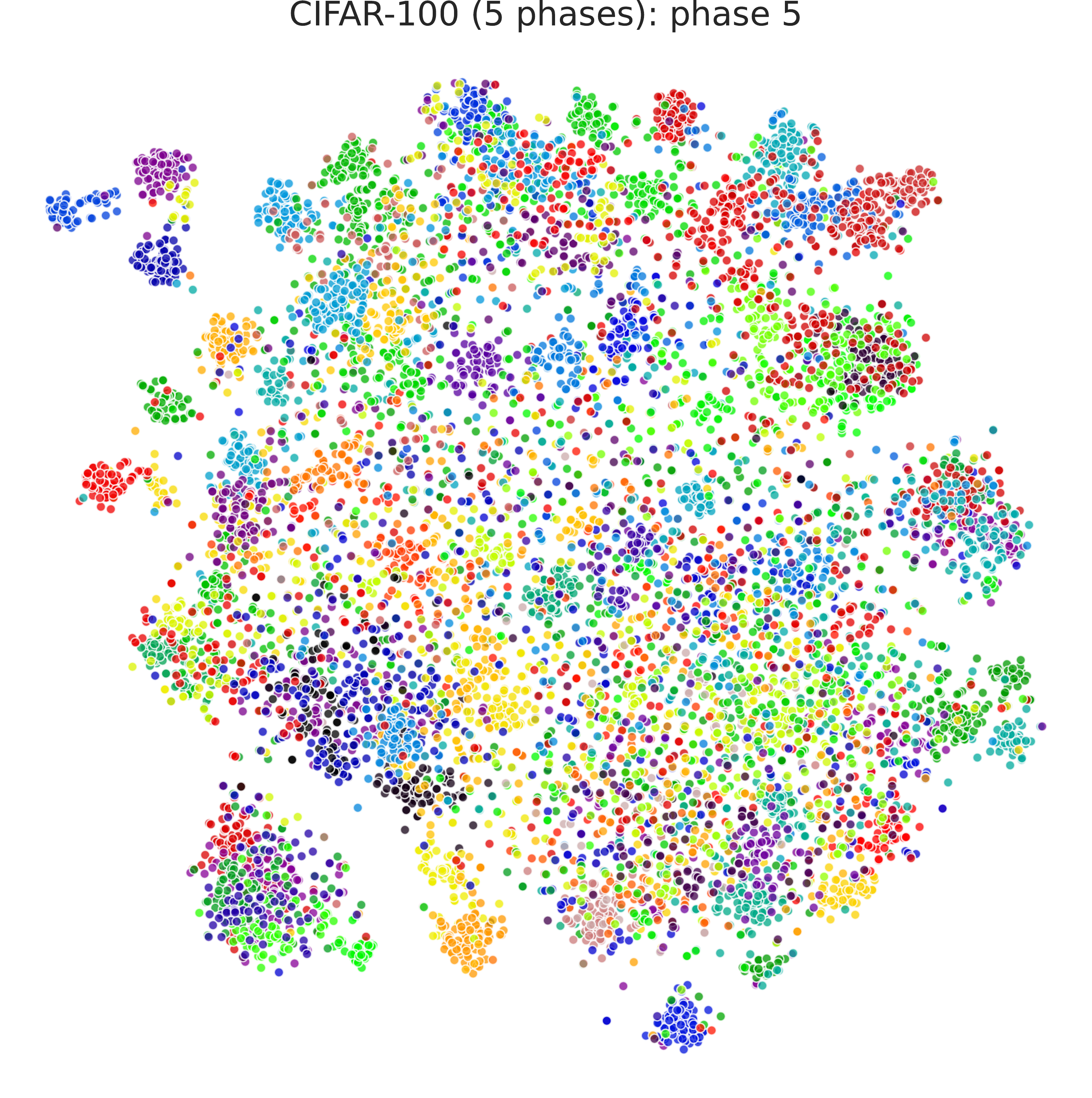}} 
    \subfloat[iVoro]{\includegraphics[width=0.5\textwidth]{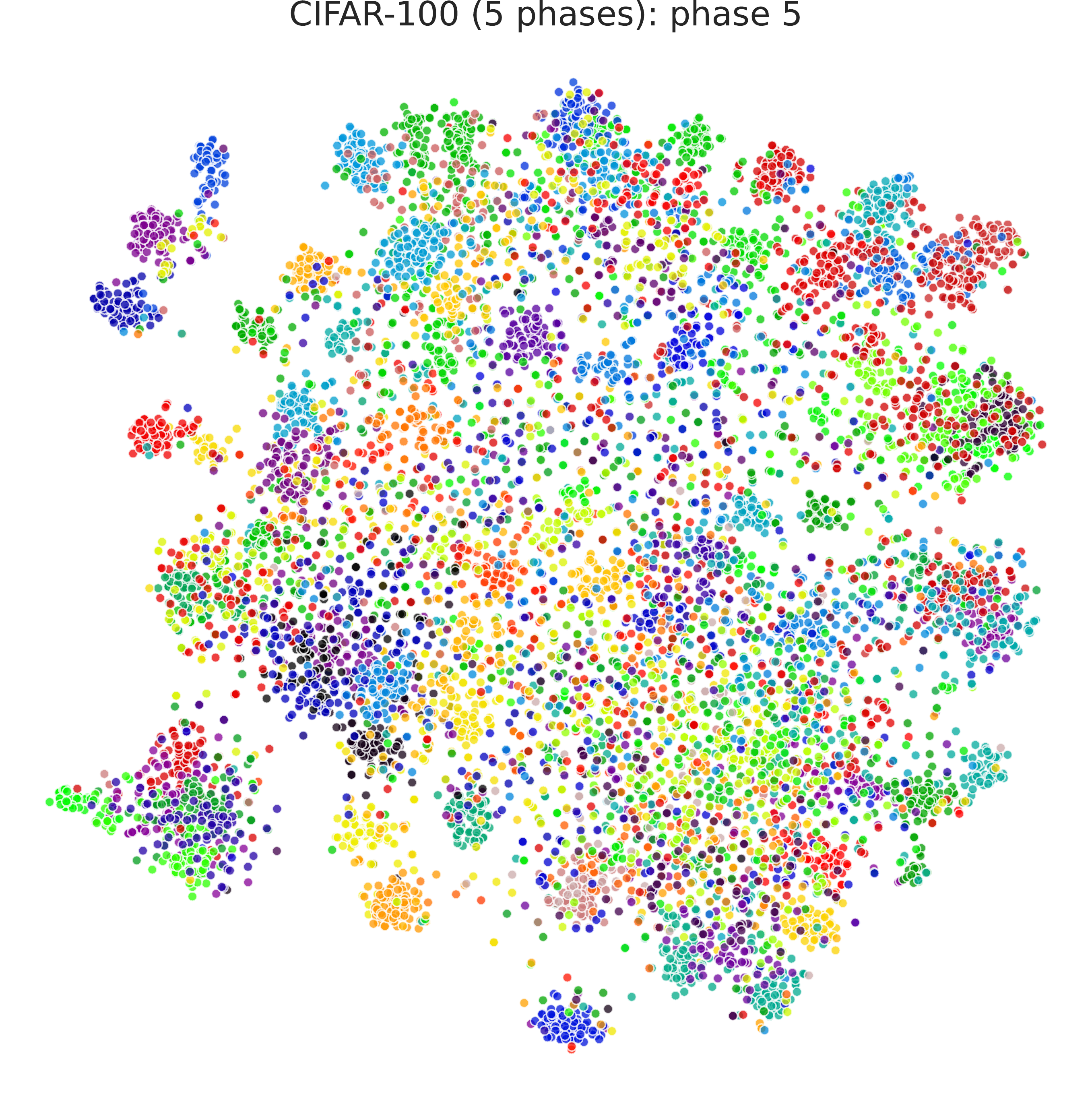}} \\ [-0.0ex]
    \subfloat[iVoro-N]{\includegraphics[width=0.5\textwidth]{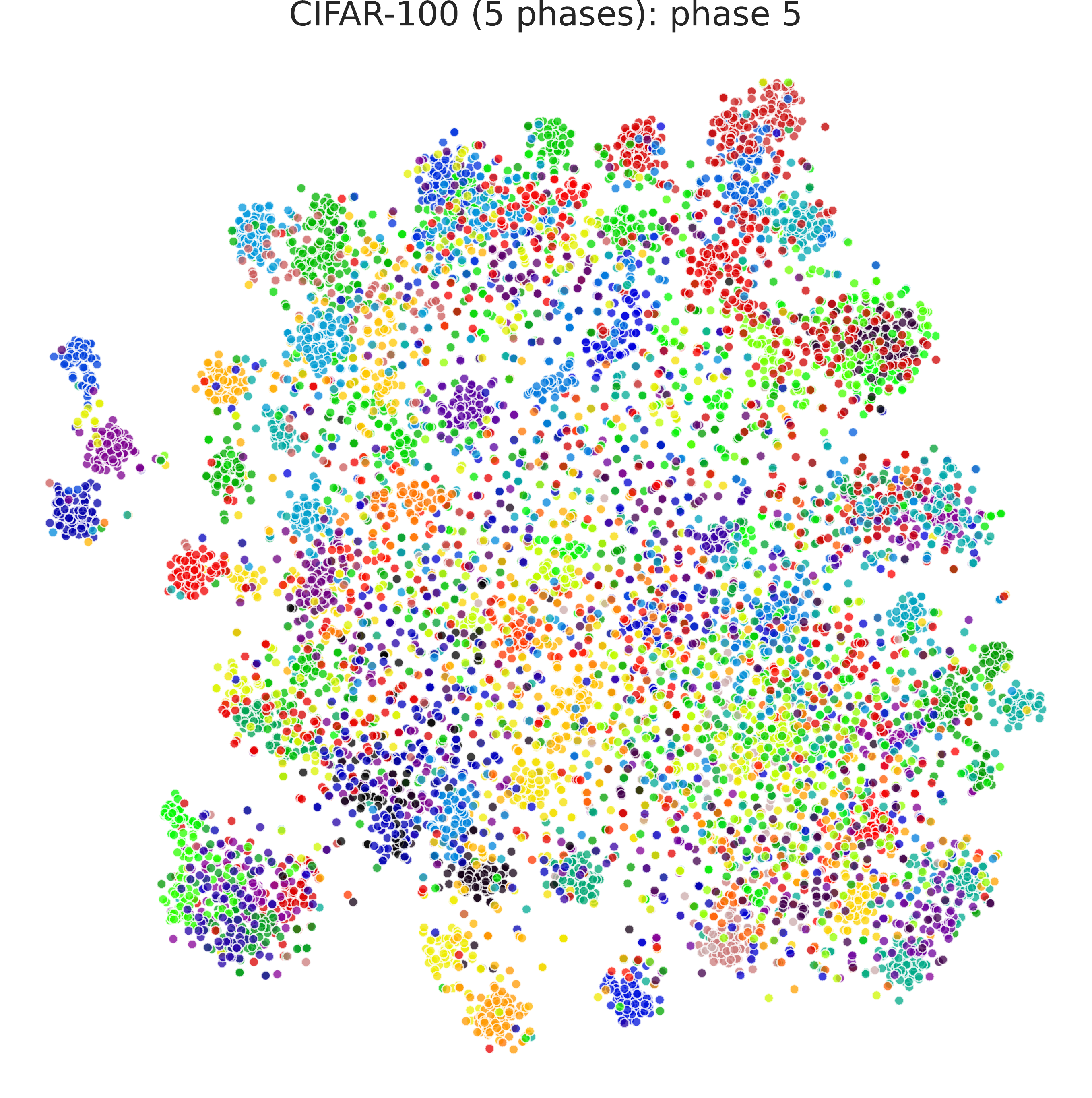}} 
    \subfloat[iVoro-NAC]{\includegraphics[width=0.5\textwidth]{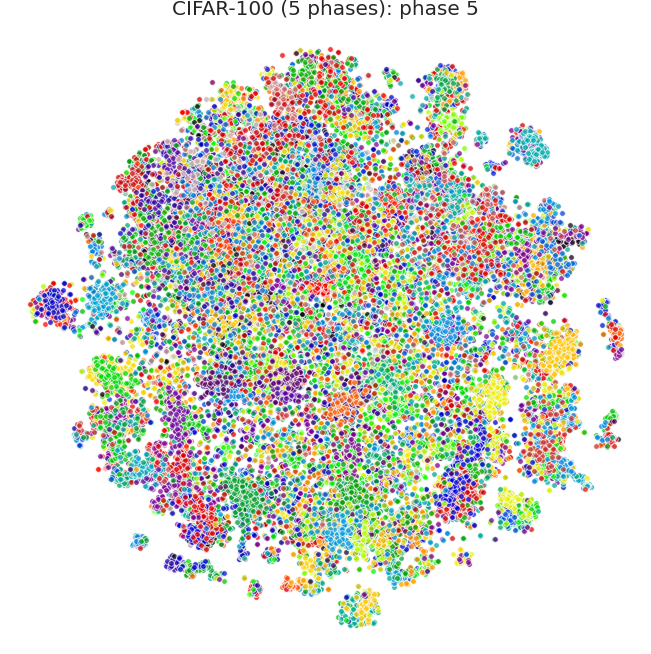}} \\ [-0.0ex]
    \caption{The t-SNE visualization of (a) PASS, (b) iVoro, (c) iVoro-N, and (d) iVoro-NAC on 5-phase CIFAR-100 dataset (phase 5).}\label{fig:tsne-5}
\end{figure}

In this section, we show the t-SNE visualization of feature distribution for PASS~\citep{zhu2021prototype}, iVoro, iVoro-N, and iVoro-NAC on CIFAR-100 dataset, in Fig.~\ref{fig:tsne-0}, Fig.~\ref{fig:tsne-1}, Fig.~\ref{fig:tsne-2}, Fig.~\ref{fig:tsne-3}, Fig.~\ref{fig:tsne-4}, and Fig.~\ref{fig:tsne-5}. Note that for the SSL-based label augmentation, there are 400 classes in total, which 2D t-SNE is unable to clearly illustrate the distributions of all the classes (including the rotated classes).

% ==================== phase 0
% ==================== phase 1
% ==================== phase 2
% ==================== phase 3
% ==================== phase 4
% ==================== phase 5

% ==================== Norm ====================
\clearpage
\section{Detailed Analysis on Parameterized Feature Normalization}\label{supp:norm}
\begin{figure}[ht]
    \centering
    \subfloat{\includegraphics[width=0.47\textwidth]{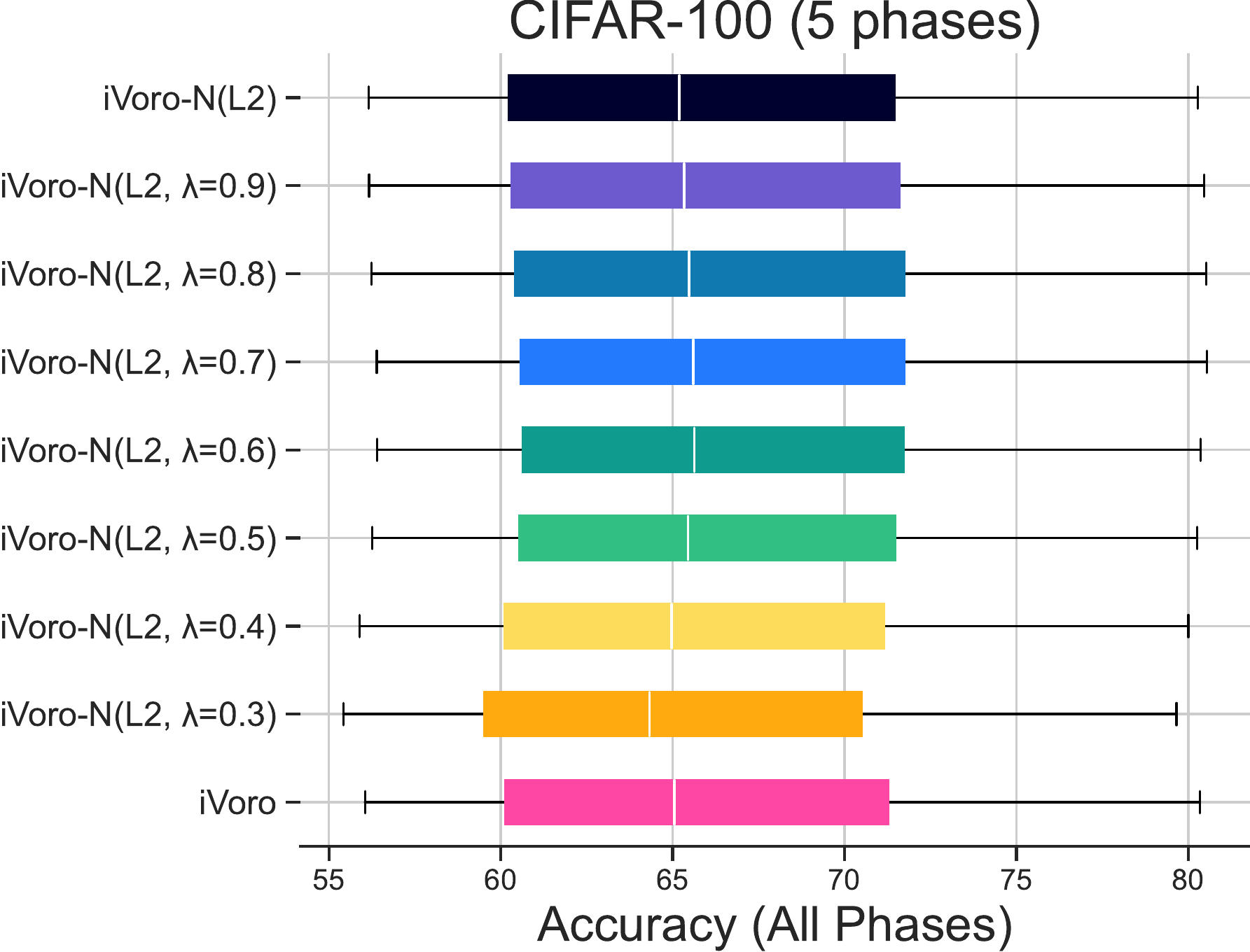}} 
    \subfloat{\includegraphics[width=0.47\textwidth]{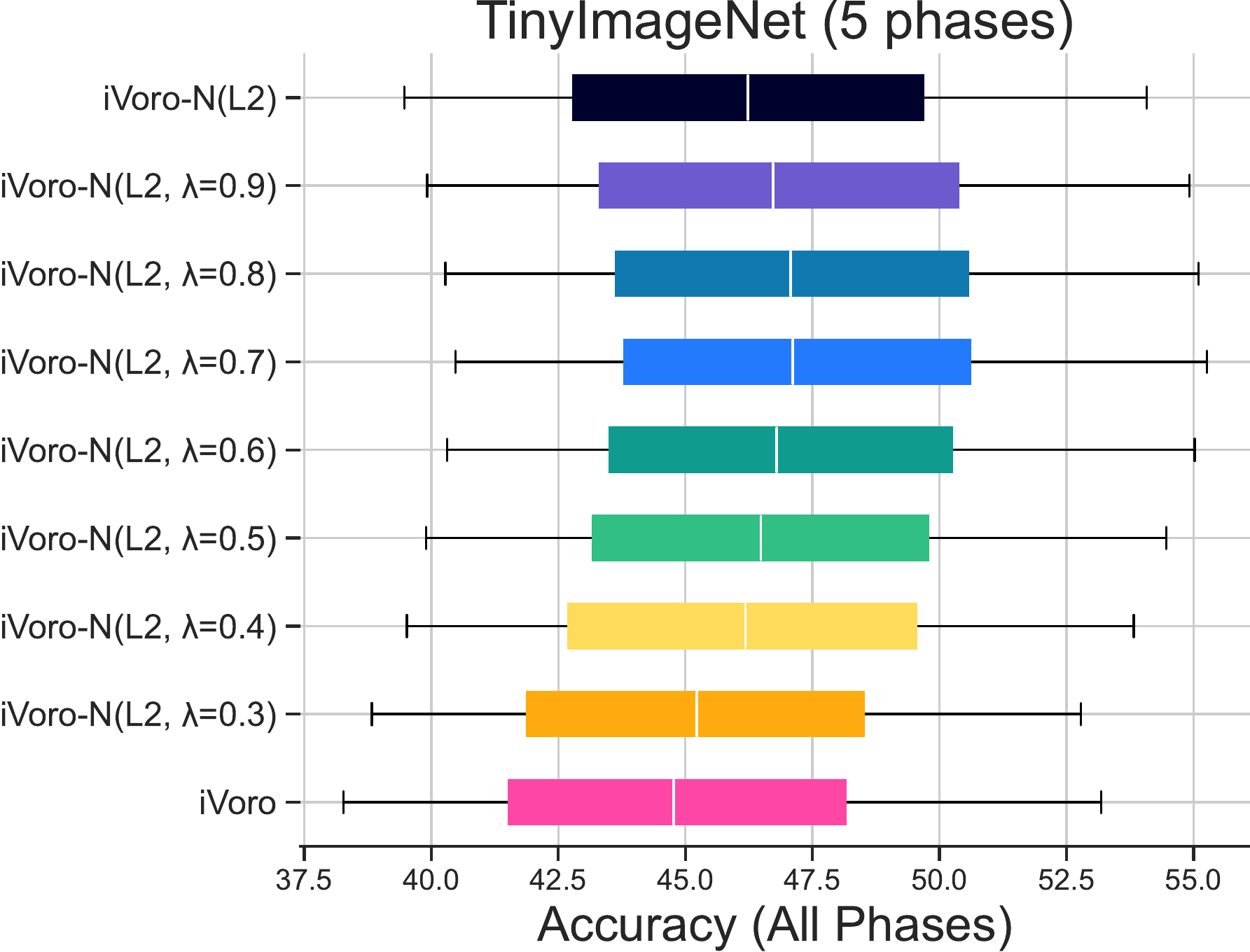}} \\
    \subfloat{\includegraphics[width=0.47\textwidth]{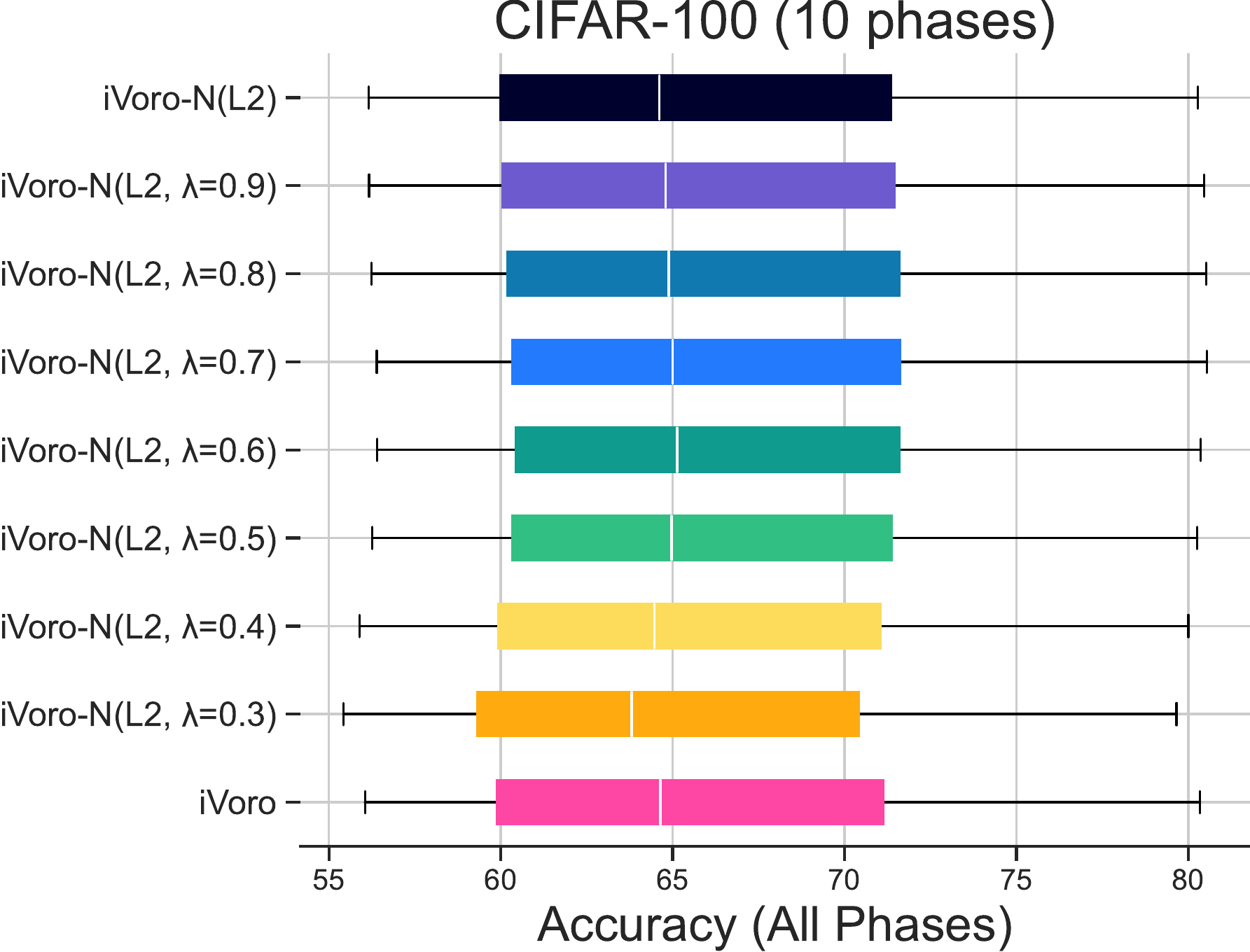}} 
    \subfloat{\includegraphics[width=0.47\textwidth]{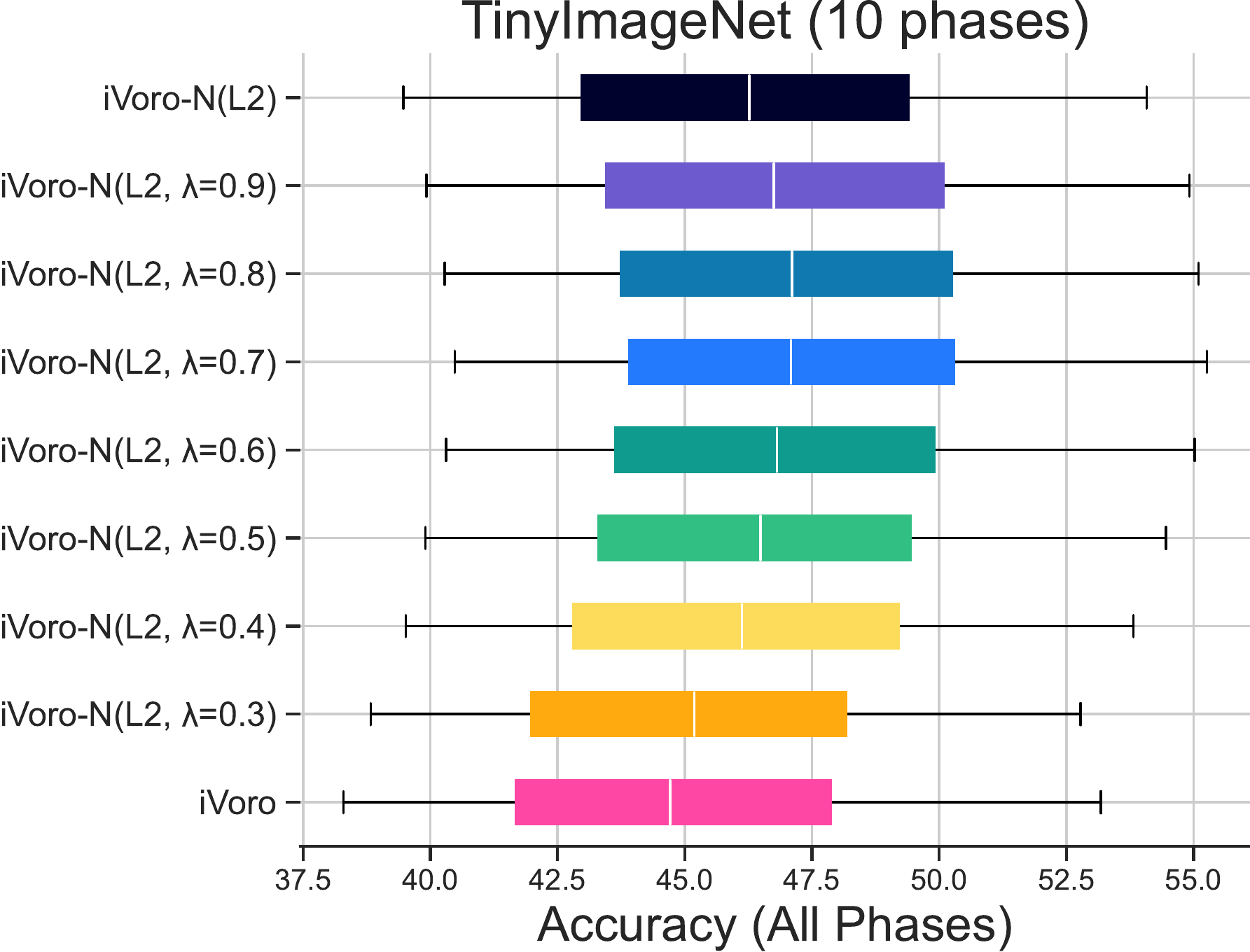}} \\
    \subfloat{\includegraphics[width=0.47\textwidth]{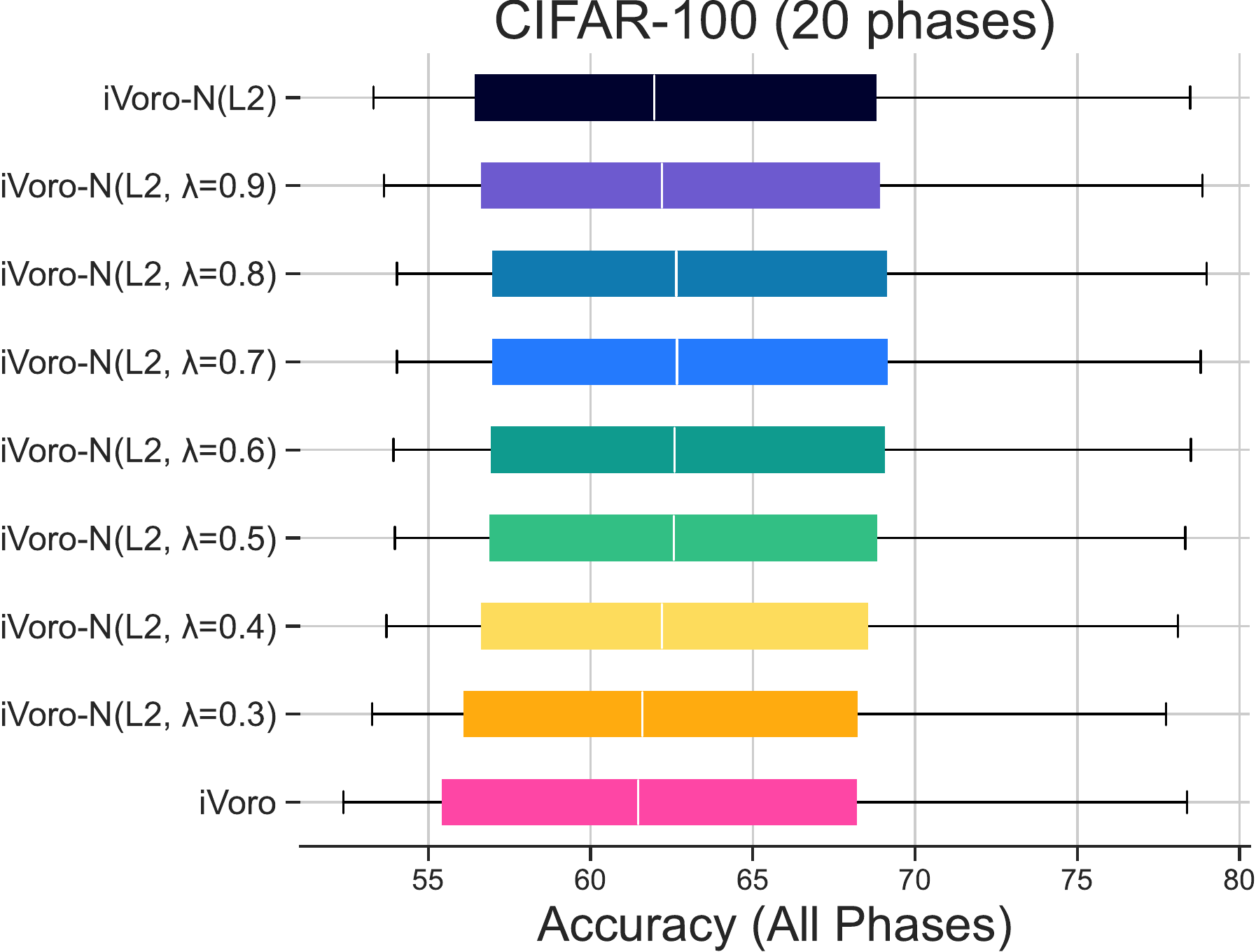}} 
    \subfloat{\includegraphics[width=0.47\textwidth]{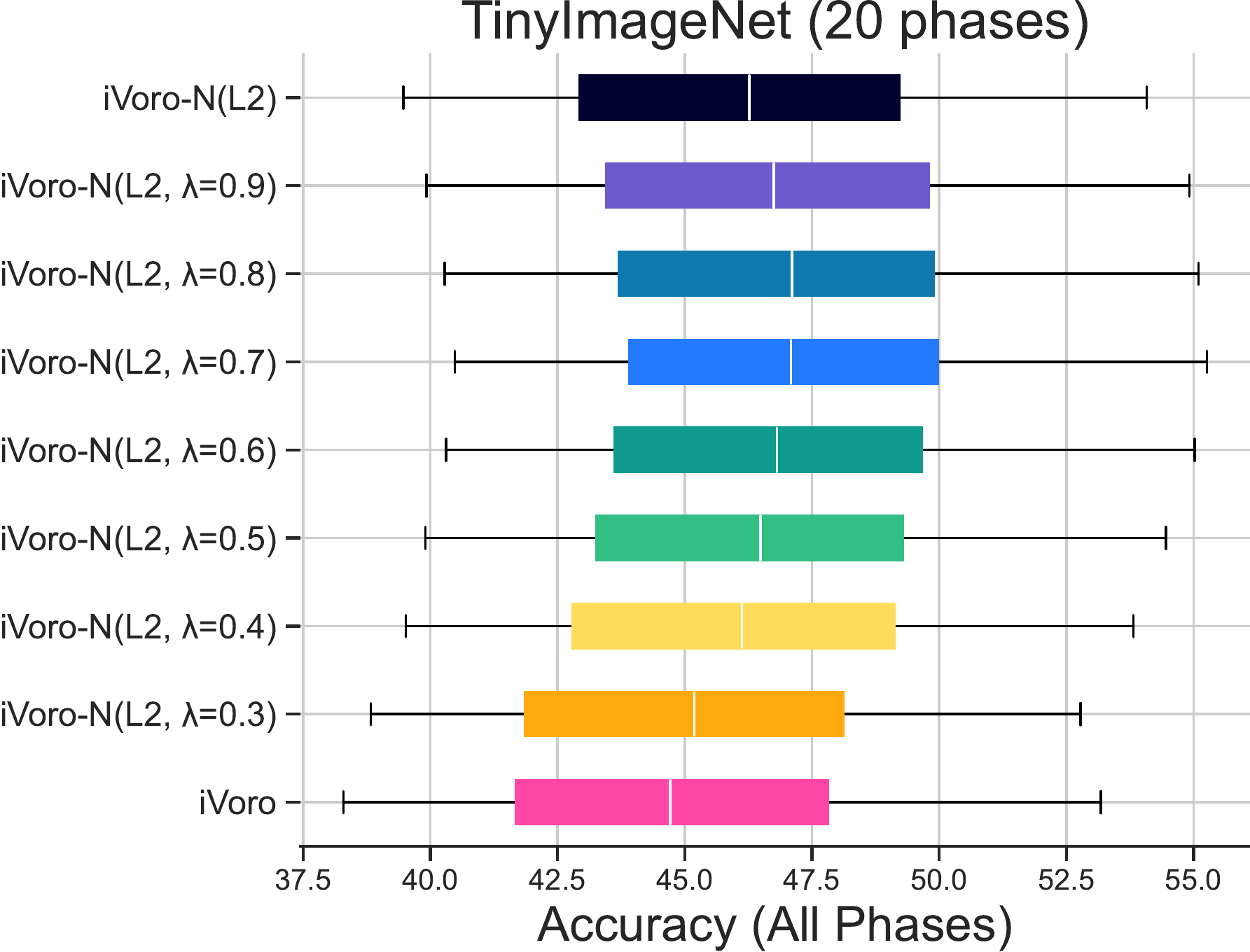}} \\ 
    \caption{Effect of parameterized feature transformation on CIFAR-100 and TinyImageNet.}\label{fig:norm}
\end{figure}

In this section we give a detailed comparison between iVoro (no normalization) and iVoro-N (parameterized feature normalization). On both CIFAR-100 and TinyImageNet with different phases, we show the distribution of accuracy (across all phases) for iVoro, iVoro-N with only $L_2$ normalization, and iVoro-N with both $L_2$ normalization and Tukey’s ladder of powers transformation ($\lambda$ varying from 0.3 to 0.9). As shown in Fig.~\ref{fig:norm}, feature normalization benefits both datasets, but the efficacy is more prominent in more complex dataset e.g. TinyImageNet. Overall, iVoro-N improves the accuracy in the last phase by up to 1.65\% on CIFAR-100, 2.21\% on TinyImageNet, and 2.40 on ImageNet-Subset, respectively, compared to iVoro. See Sec.~\ref{supp:tsne} for the t-SNE visualizations for various methods.
% ==================== Uncertainty ====================
\clearpage
\section{Uncertainty Analysis}\label{supp:unc}
\begin{figure}[ht]
    \centering
    \subfloat{\includegraphics[width=0.95\textwidth]{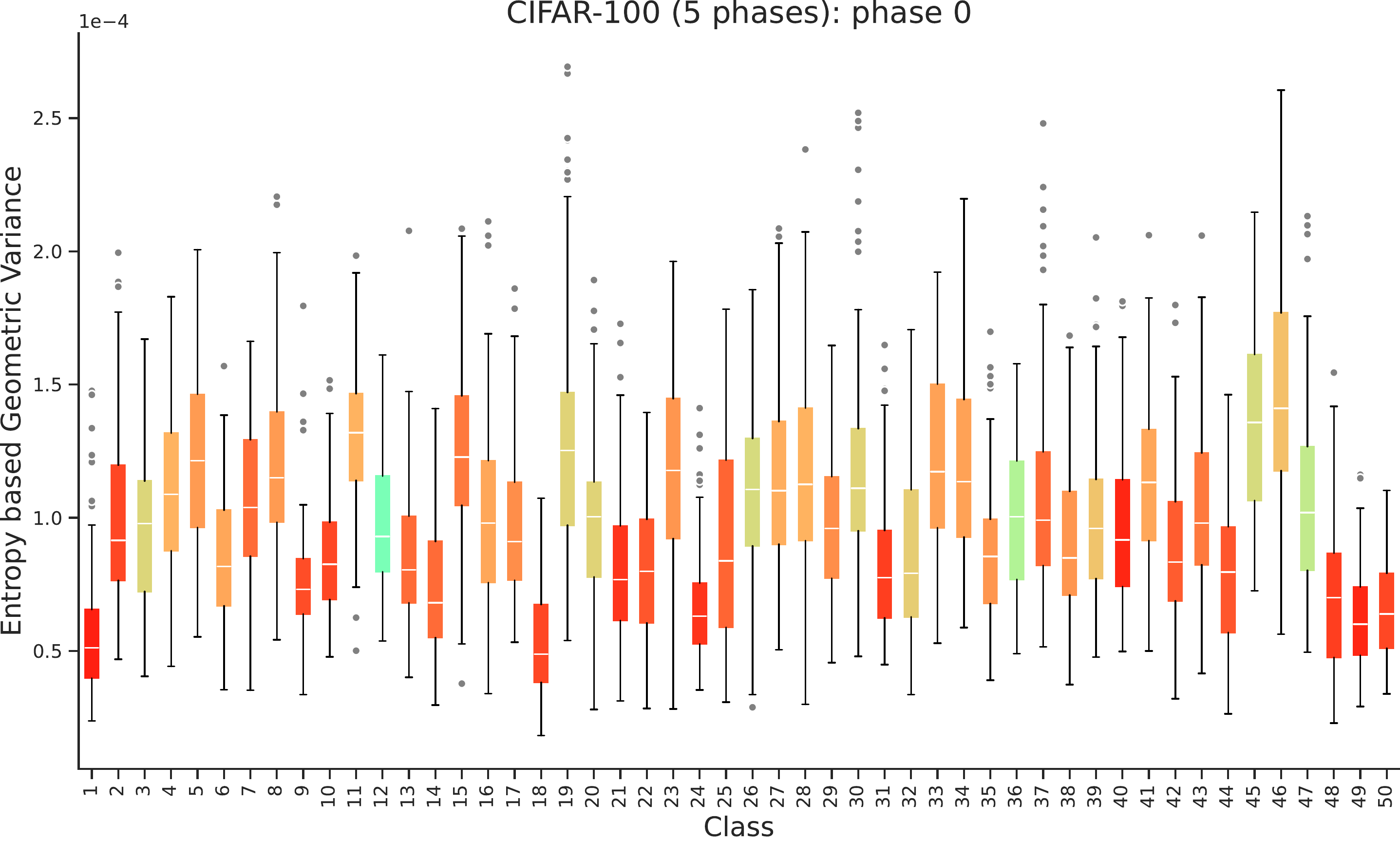}} \\
    \subfloat{\includegraphics[width=0.95\textwidth]{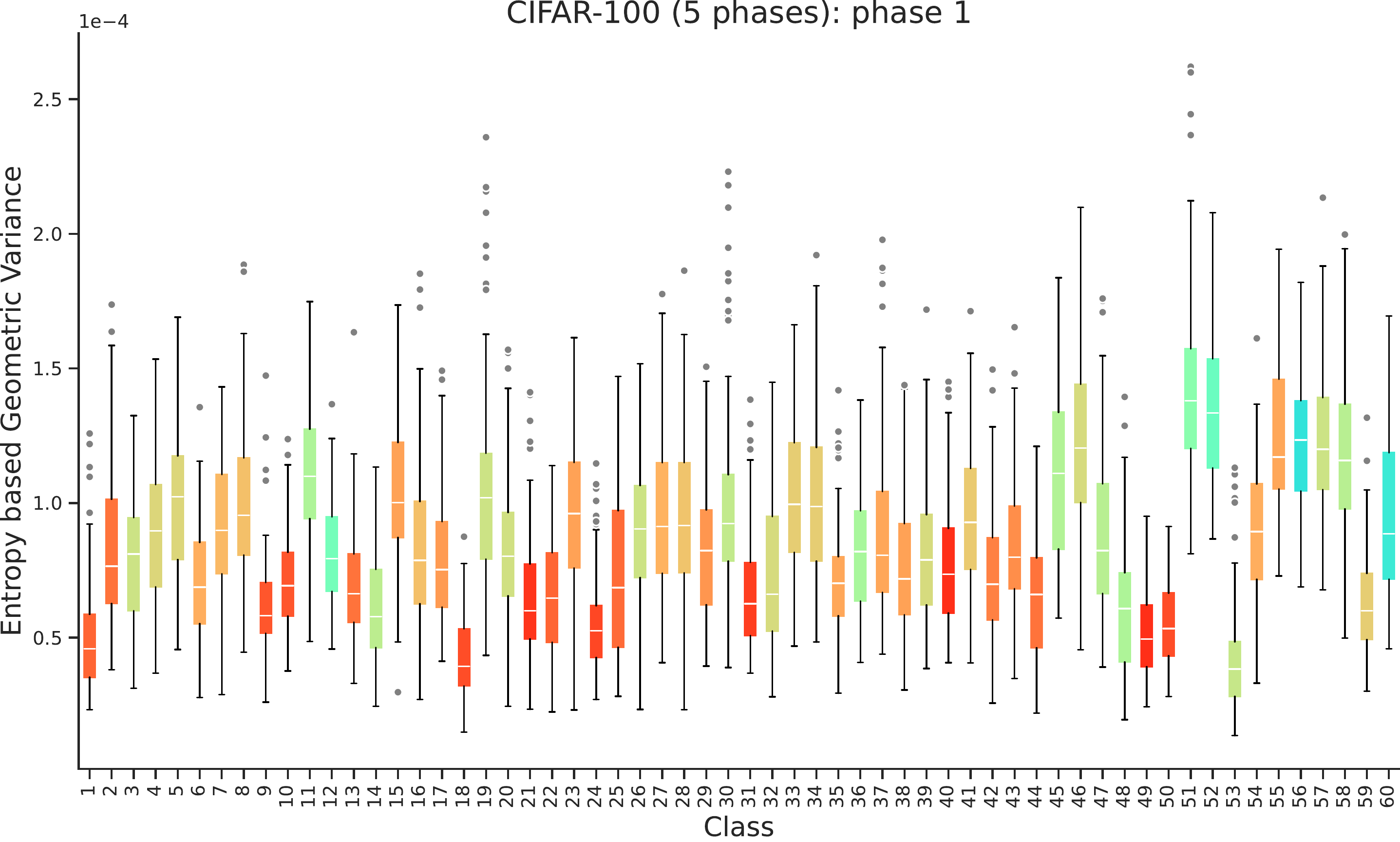}} \\
    \caption{The distributions of Entropy-based geometric variance for each class in each phase on 5-phase CIFAR-100 (phase 0 to phase 1).}\label{fig:unc-1}
\end{figure}

\begin{figure}[ht]
    \centering
    \subfloat{\includegraphics[width=0.95\textwidth]{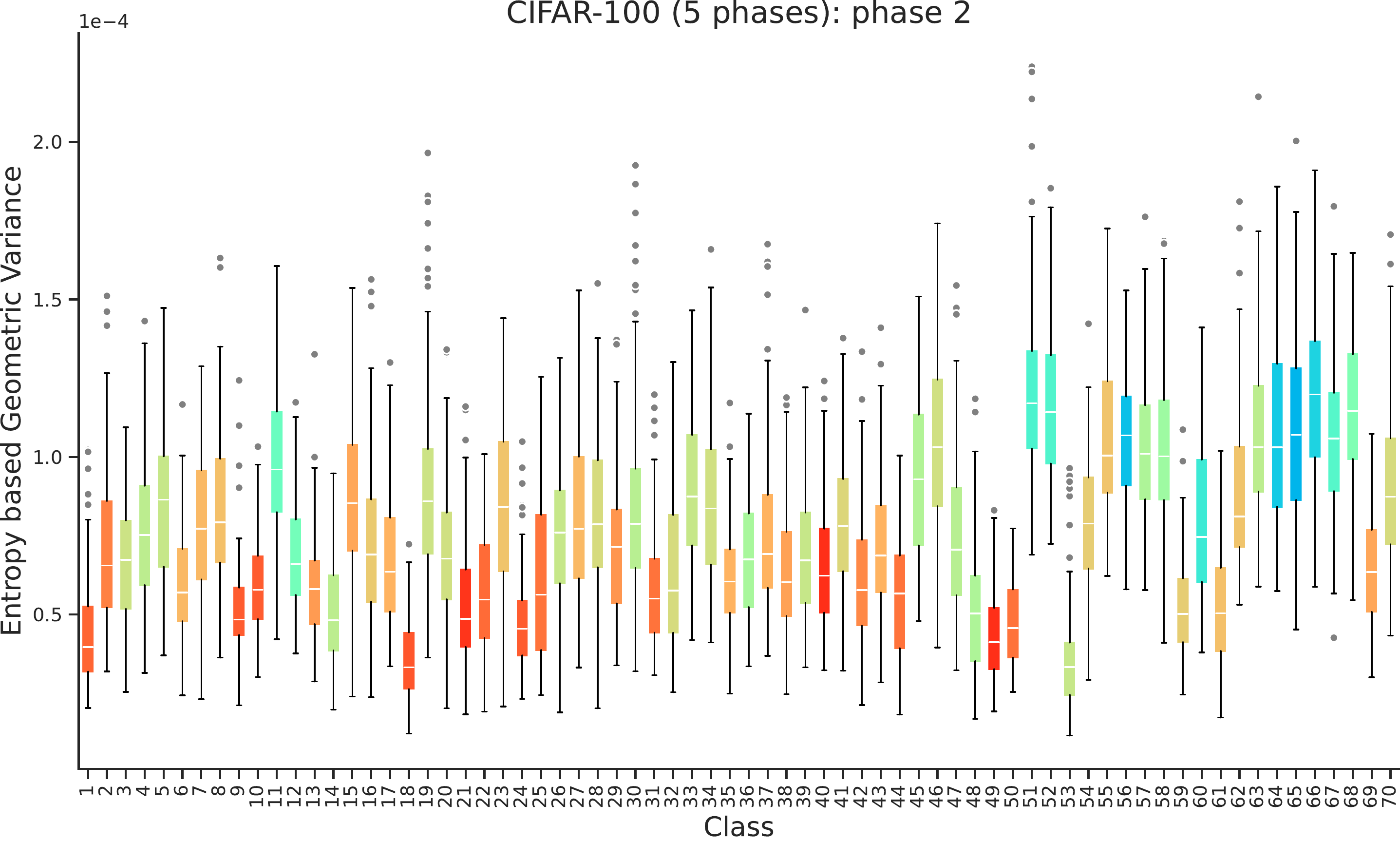}} \\
    \subfloat{\includegraphics[width=0.95\textwidth]{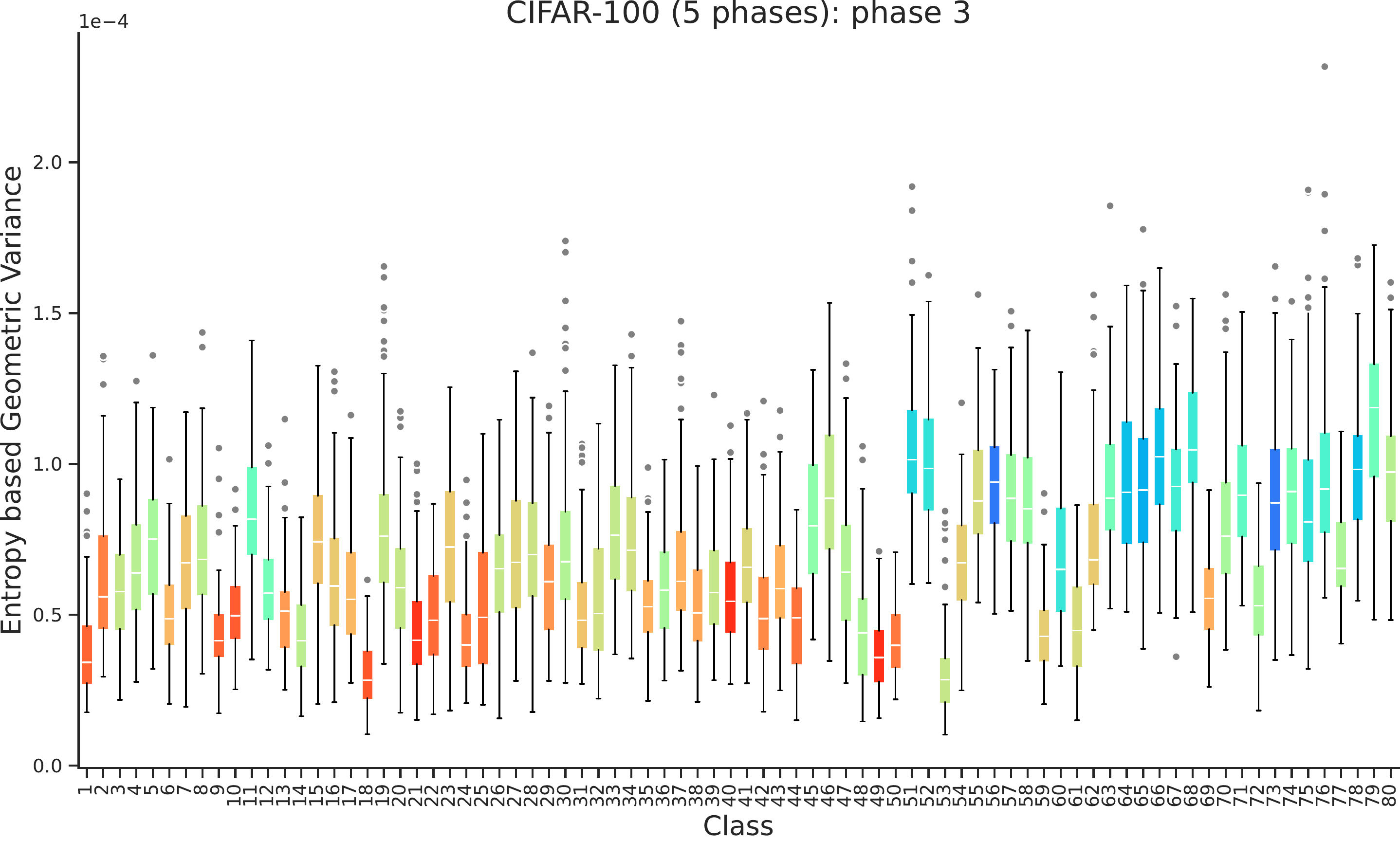}} \\
    \caption{The distributions of Entropy-based geometric variance for each class in each phase on 5-phase CIFAR-100 (phase 2 to phase 3).}\label{fig:unc-2}
\end{figure}

\begin{figure}[ht]
    \centering
    \subfloat{\includegraphics[width=0.95\textwidth]{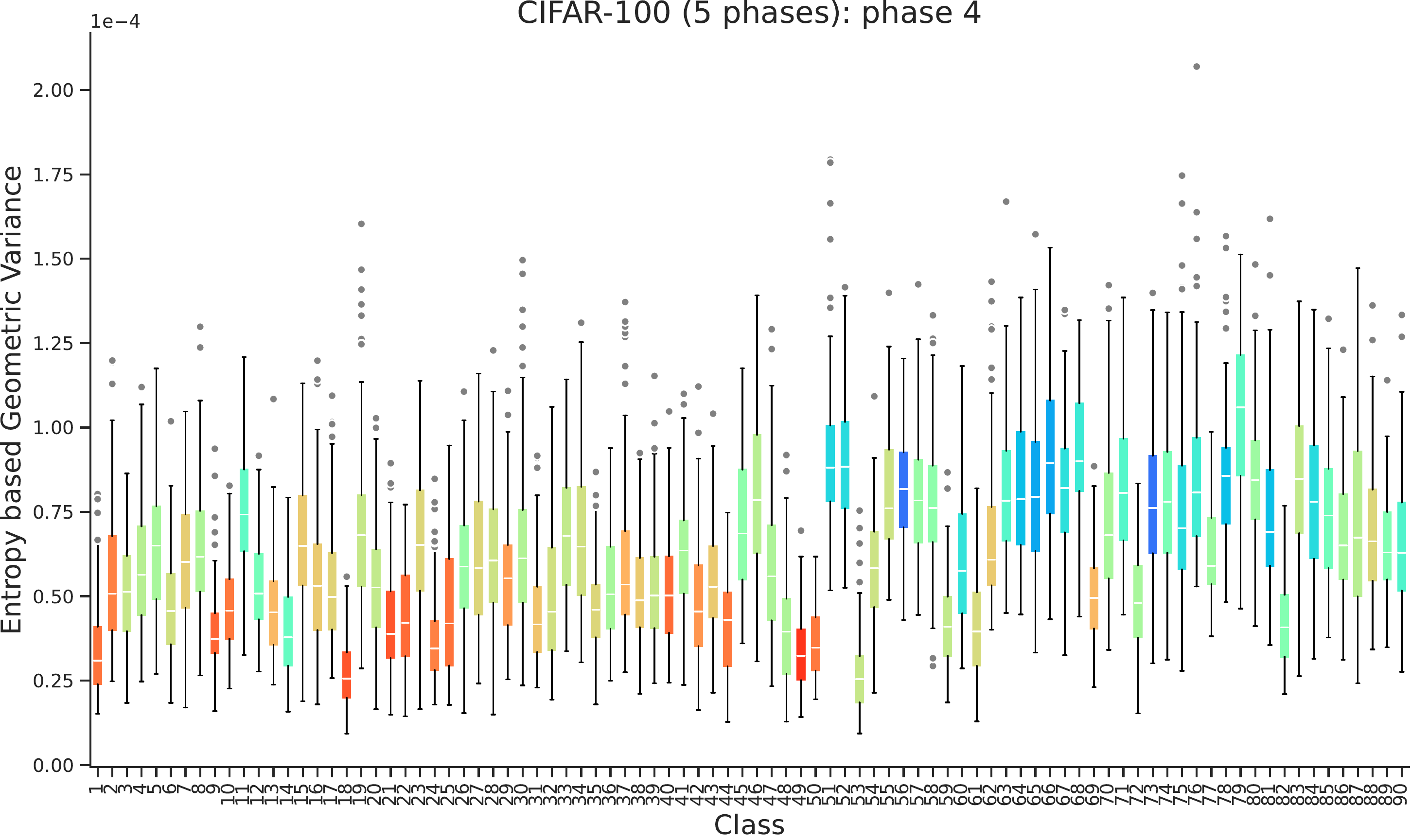}} \\
    \subfloat{\includegraphics[width=0.95\textwidth]{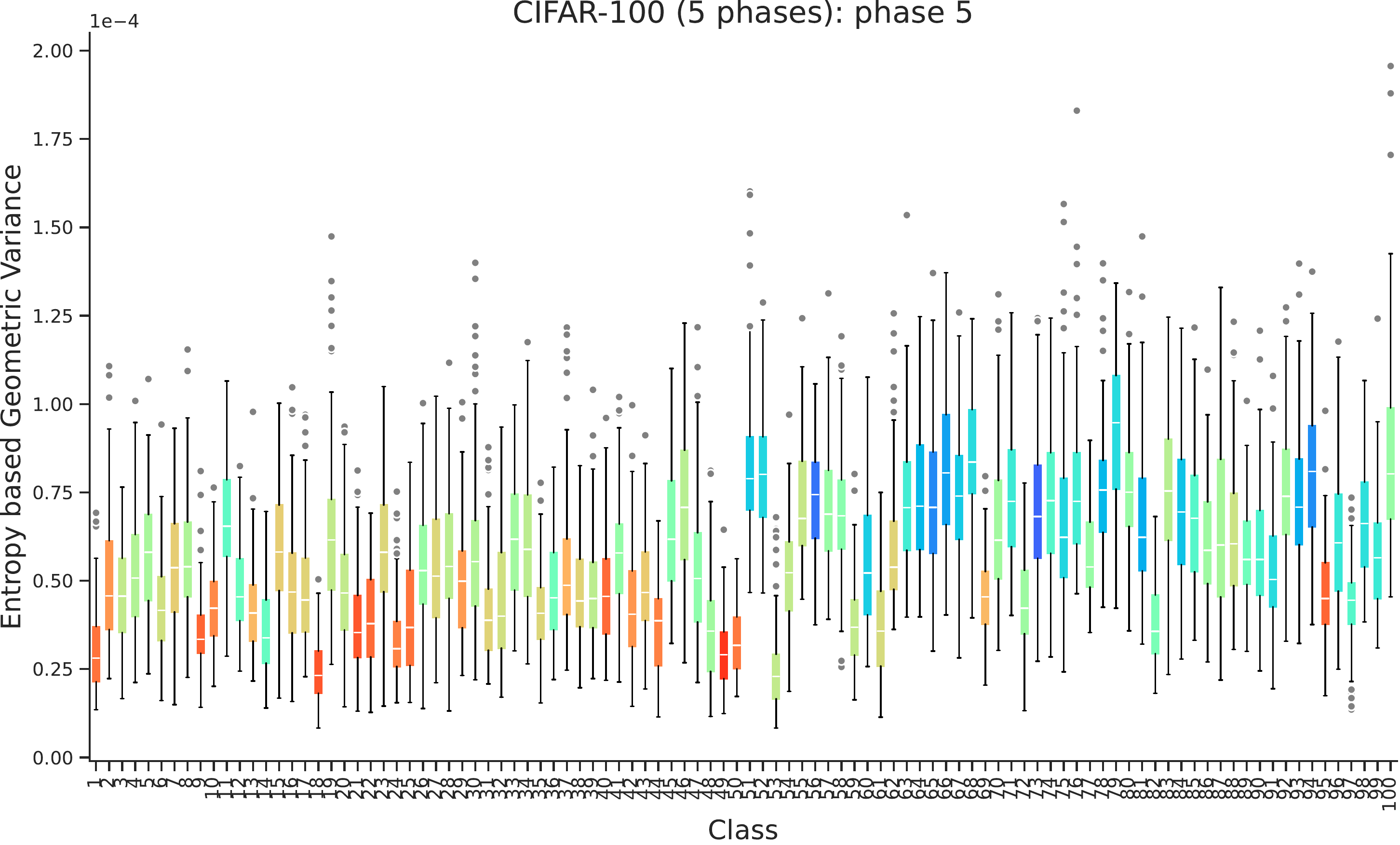}} \\
    \caption{The distributions of Entropy-based geometric variance for each class in each phase on 5-phase CIFAR-100 (phase 4 to phase 5).}\label{fig:unc-3}
\end{figure}

In this section, we demonstrate that the Entropy-based geometric variance (HV) is a good indicator of predictive uncertainty induced by SSL-based label augmentation, and also exhibits high correlation with predictive accuracy.
Fig.~\ref{fig:unc-1}, Fig.~\ref{fig:unc-2} and Fig.~\ref{fig:unc-3} present the distribution of HV of each class during each phase (0 to 5) in 5-phase CIFAR-100 data. The color of the box reflects the accuracy of iVoro for the specific class. 
It can be observed in these figures that, the class with higher predictive accuracy will typically exhibit lower geometric uncertainty.

\vspace{1em}

% ==================== Feature extractor ====================
\clearpage
\section{Analysis on the Feature Extractor}\label{supp:ext}
In order to examine the effect of the feature extractor on the final result, we gradually decrease the number of classes in the first phase from 50 to 40, 30, 20, and 10 and still include 5 classes in each subsequent phase.
When compared with PASS, the best version of iVoro still has 17.75\%, 13.59\%, 10.89\%, and 1.60\% improvements with 40, 30, 20, and 10 initial classes, respectively.

\vspace{1em}
\begin{figure}[ht]
    \centering
    \subfloat{\includegraphics[width=0.78\textwidth]{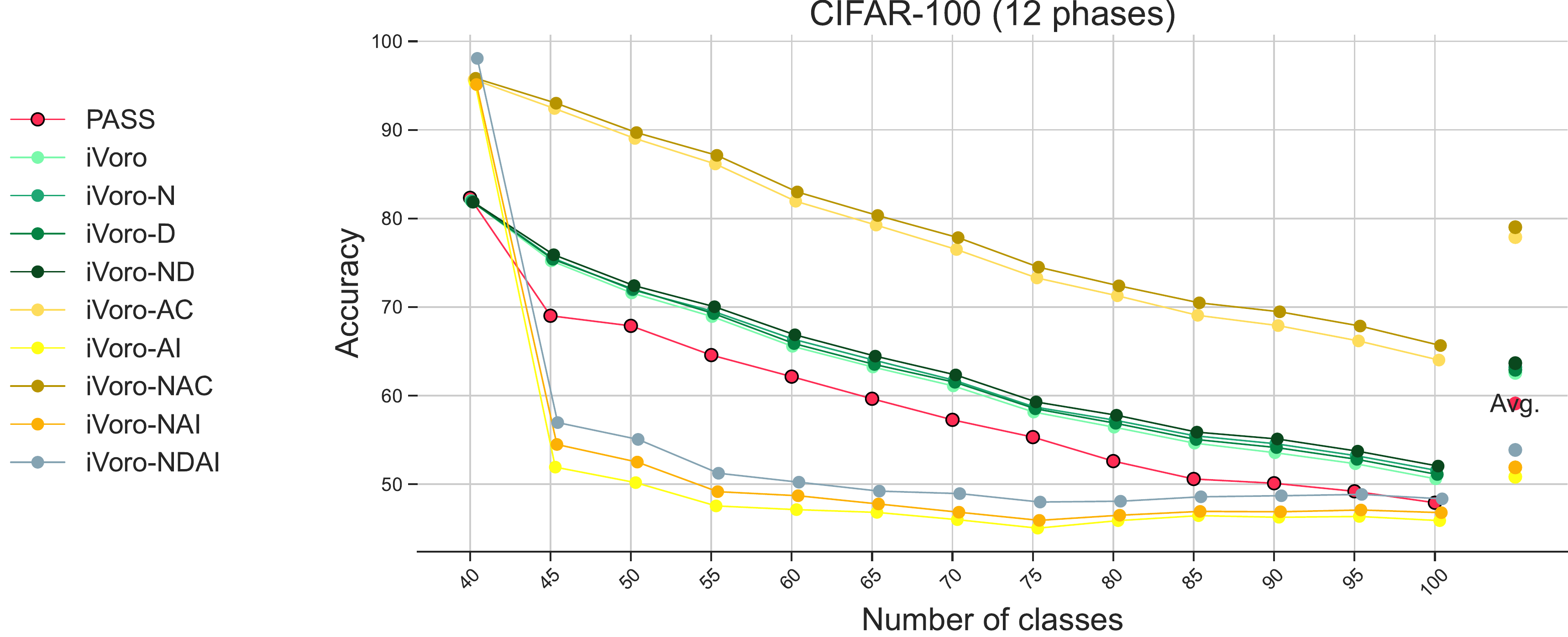}} \\
    \subfloat{\includegraphics[width=0.78\textwidth]{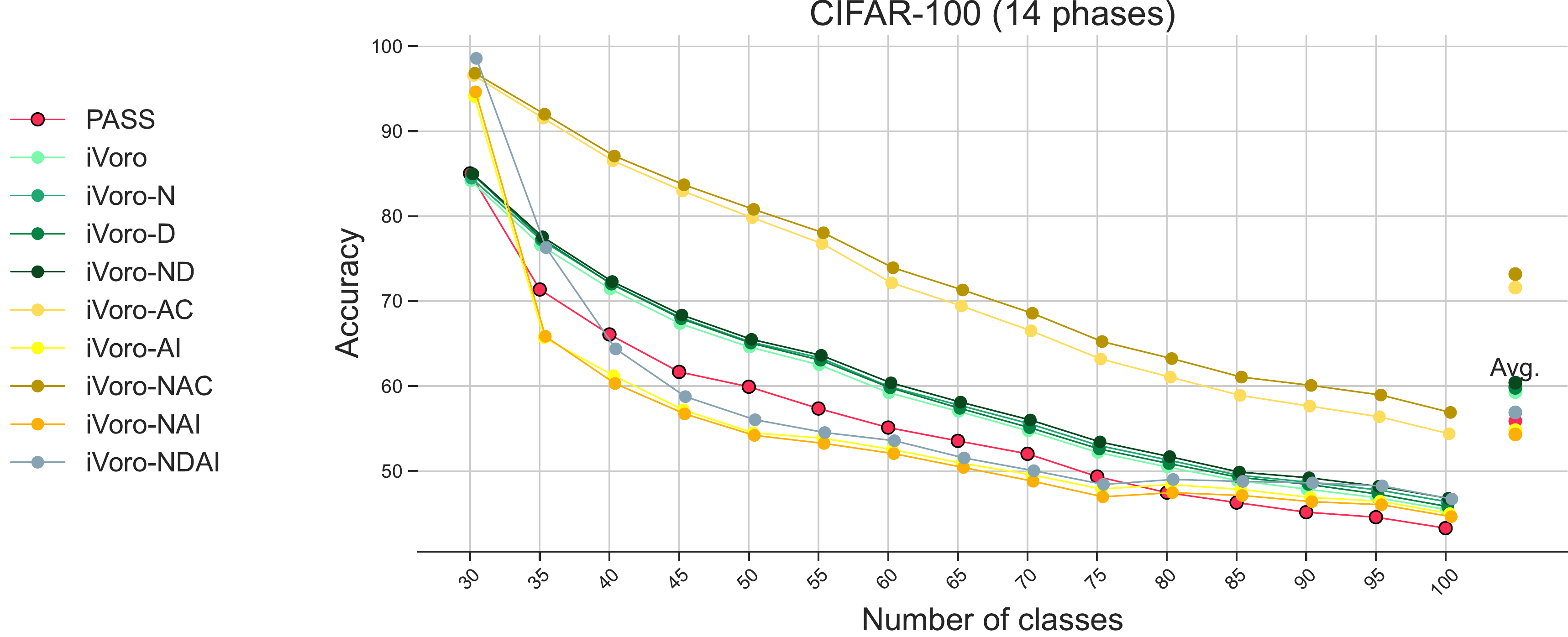}} \\
    \subfloat{\includegraphics[width=0.78\textwidth]{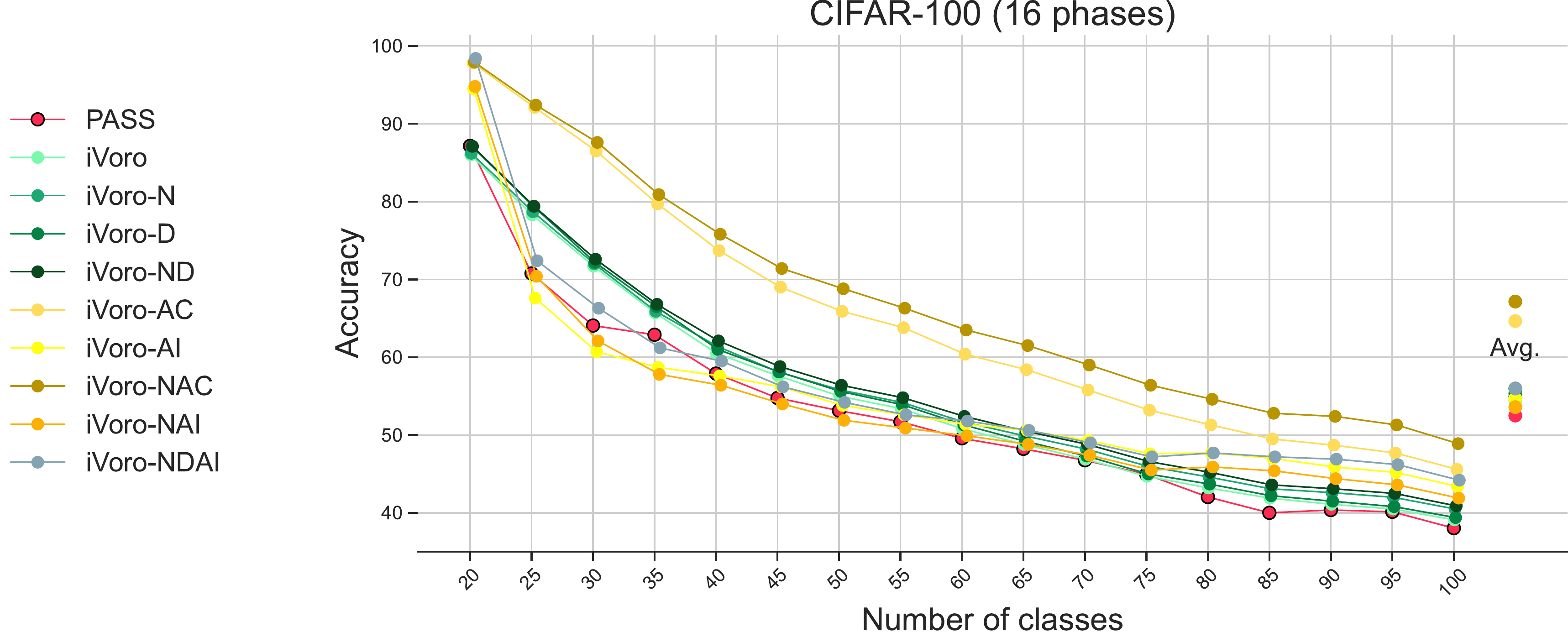}} \\ 
    \subfloat{\includegraphics[width=0.78\textwidth]{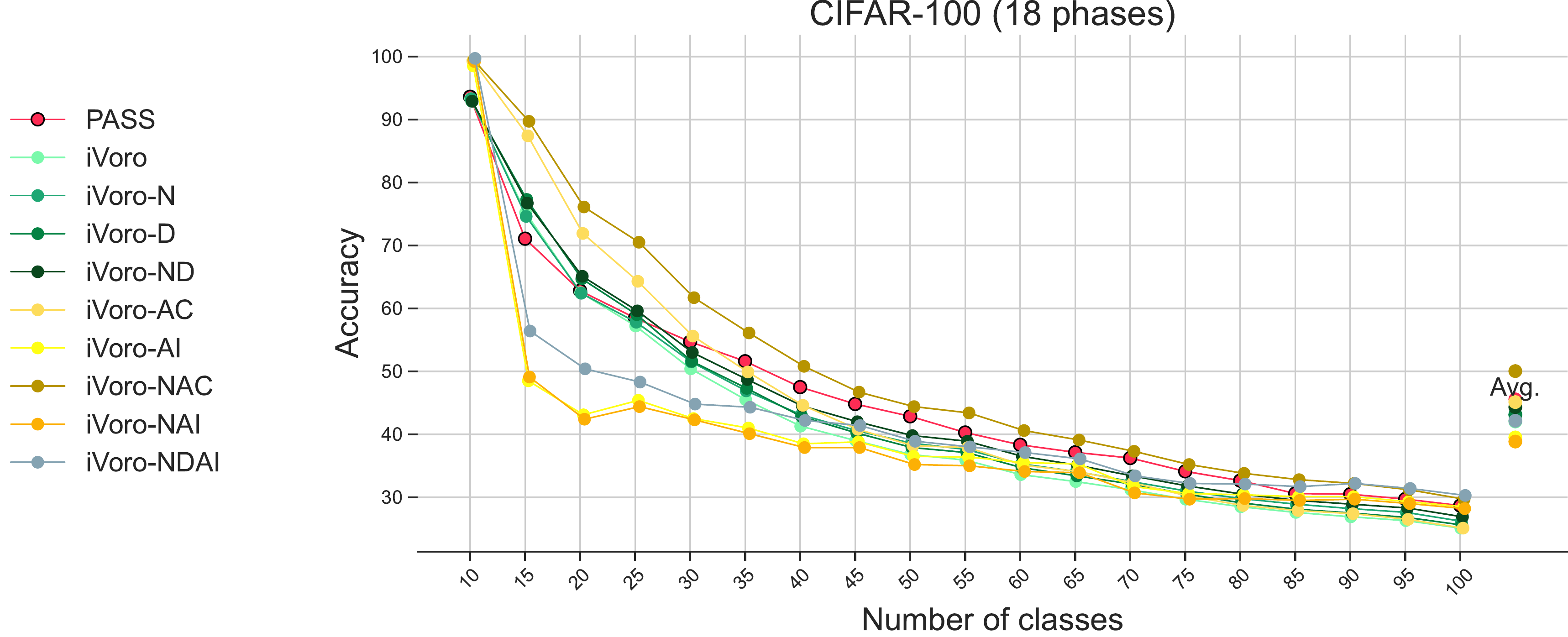}}
    \caption{Top-1 classification accuracy on CIFAR-100 during 12/14/16/18 phases of CIL, in which the number of classes in the first phase are decreased to 40/30/20/10, respectively.}\label{fig:base-phase}
\end{figure}

% ==================== CIVD ====================
\section{Analysis on the Layered Voronoi Diagrams}\label{supp:civd}
In this section, we extract the feature from the $3^{rd}$ block and rebuild iVoro, to be iVoro3, etc. The results for iVoro3, iVoro3-N, iVoro3-NAC, and iVoro3-NAI are shown in Fig.~\ref{fig:abl-cifar}, Fig.~\ref{fig:abl-tiny}, and Fig.~\ref{fig:abl-subset}. As expected, there is always a substantial decrease in accuracy if only the features from the $3^{rd}$ block are used. For example, iVoro drops from 38.27\% to 18.71\%, from 56.05\% to 43.22\%, and from 55.40\% to 35.10\% on TinyImageNet, CIFAR-100, and ImageNet-Subset, respectively. However, when augmentation integration is used, the accuracy of iVoro3-NAI becomes 42.03\% (TinyImageNet), 66.13\% (CIFAR-100), and 64.32\% (ImageNet-Subset), even much higher than iVoro and PASS. Moreover, when integrating the features from $\phi$ and $\phi^{(3)}$ using CCVD, iVoro-NDAIL achieves the best performance across all variants of iVoro, 72.34\% on TinyImageNet and 83.84\% on ImageNet-Subset.
% ==================== All algorithms ====================
\section{Complete List of Algorithms}\label{supp:alg}
Here we provide the algorithm for Voronoi Diagram reduction~\citep{ma2022fewshot} in Alg.~\ref{alg:power}, the algorithm for iVoro in Alg.~\ref{alg:ivoro}, the algorithm for iVoro-D in Alg.~\ref{alg:ivoro_d}.
\vspace{2em}
\IncMargin{1em}
\begin{algorithm}
  \SetAlgoNoLine
  \KwData{Local data $\gD_t$ in phase $\tau$}
  \KwResult{$\mW_{\tau}$} 
  Initialize $\mW_{\tau} \gets \mW_{\tau}^{(0)}$\;
  \For(){$\textrm{epoch} \gets 1,...,\textrm{\#epoch}$}{
    $\evb_{\tau,k} \gets -\frac{1}{4} ||\mW_{\tau,k}||_2^2, \forall k = 1,...,K_{\tau}$ \Comment*[r]{Apply Theorem \ref{thm:thm}}
    Compute loss $\gL(\mW_{\tau}, \vb_{\tau})$ \Comment*[r]{forward propagation}
    Update $\mW_{\tau}$ \Comment*[r]{backward propagation}    
  }      
  \Return{$\mW_{\tau}$}
  \caption{Voronoi Diagram-based Logistic Regression.}\label{alg:power}
  \end{algorithm}
\DecMargin{1em}

\vspace{2em}
\IncMargin{1em}
\begin{algorithm}
  \SetAlgoNoLine
  \KwData{Training datasets until phase $t$: $\gD_{\tau} = \{(\vx_{\tau,i}, y_{\tau,i})\}_{i=1}^{N_{\tau}}, \vx_{\tau,i} \in \sD, y_{\tau,i} \in \gC_{\tau}, \tau \in \{1,...,t\}$, query example $\vx$}
  \KwResult{prediction $\hat{y}$} 
     \For(){$\tau \in \{1,...,t\}$}{
        \For(){$k \in \{1,...,K_{\tau}\}$}{
            $\vc_{\tau, k} \gets \frac{1}{N_{\tau, k}}\ {\textstyle\sum}_{i \in \{1,...,N_{\tau, k}\}, y = k}\ \phi(\vx_{\tau, i})$ \Comment*[r]{prototypical centers}
            $\nu_{\tau, k} \gets 0$
        }
     }
     $\vz \gets \phi(\vx)$ \\
     $\vz \gets (h_\lambda \circ g_{w,\eta} \circ f)(\vz)$ \Comment*[r]{iVoro-N, optional}
     $\hat{y} \gets \gC_{\tau', k'}$ s.t. $d(\vz,\vc_{\tau', k'}) = \min_{\tau, k} d(\vz, \vc_{\tau, k})$ \\
  \Return{$\{\vc_{\tau, k}\}, \hat{y}$}    
  \caption{iVoro Algorithm.}\label{alg:ivoro}
  \end{algorithm}
\DecMargin{1em}

\vspace{1em}
\IncMargin{1em}
\begin{algorithm}
  \SetAlgoNoLine
  \KwData{Training datasets until phase $t$: $\gD_{\tau} = \{(\vx_{\tau,i}, y_{\tau,i})\}_{i=1}^{N_{\tau}}, \vx_{\tau,i} \in \sD, y_{\tau,i} \in \gC_{\tau}, \tau \in \{1,...,t\}$, query example $\vx$}
  \KwResult{Prediction $\hat{y}$} 
     \For(){$\tau \in \{1,...,t\}$}{
        $\mW_{\tau}, \vb_{\tau} \gets$ Algorithm~\ref{alg:power}$(\gD_{\tau})$ \\
        $\tilde{\vc}_{\tau,k} \gets \frac{1}{2} \mW_{\tau,k}$ \Comment*[r]{probing-induced centers}
        $\vc_{\tau, k} \gets$ Algorithm~\ref{alg:ivoro}$(\gD_{\tau})$ \Comment*[r]{prototypical centers}
        \For(){$\gC_{\tau,k_1}, \gC_{\tau,k_2} \in \gC_{\tau}$}{
            $\vv \gets \frac{\tilde{\vc}_{\tau,k_1} - \tilde{\vc}_{\tau,k_2}}{||\tilde{\vc}_{\tau,k_1} - \tilde{\vc}_{\tau,k_2}||_2}$ \\
            $q \gets \frac{||\tilde{\vc}_{\tau,k_1}||_2^2 - ||\tilde{\vc}_{\tau,k_2}||_2^2}{2||\tilde{\vc}_{\tau,k_1} - \tilde{\vc}_{\tau,k_2}||_2}$ \Comment*[r]{within-clique boundaries}
        }
     }
     \For(){$\gC_{\tau_1,k} \in \gC_{\tau_1}, \gC_{\tau_2,k'} \in \gC_{\tau_2}$}{
        $\vv \gets \frac{\vc_{\tau_1,k} - \vc_{\tau_2,k'}}{||\vc_{\tau_1,k} - \vc_{\tau_2,k'}||_2}$ \\
        $q \gets \frac{||\vc_{\tau_1,k}||_2^2 - ||\vc_{\tau_2,k'}||_2^2}{2||\vc_{\tau_1,k} - \vc_{\tau_2,k'}||_2}$ \Comment*[r]{cross-clique boundaries}
     }
     \For(){$\Gamma \in \{\Gamma\}$}{
        Delete candidate and its boundaries w.r.t. $\mathrm{sign}(\vv^T \vz - q)$ \\
        \If(){only one candidate $\gC_{\tau',k'}$ remains}{
            $\hat{y} \gets \gC_{\tau',k'}$
        }
     }
  \Return{$\hat{y}$}    
  \caption{iVoro-D Algorithm. The time complexity for the establishment of VD is $\mathcal{O}(({\textstyle\sum}_{\tau=1}^{t}K_\tau)^2)$, the for querying the VD is $\mathcal{O}({\textstyle\sum}_{\tau=1}^{t}K_\tau)$.}\label{alg:ivoro_d}
  \end{algorithm}
\DecMargin{1em}

\vspace{1em}
\IncMargin{1em}
\begin{algorithm}
  \SetAlgoNoLine
  \KwData{Local data $\gD_t$ in phase $\tau$}
  \KwResult{$\Delta \mW_{\tau}$} 
  Initialize $\mW_{\tau,k}^{(0)} \gets 2\vc_{\tau, k}, \evb_{\tau,k}^{(0)} \gets -{1}/{4} ||\mW_{\tau,k}^{(0)}||_2^2$\;
  Initialize $\Delta \mW_t \gets \mathbf{0}, \Delta \vb_t \gets 0$\;
  \For(){$\textrm{epoch} \gets 1,...,\textrm{\#epoch}$}{
    $\mW_{\tau} \gets \mW_{\tau}^{(0)} + \Delta \mW_{\tau}$ \\
    $\evb_{\tau,k} \gets -\frac{1}{4} ||\mW_{\tau,k}||_2^2, \forall k = 1,...,K_{\tau}$ \Comment*[r]{Apply Theorem \ref{thm:thm}}    Compute loss $\tilde{\gL}(\Delta \mW_{\tau}, \Delta \vb_{\tau}) = \gL(\mW_{\tau}, \vb_{\tau}) + \beta ||\Delta \mW_{\tau}||_2^2$ \Comment*[r]{forward propagation}
    Update $\Delta \mW_{\tau}$ \Comment*[r]{backward propagation}
  }
  \Return{Residue $\Delta \mW_{\tau}$}
  \caption{Voronoi Residual Prototypical Networks.}\label{alg:ivoro_r}
  \end{algorithm}
\DecMargin{1em}

% ==================== More figures ====================
\clearpage
\section{Additional Figures}\label{supp:fig}
This section includes two additional figures, classification accuracy on CIFAR-100 (5/10/20 phases) and on ImageNet-Subset (10 phases), accompanying Fig.~\ref{fig:tiny} (on TinyImageNet).

\begin{figure}[ht]
    \centering
    \subfloat{\includegraphics[height=1.3in]{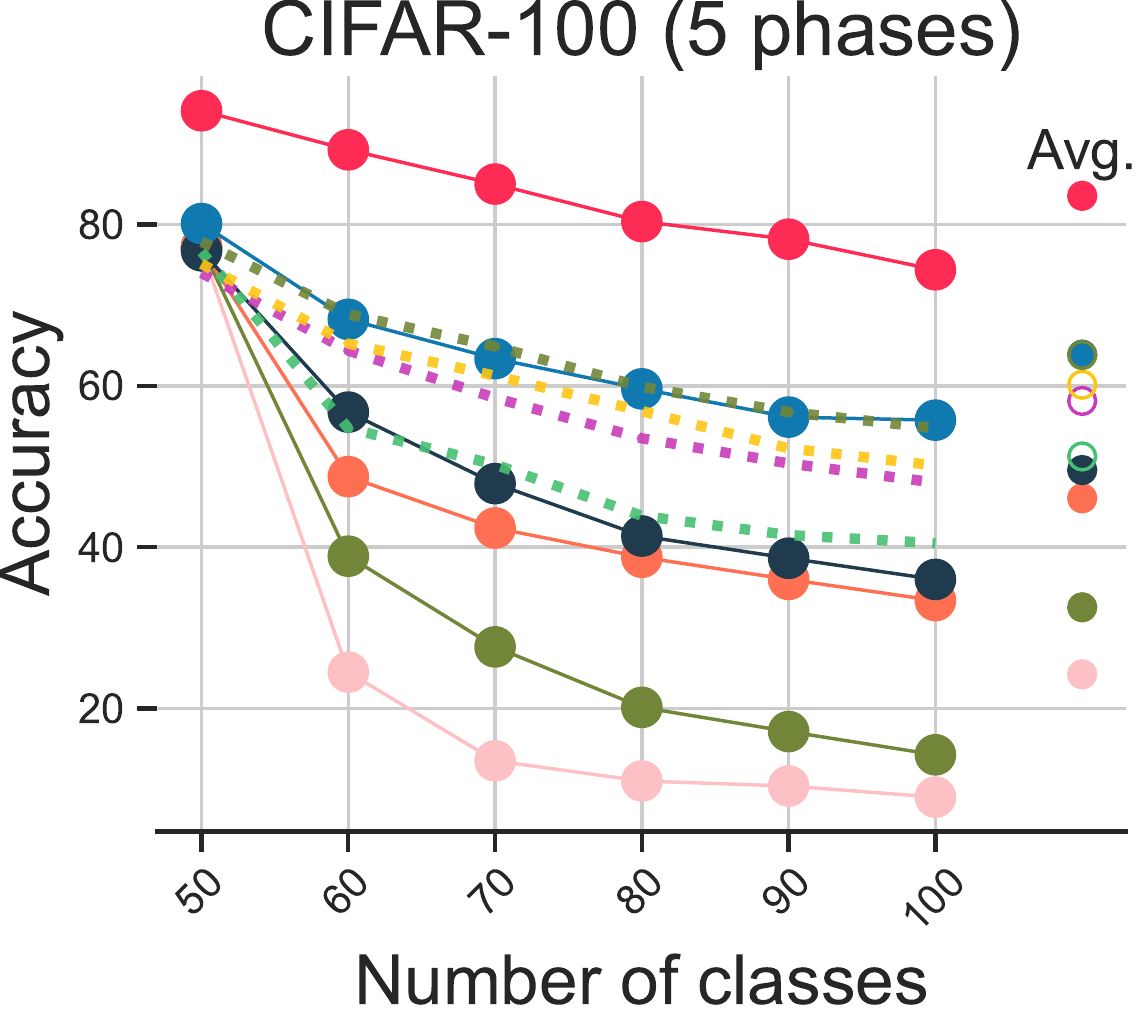}} \hspace{1mm}
    \subfloat{\includegraphics[height=1.3in]{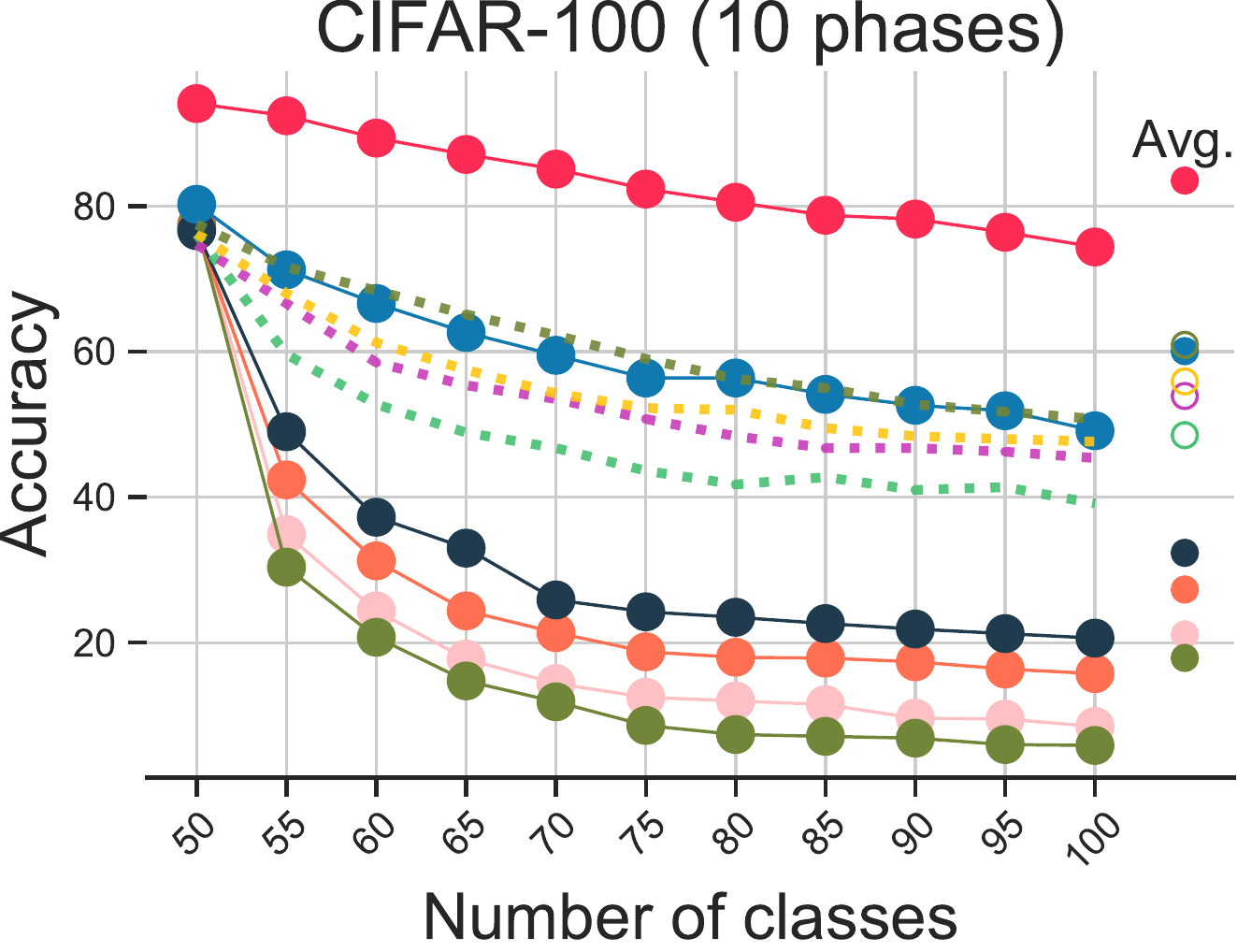}} \hspace{1mm}
    \subfloat{\includegraphics[height=1.3in]{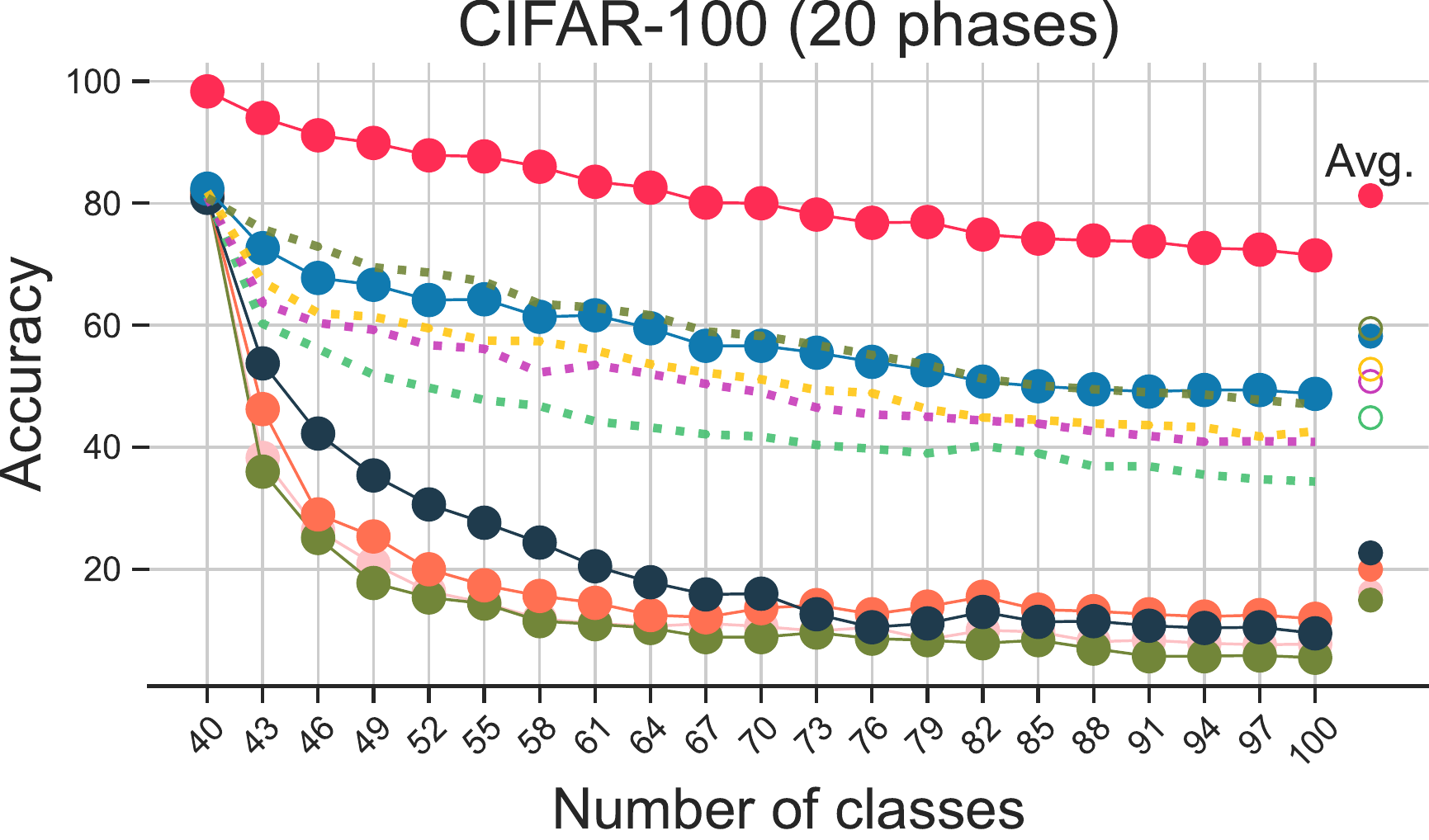}} \\ [-0.0ex]
    \subfloat{\includegraphics[width=0.7\linewidth]{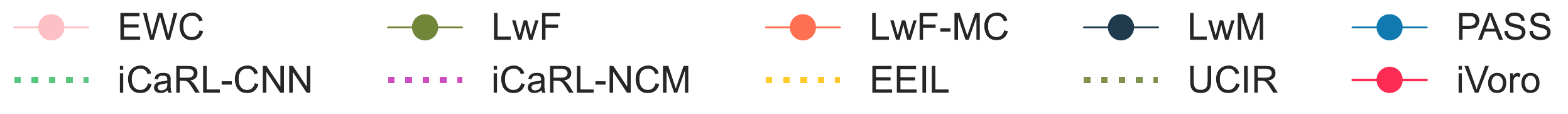}}
    \caption{Top-1 classification accuracy on CIFAR-100 during 5/10/20 phases of CIL.}\label{fig:cifar}
\end{figure}

\vspace{10mm}
\begin{figure}[ht]
    \centering
    \includegraphics[width=0.75\linewidth]{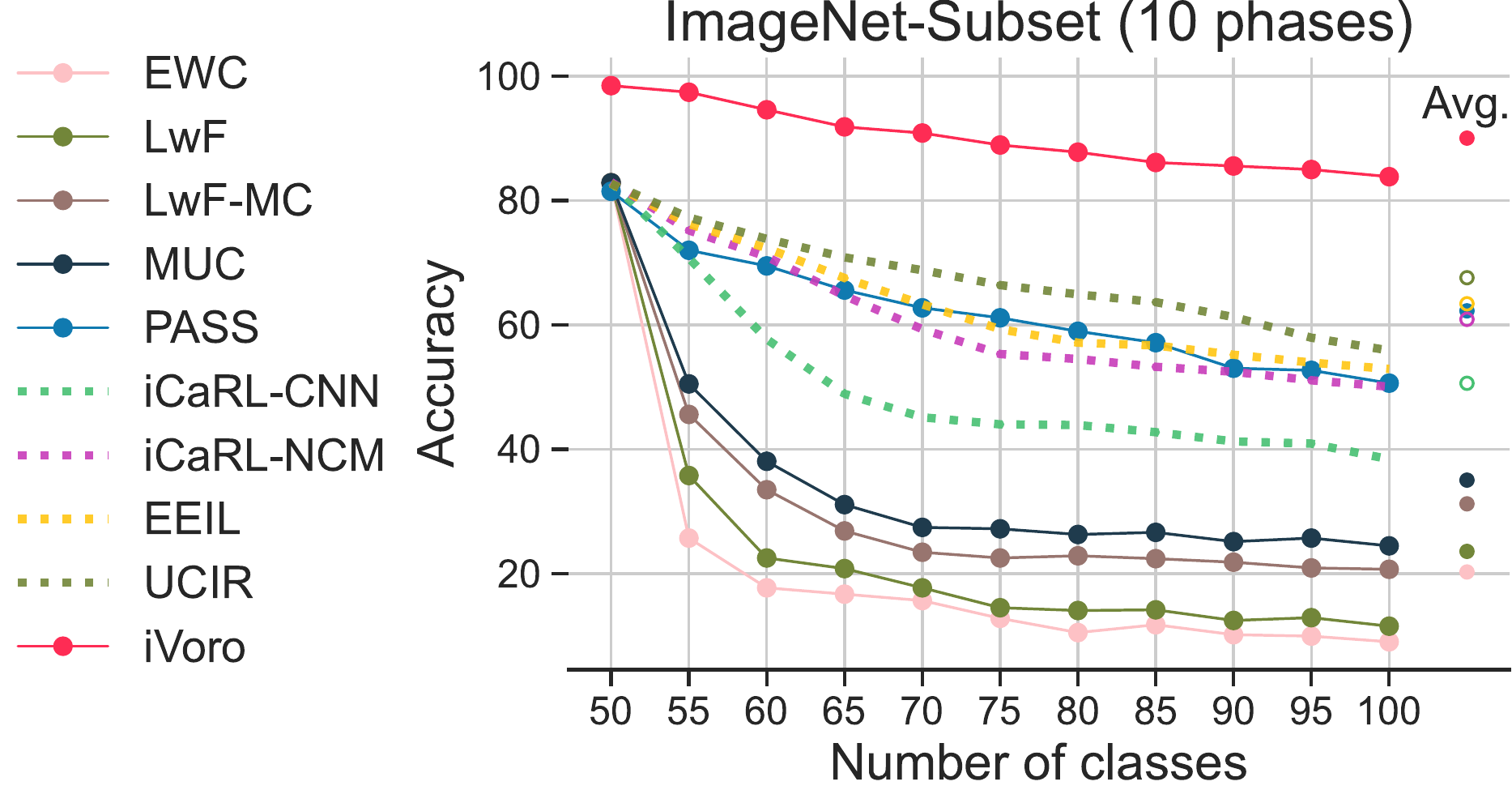}
    \caption{Top-1 classification accuracy on ImageNet-Subset during 10 phases of CIL.}\label{fig:subset}
\end{figure}
% ==================== Forgetting ====================
\clearpage
\section{Forgetting Analysis}\label{supp:forget}
\begin{figure}[ht]
    \centering
    \subfloat{\includegraphics[width=0.98\textwidth]{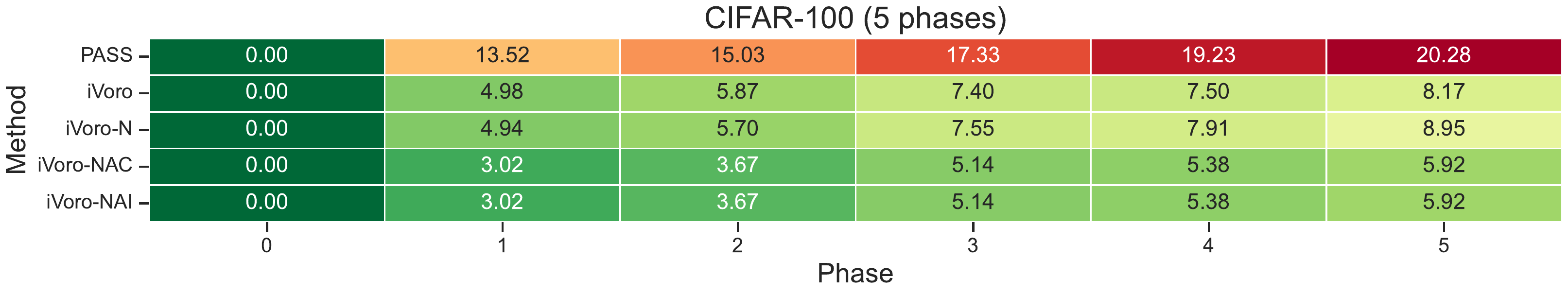}} \\ [-0.3ex]
    \subfloat{\includegraphics[width=0.98\textwidth]{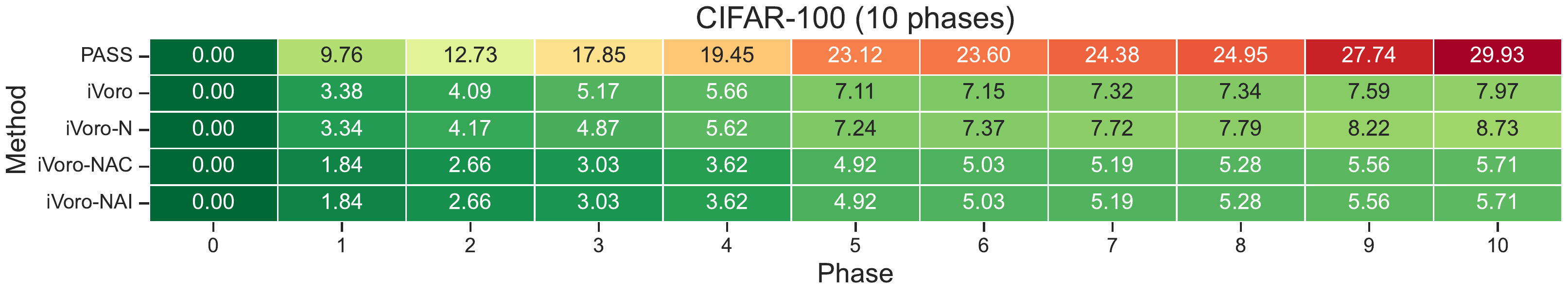}} \\ [-0.3ex]
    \subfloat{\includegraphics[width=0.98\textwidth]{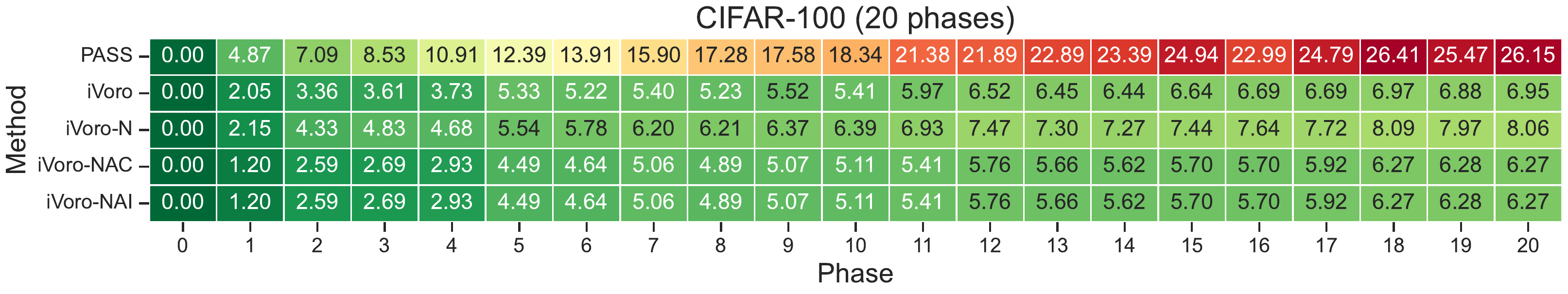}} \\
    \caption{Results of average forgetting on CIFAR-100.}\label{fig:forget-1}
\end{figure}

\begin{figure}[ht]
    \centering
    \subfloat{\includegraphics[width=0.98\textwidth]{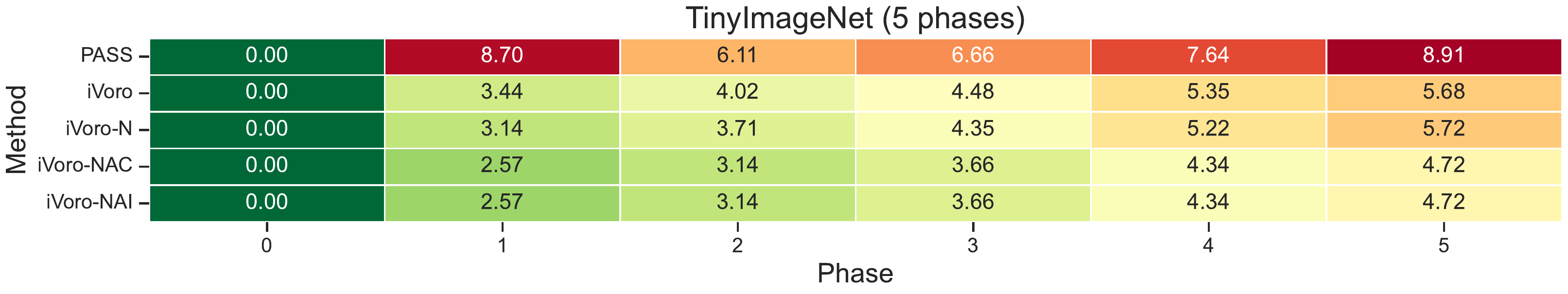}} \\ [-0.3ex]
    \subfloat{\includegraphics[width=0.98\textwidth]{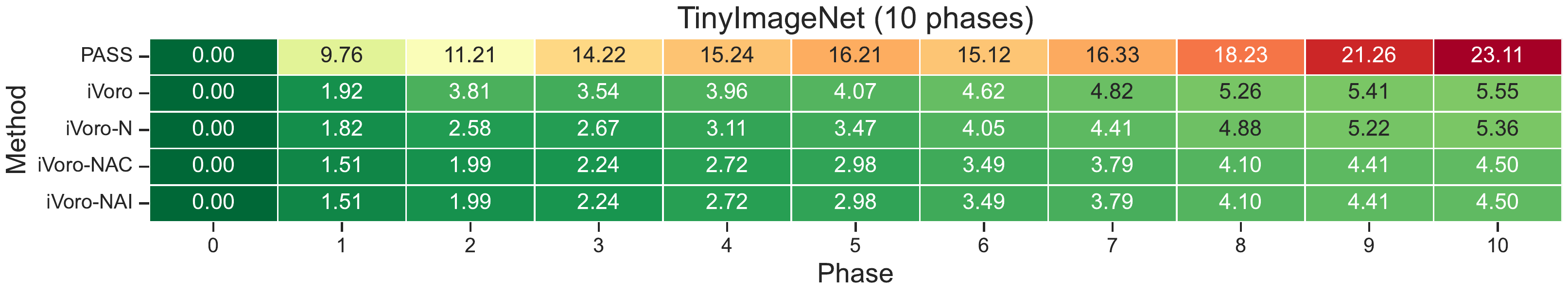}} \\ [-0.3ex]
    \subfloat{\includegraphics[width=0.98\textwidth]{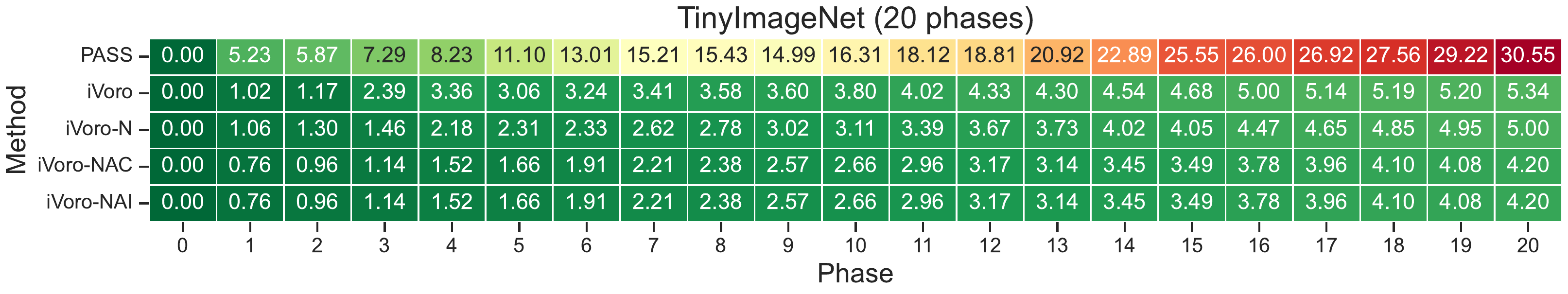}} \\ 
    \caption{Results of average forgetting on TinyImageNet.}\label{fig:forget-2}
\end{figure}

Following PASS~\citep{zhu2021prototype}, we also quantitatively measure the degree catastrophic forgetting. Specifically, at phase $t$, the accuracy drop from the maximum accuracy to current accuracy for datasets from each phase $\tau \in \{1,...,t\}$ is denoted as average forgetting, as shown in Fig.~\ref{fig:forget-1} and Fig.~\ref{fig:forget-2} for CIFAR-100 and TinyImageNet, respectively. For PASS, the average forgetting keeps growing during phases. However, for iVoro/iVoro-N/iVoro-NAC/iVoro-NAI, the catastrophic forgetting is significantly overcome. For example, on CIFAR-100 (5 phase), forgetting in the last phase is decreased from 20.28 to 8.17/8.95/5.92 for iVoro/iVoro-N/iVoro-NAC. For a local dataset $\gD_{\tau} = \{(\vx_{\tau,i}, y_{\tau,i})\}_{i=1}^{N_{\tau}}, \vx_{\tau,i} \in \sD, y_{\tau,i} \in \gC_{\tau}$ and a set of fixed prototypes $\{\vc_{\tau,k}\}_{\tau \in \{1,...,t\},k \in \{1,...,K_{\tau}\}}$, the prediction $\hat{y} = \argmin_{k \in \{1,...,K_{\tau}\}} ||\phi(\vx) - \vc_{\tau,k}||_2^2$ is less likely to change compared to continuously updated model, as in PASS, implying that, aligning with our intuition (Sec.~\ref{sec:intro}), the VD structure can naturally and successfully combat catastrophic forgetting.

\vspace{1em}

\vspace{2em}

\vspace{2em}
%Please add the following packages if necessary:
%\usepackage{booktabs, multirow} % for borders and merged ranges
%\usepackage{soul}% for underlines
%\usepackage[table]{xcolor} % for cell colors
%\usepackage{changepage,threeparttable} % for wide tables
%If the table is too wide, replace \begin{table}[!htp]...\end{table} with
%\begin{adjustwidth}{-2.5 cm}{-2.5 cm}\centering\begin{threeparttable}[!htb]...\end{threeparttable}\end{adjustwidth}
\begin{table}[!htp]\centering
\caption{Average Forgetting ($\downarrow$): Comparison between iVoro with state-of-the-art CIL methods}\label{tab:forget}
\small
\resizebox{0.8\textwidth}{!}{%
    \begin{tabular}{lc|c|c|c|c|c}\toprule
    &\multicolumn{3}{c}{\textbf{CIFAR100}} &\multicolumn{3}{c}{\textbf{TinyImageNet}} \\\cmidrule{2-7}
    Methods &\multicolumn{1}{l}{5 phases} 
    &\multicolumn{1}{l}{10 phases} 
    &\multicolumn{1}{l}{20 phases} 
    &\multicolumn{1}{l}{5 phases} 
    &\multicolumn{1}{l}{10 phases} 
    &\multicolumn{1}{l}{20 phases} \\\midrule
    \cmark{ }iCaRL\textsubscript{CNN}~\citep{rebuffi2017icarl} &42.13 &45.69 &43.54 &36.89 &36.7 &45.12 \\
    \cmark{ }iCaRL\textsubscript{CNN}~\citep{rebuffi2017icarl} &24.90 &28.32 &35.53 &27.15 &28.89 &37.40 \\
    \cmark{ }EEIL~\citep{castro2018end} &23.36 &26.65 &32.40 &25.56 &25.91 &35.04 \\
    \cmark{ }UCIR~\citep{hou2019learning} &21.00 &25.12 &28.65 &20.61 &22.25 &33.74 \\\midrule
    \xmark{ }LwF-MC~\citep{li2017learning} &44.23 &50.47 &55.46 &54.26 &54.37 &63.54 \\
    \xmark{ }MUC~\citep{liu2020more} &40.28 &47.56 &52.65 &51.46 &50.21 &58.00 \\
    \xmark{ }PASS~\citep{zhu2021prototype} &20.28 &29.93 &26.15 &8.91 &23.11 &30.55 \\
    \cellcolor[HTML]{f2f2f2}\xmark{ }iVoro (ours) &\cellcolor[HTML]{f2f2f2}\textbf{8.17} &\cellcolor[HTML]{f2f2f2}\textbf{7.97} &\cellcolor[HTML]{f2f2f2}\textbf{6.95} &\cellcolor[HTML]{f2f2f2}\textbf{5.68} &\cellcolor[HTML]{f2f2f2}\textbf{5.55} &\cellcolor[HTML]{f2f2f2}\textbf{5.34} \\
    \bottomrule
    \end{tabular}}
\end{table}

% ====================
\end{document}